 \documentclass[pmlr,twocolumn,10pt]{jmlr} 

\usepackage{booktabs}
\usepackage{siunitx}

\usepackage[switch]{lineno}

\usepackage{svg}
\usepackage{adjustbox}
\usepackage{multirow}
\usepackage{arydshln}
\usepackage{amssymb}

\newcommand{\equal}[1]{{\hypersetup{linkcolor=black}\thanks{#1}}}

\theorembodyfont{\upshape}
\theoremheaderfont{\scshape}
\theorempostheader{:}
\theoremsep{\newline}

\jmlrvolume{259}
\jmlryear{2024}
\jmlrsubmitted{LEAVE UNSET}
\jmlrpublished{LEAVE UNSET}
\jmlrworkshop{Machine Learning for Health (ML4H) 2024} 

 \title[Are Clinical T5 Models Better for Clinical Text?]{Are Clinical T5 Models Better for Clinical Text?}

\author{%
\Name{Yahan Li}\equal{These authors contributed equally} \Email{yahanli@usc.edu} \\
\addr Department of Computer Science, University of Southern California
\AND
\Name{Keith Harrigian}\footnotemark[1] \Email{kharrigian@jhu.edu} \\
\addr Department of Computer Science, Johns Hopkins University 
\AND
\Name{Ayah Zirikly} \Email{azirikly@jhu.edu} \\
\addr Center for Language and Speech Processing, Whiting School of Engineering, Johns Hopkins University \\
\addr Malone Center for Engineering in Healthcare, Johns Hopkins University 
\AND
\Name{Mark Dredze} \Email{mdredze@cs.jhu.edu} \\
\addr Department of Computer Science, Johns Hopkins University
}

\begin{document}

\maketitle

\begin{abstract}
Large language models with a transformer-based encoder/decoder architecture, such as T5 \citep{t5}, have become standard platforms for supervised tasks. To bring these technologies to the clinical domain, recent work has trained new \citep{Lehman2023DoWS} or adapted existing \citep{luClinicalT5} models to clinical data. However, the evaluation of these clinical T5 models and comparison to other models has been limited. Are the clinical T5 models better choices than FLAN-tuned \citep{flan-t5} generic T5 models? Do they generalize better to new clinical domains that differ from the training sets? We comprehensively evaluate these models across several clinical tasks and domains. We find that clinical T5 models provide marginal improvements over existing models, and perform worse when evaluated on different domains. Our results inform future choices in developing clinical LLMs.
\end{abstract}

\paragraph*{Data and Code Availability}
 We do not release any new data or models as part of this study. The code in available on GitHub: \url{https://github.com/yli-z/ml4h_are_clinical_t5_models_better_for_clinical_text}. 

\paragraph*{Institutional Review Board (IRB)}
Our clinical datasets were obtained from the PhysioNet website under the PhysioNet Credentialed Health Data Use Agreement and from an anonymized Hospital System following IRB approval. 

\section{Introduction}
There is an ongoing conversation in the community about the best strategy for developing Large Language Models (LLMs) for specialized domains. general-purpose LLMs, trained on massive amounts of diverse text, can generalize remarkably well to new domains and tasks \citep{brown2020language, touvron2023llama, bubeck2023sparks}, especially with a small amount of domain-specific alignment data \citep{singhal2022medpalm}. At the same time, specialized LLMs trained using domain-specific data can outperform their generic counterparts \citep{wu2023bloomberggpt, taylor2022galactica, Lehman2023DoWS, luClinicalT5}. 
For supervised tasks, Text-to-Text Transfer Transformer (T5) models are a popular choice as they provide the benefits of both extensive pre-training and task generalization\citep{t5}; however, there are few studies on the benefits of domain-specialized T5 training.

Consider clinical text from Electronic Health Records (EHR). Access to this data is severely restricted and the data differs substantially from popular pre-training data sets. 
Recent work has explored developing T5 models for this domain. 
One approach adapts an existing T5 model first to the biomedical domain (journal articles) and then to the clinical domain using a small amount of clinical text \citep{luClinicalT5}. Another approach starts from scratch, developing a new vocabulary and model focused on clinical data \citep{Lehman2023DoWS}. As these works were completed concurrently, they do not include a direct comparison between the approaches, leaving the best strategy for T5 models an open question.

Furthermore, while a specialized clinical T5 model sounds attractive, it has several potential shortcomings. First, the severe limitation of available EHR pre-training text means specialized T5 models are not only trained on less data overall, but also that the pre-training text is unlikely to adequately represent the true diversity of the domain. In contrast, pre-trained T5 models benefit from massive amounts of diverse data, which may yield better generalization abilities. Second, existing general T5 models may make use of extensive supervised datasets for FLAN tuning \citep{flan-t5}, potentially leading to better generalization to new supervised tasks. Specialized clinical T5 models do not have the same opportunity nor benefit. Does FLAN training mean better generalization, even without access to EHR pre-training? 

We thereby present a series of evaluations of existing general-purpose and clinical T5 models to answer two questions -- 1) What existing T5 model should clinical NLP practitioners utilize when training a supervised system, and 
2) What training strategy is more effective in the clinical domain with severe data limitations nowadays?

We make the following contributions.
    \begin{itemize}
        \item We evaluate 2 clinical, 2 non-clinical, and 1 FLAN-tuned T5 variants across 7 clinical and biomedical tasks. We independently reproduce prior results \citep{luClinicalT5, Lehman2023DoWS} and extend the evaluation to new clinical datasets. We find that general T5 models achieve similar results as the clinical variants.
        \item We investigate model performance under data-limited conditions and find that FLAN-T5 excels on clinical tasks in the low-data settings.
        \item We evaluate clinical T5 models on new clinical domains and show they perform worse than general T5 models, suggesting that existing clinical models are overfitted to the limited clinical datasets available.
    \end{itemize}

Taken together with other studies on clinical LLMs \citep{ alsentzer2019clinicalbert, lewis-etal-2020-bioclinroberta, singhal2022medpalm, luClinicalT5, Lehman2023DoWS}, our findings suggest that future work should focus on adapting general T5 models to the clinical domain through new supervised training sets rather than training new methods from scratch.

\section{Clinical T5 Models}

LLMs have become the standard base models upon which NLP systems are built. Instruction following tasks typically leverage general-purpose decode LLMs (e.g., GPT, LlaMa) \citep{Radford2019LanguageMA, brown2020language, touvron2023llama, bubeck2023sparks}, whereas supervised tasks typically rely on encoder/decoder \citep{t5} or encoder models \citep{devlin2019bert}. 

Historically, clinical NLP systems have benefited from specialized versions of these models due to differences in the topic, style, and vocabulary of clinical text relative to other language domains. However, these advantages have been tempered under various circumstances \citep{gutiérrez2023biomedical,lewis-etal-2020-bioclinroberta}, and are not necessarily guaranteed to hold given the existence of much larger contemporary LLMs. This uncertainty raises several key questions for clinical NLP practitioners.

\subsection*{What existing models should be used to develop a supervised clinical NLP system?}

T5 \citep{t5} and Flan-T5 \citep{flan-t5} do exceedingly well on a wide variety of supervised tasks, benefited from large amounts of pre-training and diverse task-specific fine-tuning, respectively. These models balance out-of-the-box zero-shot performance and supervised fine-tuning when a moderate amount of supervised data is available. Within the clinical domain, two T5-based models \citep{Lehman2023DoWS,luClinicalT5} have been trained on clinical text from MIMIC \citep{mimic3,mimic4}.
Which model should we prefer and in what setting?
Unfortunately, the publication of these two models was concurrent, so a head-to-head comparison is unavailable. Therefore, we conduct a direct comparison with the same training and evaluation settings to determine which model should be preferred.

\subsection*{Should we train from scratch on clinical data or adapt an existing pre-trained model?}

Beyond determining which model is preferred, these two models represent different strategies for developing domain-specific models:

\citet{luClinicalT5} utilize continued pre-training to tune an existing model for the clinical setting, which allows them to benefit from large amounts of previous training on general-purpose and related data. They use MIMIC-III \citep{mimic3} to train SciFive \citep{phan2021SciFive}, which is itself adapted to the scientific domain from the base T5 model \citep{t5}. In contrast, \citet{Lehman2023DoWS} train a new T5 model from scratch on clinical text from MIMIC-III and MIMIC-IV \citep{mimic3, mimic4}, which enables them to leverage a domain-specific vocabulary. 

These models follow larger trends in developing domain-specific models, including training from scratch \citep{wu2023bloomberggpt,taylor2022galactica, lewis-etal-2020-bioclinroberta} and continued pre-training  or aligning existing models \citep{alsentzer2019clinicalbert,singhal2022medpalm}.
In the clinical domain specifically, encoder (masked) LLMs have been adapted from pre-trained models (e.g., BioBERT \citep{Lee_2019_biobert} and ClinicalBioBERT \citep{alsentzer2019clinicalbert} from BERT \citep{devlin2019bert}) and have also been trained from scratch (e.g., GatorTron was pre-trained on a combination of Wikipedia articles, PubMed publications, and de-identified clinical notes \citep{yang2022gatortron}). Decoder (GPT-style) LLMs have been adapted to the medical domain, but not necessarily the clinical domain (e.g., Med-PaLm \citep{singhal2022medpalm} from FLAN-PaLm \citep{chowdhery2022palm}).

Arguments have been made in support of both pre-training strategies. We are interested in understanding whether one strategy is preferable over another when a) there is a need to generalize to a new clinical domain, and b) when only a limited amount of supervised data in the target domain is available.

\subsection*{Should we prefer domain-specialized models over powerful, general-purpose models? }

Domain-specificity might not be optimal: biomedical language models have been shown to be robust to domain-general tokenization \citep{gutiérrez2023biomedical}, and pre-trained LLMs have been shown to possess significant knowledge about the medical domain despite not having been trained specifically on biomedical data \citep{agrawal_etal_2022_large,Singhal2023-zo}. In the clinical domain, pre-trained clinical BERT models do not consistently demonstrate a significant improvement over non-clinical variants of BERT \citep{lewis2020pretrained,harrigian2023eye, yue-etal-2020-analysis-emrqa}. This suggests that the benefits of in-domain pre-training may not always justify the additional complexity and computational cost.

Morover, in the case of T5 models, there exist high-quality FLAN-tuned versions that do well on a variety of supervised tasks. While they are not clinically tuned, they do incorporate a lot of supervision. Should we prefer these general-purpose models to clinical models, especially in new clinical settings or those with limited supervision? We compare these models to clinical T5 models to determine which should be preferred in clinical settings.

Overall, our analysis provides practical guidance to practitioners, provides new evidence on LLM domain specialization methods, and explores the tradeoff between domain specialized and general-purpose LLMs for supervised tasks.

\section{Experiment Setup}

\subsection{Models}

Text-To-Text Transfer Transformer~(T5) \citep{t5} and its variants are sequence-to-sequence models utilizing an encoder/decoder Transformer architecture \citep{vaswani2017attention}. 
We evaluate several existing pre-trained models, all being the large variant ($\sim$770M parameters),  by fine-tuning them on $(x, y)$ pairs, where $x$ and $y$ are the input and output of a task, formatted as text. 

Additional information about each model is included in \tableref{model-settings}.

\paragraph{T5} The original T5 model was trained in two stages. First, it was trained with a self-supervised text denoising objective on web crawl data \citep{t5}, and then with a conditional generation objective on labeled data from a diverse mixture of downstream tasks. To separate the effects of pre-training using non-clinical web data and performing supervised multitask pre-training, we evaluate two versions of T5:  {\bf T5-Den} which is trained only using the denoising objective and {\bf T5-Sup} which is further pre-trained on supervised multi-task tasks.\footnote{Most studies only evaluate {T5-Sup} and call it a pre-trained only model.}

\paragraph{FLAN-T5} FLAN-T5 \citep{flan-t5} took the LM-Adapted T5, which was trained for additional 100K steps on the LM objective from T5-Den \citep{t5, lester-etal-2021-prompt-power}, and further trained with extensive supervised instruction fine-tuning. The instruction tuning process typically improves the model's generalization ability to new tasks \citep{longpre2023pretrainers_training}. We include FLAN-T5 to evaluate its generalization capabilities on clinical tasks.

\paragraph{Clinical-T5} We evaluate two concurrently developed Clinical-T5 models -- similarly named, but pre-trained on different corpora and with different weight initialization strategies. We refer to the model introduced by \citet{Lehman2023DoWS} as {MIMIC-T5} since their model is only pre-trained on the union of all notes from MIMIC-III and MIMIC-IV \citep{mimic3, mimic4}, including discharge summaries and radiology reports, from random initialization. We refer to the model introduced by \citet{luClinicalT5} as {SciFive+MIMIC-T5} since their model is trained on MIMIC-III \citep{mimic3} but initializes from a biomedically pre-trained model ({SciFive}), which gives it additional exposure to biomedical data. {SciFive} was itself trained from the base T5 initialization.

\subsection{Datasets} 
We evaluate models on the union of datasets considered by each of the clinical T5 studies \citep{Lehman2023DoWS,luClinicalT5}, as well as clinical data not from MIMIC-III.

\paragraph{MIMIC-III Datasets} We evaluate on CLIP \citep{CLIP}, a multi-label classification dataset, RadQA \citep{soni-etal-2022-radqa}, a question-answering dataset, and MedNLI \citep{romanov2018mednli}, a natural language inference dataset. Each dataset is comprised of instances from MIMIC-III \citep{mimic3}. Note that these datasets annotate data that comes from the same corpus used for clinical model pre-training.

\paragraph{Biomedical Datasets} We evaluate on HOC \citep{Baker2015}, a multi-label classification dataset, and two named entity recognition datasets, BC5CDR-disease \citep{Li2016-ud} and NCBI-disease \citep{Dogan2014-aw}. HOC and NCBI-disease are datasets derived from PubMed abstracts \footnote{https://pubmed.ncbi.nlm.nih.gov/}, and the BC5CDR-disease dataset is derived from full PubMed articles. Note that some of them annotate data that comes from the same corpus used for {SciFive+MIMIC-T5} pre-training.

\paragraph{Clinical Stigmatizing Language Datasets} Due to the extremely limited availability of clinical text data, public clinical NLP datasets are drawn from MIMIC (MIMIC-III \citep{mimic3} in particular), the same dataset used to pre-train the clinical-T5 models. How well do these clinical models generalize to a novel clinical domain and task? We obtain data from Hospital System (anonymized) that includes five medical specialties: Internal Medicine, Emergency Medicine, Pediatrics, OB-GYN, and Surgery. We evaluate the task of detecting stigmatizing language about patients by using the setup as \citet{harrigian-etal-2023-characterization}. We also use their annotations from MIMIC-IV.

\section{Are Clinical T5 Models Better for Clinical Text? } \label{sec:clinical-ehr}

\begin{table*}[t]
\centering
\adjustbox{width=\linewidth}{%
\begin{tabular}{ccccccc}
\toprule
     \textbf{Dataset} & \textbf{Metrics} & \textbf{{T5-Den}} & \textbf{{T5-Sup}} & \textbf{FLAN-T5} & \textbf{{MIMIC-T5}} & \textbf{{SciFive+MIMIC-T5}} \\ \midrule

   MedNLI                 & Acc. & $85.9_{(85.3, 86.2)}$   & $84.9_{(84.4, 85.5)}$ & $86.0_{(85.7, 86.4)}$ & $86.8_{(86.3, 87.3)}$  & $85.6_{(85.1, 86.4)}$ \\ \midrule
\multirow{2}{*}{RadQA} & EM       & $52.6_{(51.9, 53.3)}$  & $52.0_{(51.0, 53.0)}$& $53.0_{(51.8, 54.0)}$  & $54.5_{(54.0, 55.1)}$  & $53.4_{(52.7, 54.2)}$ \\
                       & F1       & $68.9_{(68.0, 69.8)}$  & $68.7_{(67.8, 69.8)}$ & $70.6_{(69.8, 71.3)}$   & $73.4_{(72.8, 74.0)}$    & $70.4_{(69.4, 71.5)}$ \\ \midrule
\multirow{2}{*}{CLIP}   & Macro F1 & $63.9_{(62.2, 65.6)}$   & $62.4_{(61.4, 63.4)}$ & $64.3_{(62.9,  65.7)}$ & $66.1_{(65.9, 66.3)}$ & $63.5_{(62.2, 64.8)}$ \\
                       & Micro F1 & $78.6_{(77.5, 79.8)}$ & $78.4_{(78.0, 78.8)}$  & $79.4_{(78.9, 79.9)}$   & $79.3_{(78.7, 79.9)}$  & $78.2_{(77.9, 78.5)}$ \\ \midrule
\multirow{3}{*}{HOC}   & F1       & $84.9_{(84.4, 85.6)}$ & $84.8_{(84.4, 85.1)}$ & $84.8_{(84.2, 85.3)}$  & $82.8_{(82.6, 83.0)}$ & $85.1_{(84.8, 85.4)}$  \\
                       & P        & $84.7_{(84.2, 85.6)}$ & $84.8_{(84.4, 85.1)}$ & $84.6_{(83.9, 85.2)}$  & $82.9_{(82.8, 83.1)}$ & $85.0_{(84.7, 85.5)}$ \\
                       & R        & $85.1_{(84.7, 85.6)}$ & $84.8_{(84.3, 85.2)}$ & $85.0_{(84.5, 85.4)}$  & $82.6_{(82.3, 83.0)}$ & $85.2_{(84.9, 85.5)}$ \\ \midrule
\multirow{3}{*}{BC5CDR} & F1       & $82.9_{(82.3, 83.4)}$ & $82.9_{(82.5, 83.5)}$ & $83.3_{(82.9, 83.8)}$  & $81.2_{(80.9, 81.5)}$ & $83.6_{(83.4, 83.8)}$ \\
                       & P        & $81.7_{(80.8, 82.5)}$ & $81.6_{(81.3, 82.0)}$ & 
                       $82.1_{(81.5, 82.7)}$& $80.6_{(80.2, 80.9)}$ & $83.1_{(82.5, 83.6)}$  \\
                       & R        & $84.2_{(83.6, 84.6)}$ & $84.3_{(83.6, 85.0)}$ & 
                       $84.5_{(84.2, 84.9)}$& $81.7_{(81.2, 82.3)}$ & $84.2_{(83.6, 84.8)}$ \\ \midrule
\multirow{3}{*}{NCBI}  & F1       &$85.1_{(84.6, 85.5)}$ & $84.6_{(84.2, 85.0)}$ & $84.4_{(83.9, 85.0)}$  & $79.1_{(78.3, 79.9)}$ & $85.4_{(85.0, 86.2)}$ \\
                       & P        & $83.5_{(82.8, 84.2)}$ & $83.6_{(83.1, 84.1)}$ & $82.2_{(81.6, 83.0)}$  & $78.4_{(77.6, 78.9)}$ & $83.8_{(83.0, 84.7)}$\\
                       & R        & $86.7_{(86.3, 87.2)}$ & $85.6_{(85.2, 86.2)}$ & $86.8_{(86.3, 87.3)}$  & $79.9_{(78.7, 81.3)}$ & $87.1_{(86.9, 87.5)}$ \\
\midrule
\end{tabular}
}
\caption{Mean test performance and 95\% confidence intervals across 5-fold cross-validation. MIMIC-T5 outperforms alternative models on clinical tasks, but struggles on biomedical tasks. Inclusion of biomedical data during pre-training is useful for the biomedical tasks.}
\label{tab:main_cv_table}
\end{table*}

Both {MIMIC-T5} and {SciFive+MIMIC-T5} have demonstrated improved performance on clinical EHR text compared to a general-purpose model \textsc{T5}. Our first step is to replicate these findings with a more extensive and direct evaluation.

\paragraph{Methods} We began by reproducing the results reported in \citet{Lehman2023DoWS} and \citet{luClinicalT5}, adhering closely to each study's original evaluation methodology. Both studies only leveraged a single train/dev/test split for each respective task, and only \citet{Lehman2023DoWS} estimated uncertainty around their performance point estimates.\footnote{\citet{Lehman2023DoWS} used at least three random seeds in their training procedure to estimate uncertainty.} To understand whether prior results were sensitive to the initial dataset split, we conducted an additional set of cross-validation experiments. We merged existing training and development splits and randomly split them into 5 subsets to facilitate 5-fold cross-validation. Additionally, the original held-out test split was used to evaluate generalization and compare to prior work. We considered the same evaluation metrics used in prior work (\tableref{tab:dataset-statistics}), using a bootstrap procedure to estimate confidence intervals. For uniformity across the metrics (e.g., exact match, F1 score), differences in performance between models were assessed using a paired t-test (paired by fold). We referred to MIMIC-T5 and SciFive-T5's evaluation metrics on these datasets. Training and evaluation details are included in \appendixref{sec:appendix_training_details}.

\paragraph{Results} Cross-validation results are reported in \tableref{tab:main_cv_table}, while reproducibility efforts are reported alongside the originally reported performance metrics in \appendixref{sec:reproducibility_experiments}. Our results for MIMIC-T5 perform slightly worse than previously reported on MedNLI and RadQA, but better on CLIP. SciFive+MIMIC-T5 performs slightly better than previously reported on the three biomedical tasks. The average performance over the cross-validation procedure mirrored the single train/dev/test split results.

Despite minor differences from the originally-reported results, we confirm the findings of \citet{Lehman2023DoWS} that MIMIC-T5 outperforms T5-Sup across the three clinical tasks at a statistically significant level. In comparison, SciFive+MIMIC-T5 either slightly underperformed or slightly outperformed the non-clinical T5 variants across the clinical tasks, albeit not at a statistically significant level. While the former outcome seems to corroborate the efficacy of domain-specific clinical pre-training, the latter outcome suggests the effect may be moderated inconsistently by pre-training on out-of-domain text. Given that MIMIC-T5 was pre-trained on text from the same corpus used for curating the evaluation datasets, the observed improvement may be a consequence of overfitting, which we explore below. Moreover, we note that the absolute increase in performance ($\sim$1.0 to $\sim$1.5 points across clinical tasks) falls within a range that is reasonable to expect when using domain adaptive pre-training (i.e., continued pre-training) \citep{gururangan2020dont}.

Results on biomedical (journal article) datasets highlight the differences between the adaptation and train from scratch strategies. SciFive+MIMIC-T5 does better than MIMIC-T5 on these tasks, sometimes by more than 4 points, presumably since it was adapted from SciFive which is trained on similar data, whereas MIMIC-T5 has only seen clinical data. However, SciFive+MIMIC-T5 only slightly outperforms the general-purpose models, and does not do so consistently at a statistically significant level. It is possible that the general-purpose models have already seen data similar enough to the biomedical data such that further pre-training has minimal effect. 

Finally, we observe the benefits of supervised instruction fine-tuning. FLAN-T5 performs comparably to MIMIC-T5 on the three clinical datasets and to SCIFIVE+MIMIC-T5 on the three biomedical datasets. MIMIC-T5 only outperforms FLAN-T5 on RadQA dataset, and only under 1 metric (F1 score), with statistical significance. Recent work suggests that training on relevant target tasks, either through instruction tuning or multitask training, has broad downstream benefits \citep{chung2022scaling,mueller2022text}. T5-Den and T5-Sup perform similarly on biomedical tasks, but not the clinical tasks, where T5-Den outperforms T5-Sup slightly, albeit not at a statistically significant level. Whether the remaining gap in performance between non-clinical and clinical T5 models can be decreased via supervised pre-training on related tasks or extensive pre-training in general is an important open area of study, 
especially given that non-clinical task datasets are significantly more abundant than clinical ones.

\begin{table*}[t!]
\adjustbox{width=\linewidth}{
\begin{tabular}{lcccccccc}
\toprule
~ & \multicolumn{4}{c}{Hospital System (anonymized)} & \multicolumn{4}{c}{MIMIC-IV} \\
\cmidrule(lr){2-5} 
\cmidrule(lr){6-9}
\textbf{Model} & \begin{tabular}{@{}c@{}}Credibility \&\\Obstinacy\end{tabular}  & Compliance & Descriptors & \textbf{Average} & \begin{tabular}{@{}c@{}}Credibility \&\\Obstinacy\end{tabular} & Compliance & Descriptors & \textbf{Average}  \\
\midrule
T5-Sup & $88.2_{(86.1, 90.6)}$ & $86.7_{(85.5, 88.2)}  $ & $90.6_{(89.3, 91.4)}  $ & $\textbf{88.5}_{(87.8, 89.0)}$ & $76.0_{(74.1, 77.9)}$ & $92.9_{(91.7, 94.2)}$ & $85.7_{(83.4, 88.0)}$ & ${84.9}_{(84.2, 85.8)}$\\
MIMIC-T5 & $88.1_{(83.1, 94.1)}$ & $84.7_{(81.0, 88.7)}$ & $88.9_{(88.2, 89.5)}$ & $87.2_{(85.1, 89.7)}$ & $75.0_{(72.1, 77.8)}$ & $91.6_{(90.4, 92.7)}$ & $84.8_{(81.9, 88.2)}$ & $83.8_{(82.5, 84.5)}$\\
SciFive+MIMIC-T5 & $87.3_{(82.1, 91.8)}$ & $87.2_{(86.8, 87.6)}$ & $90.0_{(89.2, 90.7)}$ & $88.2_{(86.6, 89.2)}$ & $74.6_{(70.6,  78.8)}$ & $91.0_{(89.9, 92.1)}$ & $86.3_{(85.0, 87.8)}$ & $84.0_{(82.0, 85.8)}$\\
\hdashline\noalign{\vskip 0.5ex}
FLAN-T5 & $90.4_{(88.3, 92.3)}$ & $ 88.0_{(86.8, 88.9)}$ & $86.8_{(80.0, 90.7)}$ & ${88.4}_{(86.6, 90.0)}$ & $77.4_{(75.0, 79.3)}$ & $92.9_{(91.3, 94.5)}$ & $86.3_{(85.3, 87.0)}$ & $\mathbf{85.6}_{(85.0, 86.1)}$ \\
\bottomrule
\end{tabular}}
\caption{Macro F1 scores on stigmatizing language datasets from the Hospital System (anonymized) and MIMIC-IV.}
\label{stigma_table}
\end{table*}

\section{Do Clinical T5 Models Generalize to New Clinical Text?}

A key limitation in developing clinical NLP systems is the lack of publicly available data due to PHI concerns. Mostly public available clinical models, including MIMIC-T5 and SCIFIVE+MIMIC-T5, and evaluation sets utilize MIMIC, so we do not know if they generalize to new clinical text sources. 
The weaker performance of MIMIC-T5 on the biomedical datasets may suggest its limited ability to generalize, or perhaps its benefits remain when focused on clinical EHR text.
Since the clinical pre-training and evaluation sets represent only a single medical domain from a single institution, we ask: do clinical T5 models generalize to new clinical text sources and tasks better than general-purpose models?

\paragraph{Methods} To evaluate generalization, we consider a new task for which we can evaluate on both a MIMIC dataset and a Hospital System dataset. The task is to characterize stigmatizing language in medical records, which includes three multi-class classification tasks that operate similarly to word sense disambiguation. 
The two datasets were curated using the annotation guidelines set forth by \citet{harrigian-etal-2023-characterization}. 
We perform 5-fold cross-validation experiments as above on these two datasets, and compare the in-domain (MIMIC-IV) and out-of-domain (Hospital System (anonymized)) performance.

\paragraph{Results}
\tableref{stigma_table} shows the performance of the T5 models on the two stigmatizing language datasets. On the MIMIC-IV dataset, T5-Sup, MIMIC-T5, and SciFive+MIMIC-T5 achieve similar levels of performance. 
For this task, the benefits of MIMIC-T5 do not materialize, despite the MIMIC-IV dataset being within the distribution of its pre-training data.\footnote{The MIMIC-IV stigmatizing language dataset may have been seen verbatim during MIMIC-T5 pre-training, as they did not leverage the same evaluation splits.} 

When we consider the out-of-distribution Hospital System (anonymized) data, MIMIC-T5 falls behind these other two models, perhaps either because MIMIC-T5 has been overfit to MIMIC data, or because it lacks the large-scale pre-training of the other models. SciFive+MIMIC-T5 performs slightly better than MIMIC-T5 on these two clinical datasets. However, neither MIMIC-T5 or SciFive+MIMIC-T5 show improvement over T5-Sup or FLAN-T5.

Finally, these evaluations further highlight the generalization ability of T5-Sup and FLAN-T5, which do better than clinical T5 models on both datasets. Even without clinical tasks in hand, the additional supervised training yields a model that does better on a novel clinical task across both data sources.

\section{Do Clinical T5 Models Perform Well in Low-Resource Settings?} \label{low-resource}

One of the possible reasons we have observed only a small performance gap between general-purpose and clinical models may be the presence of sufficient task training data. Indeed, previous work has found that tuning general-purpose BERT models for a clinical task using supervised data can nullify a clinical model's advantage \citep{harrigian2023eye}. In this regard, we may expect to see larger benefits to clinical models in a low-resource setting. Importantly, such settings are common in clinical tasks, where annotation typically requires advanced expertise, or is nigh impossible due to data privacy constraints. We therefore explore if the advantages of clinical models emerge more prominently in low-resource settings.

\paragraph{Methods} We selected MedNLI and the two stigmatizing language datasets to test a low-resource setting. For MedNLI, we used the same cross validation splits as before, but downsampled the training subset to 1\% ($\sim$99 examples on average per cross validation fold). For the stigmatizing language datasets, which has less data than MedNLI to begin with, we were able to explore multiple downsampled training size settings -- 1\% ($\sim$2  examples on average per fold), 5\% ($\sim$12 examples on average per fold), and 25\% ($\sim$59 examples on average per fold) for every task. We note that 1\% of training data represents an extreme scenario, albeit one that is not inconceivable given the complexities of annotating some clinical datasets \citep{clinical_text_review}.
Experimental details are described in the \appendixref{sec:appendix_training_details}.

\paragraph{Results} 

\begin{figure}[h]
\centering
\includegraphics[width=\linewidth]{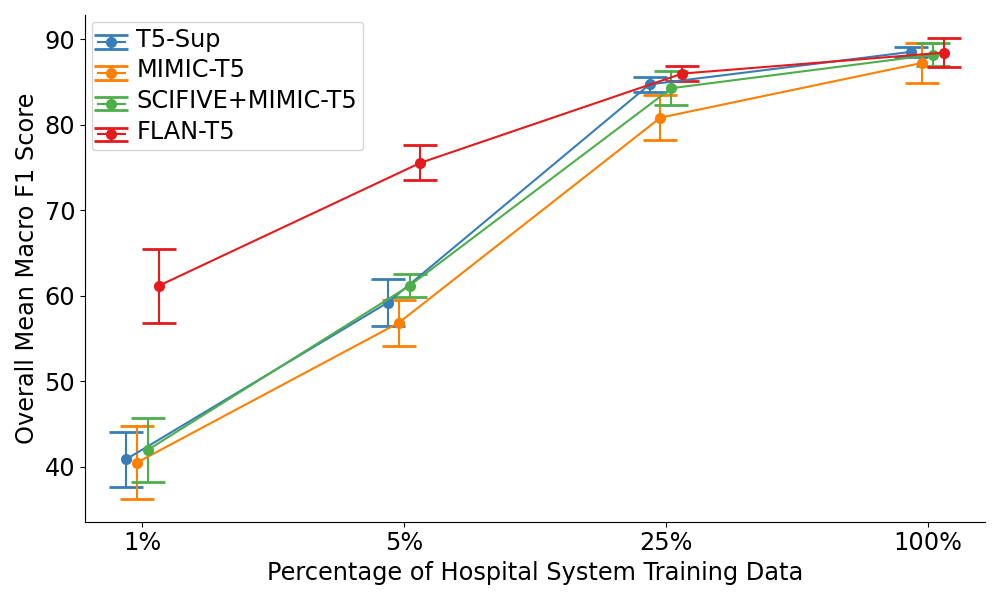}
\caption{Mean macro-F1 scores across the three stigmatizing language classification tasks for the Hospital System (anonymized) dataset using a random 1\%, 5\%, and 25\% sample of the training data.}
\label{fig:internal-downsampling-results}
\end{figure}

\begin{figure}[h]
\centering
\includegraphics[width=\linewidth]{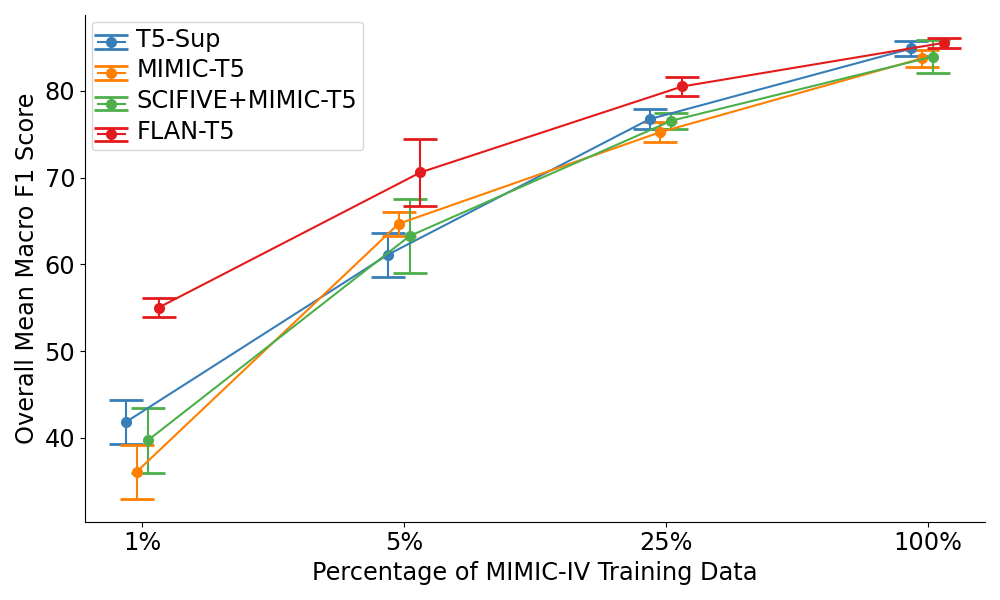}
\caption{Mean macro-F1 scores across the three stigmatizing language classification tasks for the MIMIC-IV dataset using a random 1\%, 5\%, and 25\% sample of the training data.}
\label{fig:mimic-downsampling-results}
\end{figure}

\begin{table}[h]
    \centering
    \begin{tabular}{cc}
  \toprule
  Model & Accuracy \\
  \midrule
       T5-Sup &  $78.80_{(77.69, 79.90)}$ \\
       MIMIC-T5 & $63.87_{(60.35, 66.95)}$\\
       SciFive+MIMIC-T5 & $78.92_{(78.35, 79.84)}$ \\
       \hdashline
       FLAN-T5 & $79.49_{(78.97, 80.06)}$ \\
    \bottomrule
    \end{tabular}
    \caption{Performance with 1\% training data on MedNLI dataset }
    \label{tab:downsampled_mednli_table}
\end{table}

\figureref{fig:internal-downsampling-results} and \figureref{fig:mimic-downsampling-results} show results for the Hospital System (anonymized) and MIMIC-IV stigmatizing language datasets, respectively, while \tableref{tab:downsampled_mednli_table} shows results for MedNLI. Additional task-specific results are reported in the \appendixref{sec:appendix_cross_validation_results}. 

In the 1\% MedNLI experiment, MIMIC-T5 is no longer leading, nor is it even competitive with, the other T5 models. For the two stigmatizing language datasets, T5-Sup, MIMIC-T5, and SciFive+MIMIC-T5 all perform roughly equivalently to one another in the low-resource settings as we gradually increase the sample size. That said, MIMIC-T5 finds itself at the bottom of relative performance rankings in more cases than not. Across all three tasks, FLAN-T5 excels in the low-resource settings compared to the three non-FLAN-tuned T5 models. Altogether, our results suggest that strong general-purpose language models are more appropriate for new distributions and those with limited amounts of task-specific training data.

\section{Discussion}

In this study, we independently replicated and extended prior work concerning specialized clinical T5 models \citep{luClinicalT5,Lehman2023DoWS}. 
By placing the models in an expanded context (i.e., additional non-clinical variants, evaluation datasets), we find ourselves in a position to make the following recommendations for clinical NLP practitioners.

\subsection*{Clinical models pre-trained from scratch using comparatively small datasets should not be used beyond their pre-training distributions.}

Language models that are pre-trained on the target domain from scratch can achieve strong target-domain performance due to their in-domain pre-training \citep{gupta2023warmup}. For example, PubMedBERT \citep{pubmedbert}, which was pre-trained from scratch on 3.2 billion words from PubMed Abstracts only, achieved remarkable performance in biomedical tasks drawn also from PubMed Abstracts. Unfortunately, large amounts of diverse and relevant pre-training data are generally hard to obtain for most real-world clinical applications.

Currently, MIMIC provides the largest public clinical text corpora, coming in at approximately 1.2 billion words across both MIMIC-III and MIMIC-IV \citep{Lehman2023DoWS}. However, practitioners must remember that MIMIC only represents a small subset of the broader clinical text landscape. Not only are there sample biases related to the demographics of patients in the corpora, but also topic biases (e.g., intensive care unit patients) and syntactic biases (e.g., hospital-specific documentation practices). LLMs trained on this comparatively narrow sample of data provide optimistic estimates of performance when evaluated on data also drawn from the MIMIC corpora \citep{Lehman2023DoWS}. As shown in our experiments, they struggle to adapt to the diverse needs of the clinical and biomedical space. So-called ``clinical'' language models may perhaps be better described as ``MIMIC'' language models. 

\subsection*{Clinical models should only be used in downstream tasks if they have access to sufficient supervised training data.}

In low-data settings, specificalized clinical models are more likely to underperform their generic counterparts that have been pre-trained on more data and/or using supervised fine-tuning. For example, in \S \ref{low-resource}, we saw that T5-Sup, SciFive+MIMIC-T5, and FLAN-T5 were all more qualified to operate in at the 1\% downsample regime than MIMIC-T5. Even larger generic LLMs such as ChatGPT and PaLM are likely to provide even more significant advantages in low-resource settings \citep{agrawal_etal_2022_large,Singhal2023-zo}. This leads us to our next recommendation.

\subsection*{The combination of task fine-tuning and FLAN instruction fine-tuning is hard to beat.} 

Recent research has shown that models pre-trained in a supervised manner are explicitly guided towards flat loss regions. Consequently, they are more robust to sequential fine-tuning in new domains \citep{mehta2023empirical}. It has also been shown that heterogeneity within pre-training data is important for promoting generalization in downstream tasks across diverse domains \citep{longpre2023pretrainers_training}. FLAN-T5's exceptional performance across all tasks in our study, especially in low-resource settings, affirms these prior findings. 

\subsection*{Adapting general-purpose models rather than training from scratch, or leveraging a mix of both training strategies, may be the best option for clinical language models moving forward.} 

General-domain pre-training is beneficial given that clinical text has limited availability, and domain-adaptive or task-adaptive training can further improve performance on downstream tasks \citep{gururangan2020dont}. For example, continued pre-training on small amounts of annotated clinical data (i.e., task-adaptive pre-training) has improved transfer to out-of-domain clinical data in a note-section classification task \citep{jamia_clinical_continued_pretraining}, while adapting a generic language model, such as BERT, has been shown to be more efficient than pre-training a new model from scratch in unseen target domains \citep{lamproudis_etal_2022_evaluating_bert_pretraining}. Various contemporary works have presented ways to further optimize continued pre-training for adaptation. For example, by reverse engineering the effects of instruction-tuning \citep{fleshman2024readapt}, or by finding more efficient warm-up strategies \citep{gupta2023warmup}.

\section{Conclusion}

Our study aimed to determine whether existing clinical T5 models offer performance improvements over non-clinical T5 variants in clinical tasks. Our findings suggest that these clinical language models may outperform their general counterparts, but only under specific conditions: a) the task is on clinical EHR data sourced from MIMIC; b) sufficient annotated training data is available. Meanwhile, general-purpose language models, especially those leveraging supervised multi-task instructing tuning (e.g., FLAN-T5), excel when these criteria are not met.

More broadly, our results \emph{do} provide support to previous claims that truly ``domain-specific'' language models, trained for specific data distributions and tasks, are ideal \citep{Lehman2023DoWS}. However, they also suggest that domain-specific models are not always practical. Data distributions may change over time (e.g., suddenly due to a pandemic, gradually as clinical practice evolves) \citep{khanday2020machine,jeong2024recent} or across target populations \citep{harrigian2023eye}. Likewise, not all domains are data rich and diverse enough to support training a language model from scratch that endows broader linguistic abilities. In the clinical domain, non-clinical language models will almost certainly continue to have access to pre-training corpora that are orders of magnitude larger than clinical pre-training corpora, and that data scale is greatly advantageous for training high-performing LLMs \citep{kaplan2020scaling}.

Lastly, we would be remiss not to comment on the cost and sustainability implications of training domain-specific language models. While it is true that given the same inference budget, a clinical model trained from scratch is typically able to achieve better performance than a generic model \citep{Lehman2023DoWS}, such a perspective regarding model efficiency is inherently narrow. Using floating point operations (FLOPs) at inference time as a proxy for efficiency does not accurately reflect the additional cost complexities of training a language model from scratch -- e.g., training compute, human labor (development and annotation), data access, and dealing with future distribution shift. Put another way, specialized models may achieve improvements in performance that are statistically significant, but not necessarily practically significant.

\section* {Limitations}

\paragraph{T5 Model Family} Our study focused on clinically pretrained T5 models, made possible by the released models from \citet{luClinicalT5} and \citet{Lehman2023DoWS}. Comparing them allowed us to study the effect of pretraining strategies while controlling for the architecture. Future work could extend these investigations to other pretrained model architectures.

\paragraph{Data Availability} Due to the scarcity of clinical text datasets, our evaluation was limited to a total of seven datasets. This constraint might not provide a comprehensive assessment for a wide range of clinical tasks. We have cited relevant work where applicable to supplement our findings.

\paragraph{Hyperparameter Search} Due to computational costs associated with training each T5 model over multiple cross validation folds, we were only able to perform limited amounts of hyperparameter tuning at the onset of the study. Upon identifying a reasonable set of hyperparameters that worked well across the different datasets and models, we opted to fix them across all the cross validation runs. It is certainly possible that different hyperparameters may have yielded different outcomes. That said, the computational limitations faced here reflect what many practitioners face.

\paragraph{Statistical Measures} We are aware that traditional statistical methods in NLP may not be sufficient to describe the divergence of performances in the clinical domain. While these statistical methods are effective in establishing the statistical significance of performance differences between models, they may fall short of providing a comprehensive understanding of how these differences manifest in real-world clinical settings. This limitation underscores the need for specialized evaluation frameworks in healthcare applications, thereby ensuring that the models not only perform well statistically but also deliver practical, reliable, and clinically relevant outcomes.

\paragraph{Future Directions} The clinical domain presents unique challenges and opportunities. Future work could consider extending model architectures and training strategies \citep{gema2024parameterefficientfinetuningllamaclinical, labrak2024biomistralcollectionopensourcepretrained, toma2023clinicalcamelopenexpertlevel, wu2023pmcllama}, evaluation \citep{Tang_Sun_et_al._2023_medical_evidence_summarization, Hager_Jungmann_Holland_Bhagat_Hubrecht_Knauer_Vielhauer_Makowski_Braren_Kaissis_et_al._2024, ness2024medfuzzexploringrobustnesslarge}), and techniques for addressing data scarcity and privacy issues \citep{lin-etal-2022-fednlp}.

\section*{Ethics Statement}

Our clinical datasets were obtained from the PhysioNet website, accessed under the PhysioNet Credentialed Health Data Use Agreement, and from Hospital System (anonymized) after IRB approval. All data from MIMIC-III or MIMIC-IV has been de-identified to ensure privacy and confidentiality. Clinical-T5 models were pre-trained on MIMIC; we access and use them under the PhysioNet Credentialed Health Data license and data use agreement. All data processing and model training was done on remote servers secured using OS-level security protocols.

\acks{This work was supported by the National Institute on Minority Health and Health Disparities under grant number R01 MD017048. The content is
solely the responsibility of the authors and does not
necessarily represent the official views of NIMHD,
NIH, or Johns Hopkins University.

We thank Eric Lehman, Qiuhao Lu, and Long Phan for clarifications about their code and experiments.
}

\bibliography{main}

\begin{thebibliography}{64}
\providecommand{\natexlab}[1]{#1}
\providecommand{\url}[1]{\texttt{#1}}
\expandafter\ifx\csname urlstyle\endcsname\relax
  \providecommand{\doi}[1]{doi: #1}\else
  \providecommand{\doi}{doi: \begingroup \urlstyle{rm}\Url}\fi

\bibitem[Agrawal et~al.(2022)Agrawal, Hegselmann, Lang, Kim, and Sontag]{agrawal_etal_2022_large}
Monica Agrawal, Stefan Hegselmann, Hunter Lang, Yoon Kim, and David Sontag.
\newblock Large language models are few-shot clinical information extractors.
\newblock In Yoav Goldberg, Zornitsa Kozareva, and Yue Zhang, editors, \emph{Proceedings of the 2022 Conference on Empirical Methods in Natural Language Processing}, pages 1998--2022, Abu Dhabi, United Arab Emirates, December 2022. Association for Computational Linguistics.
\newblock \doi{10.18653/v1/2022.emnlp-main.130}.
\newblock URL \url{https://aclanthology.org/2022.emnlp-main.130}.

\bibitem[Alsentzer et~al.(2019)Alsentzer, Murphy, Boag, Weng, Jin, Naumann, and McDermott]{alsentzer2019clinicalbert}
Emily Alsentzer, John~R. Murphy, Willie Boag, Wei-Hung Weng, Di~Jin, Tristan Naumann, and Matthew B.~A. McDermott.
\newblock Publicly available clinical bert embeddings, 2019.

\bibitem[Baker et~al.(2015)Baker, Silins, Guo, Ali, H\"{o}gberg, Stenius, and Korhonen]{Baker2015}
Simon Baker, Ilona Silins, Yufan Guo, Imran Ali, Johan H\"{o}gberg, Ulla Stenius, and Anna Korhonen.
\newblock Automatic semantic classification of scientific literature according to the hallmarks of cancer.
\newblock \emph{Bioinformatics}, 32\penalty0 (3):\penalty0 432–440, October 2015.
\newblock ISSN 1367-4803.
\newblock \doi{10.1093/bioinformatics/btv585}.
\newblock URL \url{http://dx.doi.org/10.1093/bioinformatics/btv585}.

\bibitem[Brown et~al.(2020)Brown, Mann, Ryder, Subbiah, Kaplan, Dhariwal, Neelakantan, Shyam, Sastry, Askell, Agarwal, Herbert-Voss, Krueger, Henighan, Child, Ramesh, Ziegler, Wu, Winter, Hesse, Chen, Sigler, Litwin, Gray, Chess, Clark, Berner, McCandlish, Radford, Sutskever, and Amodei]{brown2020language}
Tom~B. Brown, Benjamin Mann, Nick Ryder, Melanie Subbiah, Jared Kaplan, Prafulla Dhariwal, Arvind Neelakantan, Pranav Shyam, Girish Sastry, Amanda Askell, Sandhini Agarwal, Ariel Herbert-Voss, Gretchen Krueger, Tom Henighan, Rewon Child, Aditya Ramesh, Daniel~M. Ziegler, Jeffrey Wu, Clemens Winter, Christopher Hesse, Mark Chen, Eric Sigler, Mateusz Litwin, Scott Gray, Benjamin Chess, Jack Clark, Christopher Berner, Sam McCandlish, Alec Radford, Ilya Sutskever, and Dario Amodei.
\newblock Language models are few-shot learners, 2020.

\bibitem[Bubeck et~al.(2023)Bubeck, Chandrasekaran, Eldan, Gehrke, Horvitz, Kamar, Lee, Lee, Li, Lundberg, Nori, Palangi, Ribeiro, and Zhang]{bubeck2023sparks}
Sébastien Bubeck, Varun Chandrasekaran, Ronen Eldan, Johannes Gehrke, Eric Horvitz, Ece Kamar, Peter Lee, Yin~Tat Lee, Yuanzhi Li, Scott Lundberg, Harsha Nori, Hamid Palangi, Marco~Tulio Ribeiro, and Yi~Zhang.
\newblock Sparks of artificial general intelligence: Early experiments with gpt-4, 2023.

\bibitem[Chowdhery et~al.(2022)Chowdhery, Narang, Devlin, Bosma, Mishra, Roberts, Barham, Chung, Sutton, Gehrmann, Schuh, Shi, Tsvyashchenko, Maynez, Rao, Barnes, Tay, Shazeer, Prabhakaran, Reif, Du, Hutchinson, Pope, Bradbury, Austin, Isard, Gur-Ari, Yin, Duke, Levskaya, Ghemawat, Dev, Michalewski, Garcia, Misra, Robinson, Fedus, Zhou, Ippolito, Luan, Lim, Zoph, Spiridonov, Sepassi, Dohan, Agrawal, Omernick, Dai, Pillai, Pellat, Lewkowycz, Moreira, Child, Polozov, Lee, Zhou, Wang, Saeta, Diaz, Firat, Catasta, Wei, Meier-Hellstern, Eck, Dean, Petrov, and Fiedel]{chowdhery2022palm}
Aakanksha Chowdhery, Sharan Narang, Jacob Devlin, Maarten Bosma, Gaurav Mishra, Adam Roberts, Paul Barham, Hyung~Won Chung, Charles Sutton, Sebastian Gehrmann, Parker Schuh, Kensen Shi, Sasha Tsvyashchenko, Joshua Maynez, Abhishek Rao, Parker Barnes, Yi~Tay, Noam Shazeer, Vinodkumar Prabhakaran, Emily Reif, Nan Du, Ben Hutchinson, Reiner Pope, James Bradbury, Jacob Austin, Michael Isard, Guy Gur-Ari, Pengcheng Yin, Toju Duke, Anselm Levskaya, Sanjay Ghemawat, Sunipa Dev, Henryk Michalewski, Xavier Garcia, Vedant Misra, Kevin Robinson, Liam Fedus, Denny Zhou, Daphne Ippolito, David Luan, Hyeontaek Lim, Barret Zoph, Alexander Spiridonov, Ryan Sepassi, David Dohan, Shivani Agrawal, Mark Omernick, Andrew~M. Dai, Thanumalayan~Sankaranarayana Pillai, Marie Pellat, Aitor Lewkowycz, Erica Moreira, Rewon Child, Oleksandr Polozov, Katherine Lee, Zongwei Zhou, Xuezhi Wang, Brennan Saeta, Mark Diaz, Orhan Firat, Michele Catasta, Jason Wei, Kathy Meier-Hellstern, Douglas Eck, Jeff Dean, Slav Petrov, and Noah Fiedel.
\newblock Palm: Scaling language modeling with pathways, 2022.

\bibitem[Chung et~al.(2022{\natexlab{a}})Chung, Hou, Longpre, Zoph, Tay, Fedus, Li, Wang, Dehghani, Brahma, Webson, Gu, Dai, Suzgun, Chen, Chowdhery, Castro-Ros, Pellat, Robinson, Valter, Narang, Mishra, Yu, Zhao, Huang, Dai, Yu, Petrov, Chi, Dean, Devlin, Roberts, Zhou, Le, and Wei]{flan-t5}
Hyung~Won Chung, Le~Hou, Shayne Longpre, Barret Zoph, Yi~Tay, William Fedus, Yunxuan Li, Xuezhi Wang, Mostafa Dehghani, Siddhartha Brahma, Albert Webson, Shixiang~Shane Gu, Zhuyun Dai, Mirac Suzgun, Xinyun Chen, Aakanksha Chowdhery, Alex Castro-Ros, Marie Pellat, Kevin Robinson, Dasha Valter, Sharan Narang, Gaurav Mishra, Adams Yu, Vincent Zhao, Yanping Huang, Andrew Dai, Hongkun Yu, Slav Petrov, Ed~H. Chi, Jeff Dean, Jacob Devlin, Adam Roberts, Denny Zhou, Quoc~V. Le, and Jason Wei.
\newblock Scaling instruction-finetuned language models, 2022{\natexlab{a}}.

\bibitem[Chung et~al.(2022{\natexlab{b}})Chung, Hou, Longpre, Zoph, Tay, Fedus, Li, Wang, Dehghani, Brahma, et~al.]{chung2022scaling}
Hyung~Won Chung, Le~Hou, Shayne Longpre, Barret Zoph, Yi~Tay, William Fedus, Yunxuan Li, Xuezhi Wang, Mostafa Dehghani, Siddhartha Brahma, et~al.
\newblock Scaling instruction-finetuned language models.
\newblock \emph{arXiv preprint arXiv:2210.11416}, 2022{\natexlab{b}}.

\bibitem[Devlin et~al.(2019)Devlin, Chang, Lee, and Toutanova]{devlin2019bert}
Jacob Devlin, Ming-Wei Chang, Kenton Lee, and Kristina Toutanova.
\newblock Bert: Pre-training of deep bidirectional transformers for language understanding, 2019.

\bibitem[Do{\u g}an et~al.(2014)Do{\u g}an, Leaman, and Lu]{Dogan2014-aw}
Rezarta~Islamaj Do{\u g}an, Robert Leaman, and Zhiyong Lu.
\newblock {NCBI} disease corpus: A resource for disease name recognition and concept normalization.
\newblock \emph{J. Biomed. Inform.}, 47:\penalty0 1--10, February 2014.

\bibitem[Doğan et~al.(2014)Doğan, Leaman, and Lu]{ncbi}
Rezarta~Islamaj Doğan, Robert Leaman, and Zhiyong Lu.
\newblock Ncbi disease corpus: A resource for disease name recognition and concept normalization.
\newblock \emph{Journal of Biomedical Informatics}, 47:\penalty0 1–10, Feb 2014.
\newblock \doi{10.1016/j.jbi.2013.12.006}.

\bibitem[Fleshman and Durme(2024)]{fleshman2024readapt}
William Fleshman and Benjamin~Van Durme.
\newblock Re-adapt: Reverse engineered adaptation of large language models, 2024.

\bibitem[Gema et~al.(2024)Gema, Minervini, Daines, Hope, and Alex]{gema2024parameterefficientfinetuningllamaclinical}
Aryo~Pradipta Gema, Pasquale Minervini, Luke Daines, Tom Hope, and Beatrice Alex.
\newblock Parameter-efficient fine-tuning of llama for the clinical domain, 2024.
\newblock URL \url{https://arxiv.org/abs/2307.03042}.

\bibitem[Gu et~al.(2021)Gu, Tinn, Cheng, Lucas, Usuyama, Liu, Naumann, Gao, and Poon]{pubmedbert}
Yu~Gu, Robert Tinn, Hao Cheng, Michael Lucas, Naoto Usuyama, Xiaodong Liu, Tristan Naumann, Jianfeng Gao, and Hoifung Poon.
\newblock Domain-specific language model pretraining for biomedical natural language processing.
\newblock \emph{ACM Trans. Comput. Healthcare}, 3\penalty0 (1), oct 2021.
\newblock \doi{10.1145/3458754}.
\newblock URL \url{https://doi.org/10.1145/3458754}.

\bibitem[Gupta et~al.(2023)Gupta, Thérien, Ibrahim, Richter, Anthony, Belilovsky, Rish, and Lesort]{gupta2023warmup}
Kshitij Gupta, Benjamin Thérien, Adam Ibrahim, Mats~L. Richter, Quentin Anthony, Eugene Belilovsky, Irina Rish, and Timothée Lesort.
\newblock Continual pre-training of large language models: How to (re)warm your model?, 2023.

\bibitem[Gururangan et~al.(2020)Gururangan, Marasović, Swayamdipta, Lo, Beltagy, Downey, and Smith]{gururangan2020dont}
Suchin Gururangan, Ana Marasović, Swabha Swayamdipta, Kyle Lo, Iz~Beltagy, Doug Downey, and Noah~A. Smith.
\newblock Don't stop pretraining: Adapt language models to domains and tasks, 2020.

\bibitem[Gutiérrez et~al.(2023)Gutiérrez, Sun, and Su]{gutiérrez2023biomedical}
Bernal~Jiménez Gutiérrez, Huan Sun, and Yu~Su.
\newblock Biomedical language models are robust to sub-optimal tokenization, 2023.

\bibitem[Hager et~al.(2024)Hager, Jungmann, Holland, Bhagat, Hubrecht, Knauer, Vielhauer, Makowski, Braren, Kaissis, and et~al.]{Hager_Jungmann_Holland_Bhagat_Hubrecht_Knauer_Vielhauer_Makowski_Braren_Kaissis_et_al._2024}
Paul Hager, Friederike Jungmann, Robbie Holland, Kunal Bhagat, Inga Hubrecht, Manuel Knauer, Jakob Vielhauer, Marcus Makowski, Rickmer Braren, Georgios Kaissis, and et~al.
\newblock Evaluation and mitigation of the limitations of large language models in clinical decision-making.
\newblock \emph{Nature Medicine}, 30\penalty0 (9):\penalty0 2613–2622, Jul 2024.
\newblock \doi{10.1038/s41591-024-03097-1}.

\bibitem[Harrigian et~al.(2023{\natexlab{a}})Harrigian, Tang, Gonzales, Cai, and Dredze]{harrigian2023eye}
Keith Harrigian, Tina Tang, Anthony Gonzales, Cindy~X. Cai, and Mark Dredze.
\newblock An eye on clinical bert: Investigating language model generalization for diabetic eye disease phenotyping, 2023{\natexlab{a}}.

\bibitem[Harrigian et~al.(2023{\natexlab{b}})Harrigian, Zirikly, Chee, Ahmad, Links, Saha, Beach, and Dredze]{harrigian-etal-2023-characterization}
Keith Harrigian, Ayah Zirikly, Brant Chee, Alya Ahmad, Anne Links, Somnath Saha, Mary~Catherine Beach, and Mark Dredze.
\newblock Characterization of stigmatizing language in medical records.
\newblock In Anna Rogers, Jordan Boyd-Graber, and Naoaki Okazaki, editors, \emph{Proceedings of the 61st Annual Meeting of the Association for Computational Linguistics (Volume 2: Short Papers)}, pages 312--329, Toronto, Canada, July 2023{\natexlab{b}}. Association for Computational Linguistics.
\newblock \doi{10.18653/v1/2023.acl-short.28}.
\newblock URL \url{https://aclanthology.org/2023.acl-short.28}.

\bibitem[Jeong et~al.(2024)Jeong, Jabbour, Yang, Thapta, Mozannar, Han, Mehandru, Wornow, Lialin, Liu, et~al.]{jeong2024recent}
Hyewon Jeong, Sarah Jabbour, Yuzhe Yang, Rahul Thapta, Hussein Mozannar, William~Jongwon Han, Nikita Mehandru, Michael Wornow, Vladislav Lialin, Xin Liu, et~al.
\newblock Recent advances, applications, and open challenges in machine learning for health: Reflections from research roundtables at ml4h 2023 symposium.
\newblock \emph{arXiv preprint arXiv:2403.01628}, 2024.

\bibitem[Johnson et~al.(2023)Johnson, Bulgarelli, Shen, Gayles, Shammout, Horng, Pollard, Hao, Moody, Gow, and et~al.]{mimic4}
Alistair E.~W. Johnson, Lucas Bulgarelli, Lu~Shen, Alvin Gayles, Ayad Shammout, Steven Horng, Tom~J. Pollard, Sicheng Hao, Benjamin Moody, Brian Gow, and et~al.
\newblock Mimic-iv, a freely accessible electronic health record dataset, Jan 2023.
\newblock URL \url{https://www.nature.com/articles/s41597-022-01899-x}.

\bibitem[Johnson et~al.(2016)Johnson, Pollard, Shen, Lehman, Feng, Ghassemi, Moody, Szolovits, Anthony~Celi, and Mark]{mimic3}
Alistair~E.W. Johnson, Tom~J. Pollard, Lu~Shen, Li-wei~H. Lehman, Mengling Feng, Mohammad Ghassemi, Benjamin Moody, Peter Szolovits, Leo Anthony~Celi, and Roger~G. Mark.
\newblock Mimic-iii, a freely accessible critical care database.
\newblock \emph{Scientific Data}, 3\penalty0 (1), May 2016.
\newblock \doi{10.1038/sdata.2016.35}.

\bibitem[Kaplan et~al.(2020)Kaplan, McCandlish, Henighan, Brown, Chess, Child, Gray, Radford, Wu, and Amodei]{kaplan2020scaling}
Jared Kaplan, Sam McCandlish, Tom Henighan, Tom~B Brown, Benjamin Chess, Rewon Child, Scott Gray, Alec Radford, Jeffrey Wu, and Dario Amodei.
\newblock Scaling laws for neural language models.
\newblock \emph{arXiv preprint arXiv:2001.08361}, 2020.

\bibitem[Khanday et~al.(2020)Khanday, Rabani, Khan, Rouf, and Mohi Ud~Din]{khanday2020machine}
Akib Mohi Ud~Din Khanday, Syed~Tanzeel Rabani, Qamar~Rayees Khan, Nusrat Rouf, and Masarat Mohi Ud~Din.
\newblock Machine learning based approaches for detecting covid-19 using clinical text data.
\newblock \emph{International Journal of Information Technology}, 12:\penalty0 731--739, 2020.

\bibitem[Labrak et~al.(2024)Labrak, Bazoge, Morin, Gourraud, Rouvier, and Dufour]{labrak2024biomistralcollectionopensourcepretrained}
Yanis Labrak, Adrien Bazoge, Emmanuel Morin, Pierre-Antoine Gourraud, Mickael Rouvier, and Richard Dufour.
\newblock Biomistral: A collection of open-source pretrained large language models for medical domains, 2024.
\newblock URL \url{https://arxiv.org/abs/2402.10373}.

\bibitem[Lamproudis et~al.(2022)Lamproudis, Henriksson, and Dalianis]{lamproudis_etal_2022_evaluating_bert_pretraining}
Anastasios Lamproudis, Aron Henriksson, and Hercules Dalianis.
\newblock Evaluating pretraining strategies for clinical {BERT} models.
\newblock In Nicoletta Calzolari, Fr{\'e}d{\'e}ric B{\'e}chet, Philippe Blache, Khalid Choukri, Christopher Cieri, Thierry Declerck, Sara Goggi, Hitoshi Isahara, Bente Maegaard, Joseph Mariani, H{\'e}l{\`e}ne Mazo, Jan Odijk, and Stelios Piperidis, editors, \emph{Proceedings of the Thirteenth Language Resources and Evaluation Conference}, pages 410--416, Marseille, France, June 2022. European Language Resources Association.
\newblock URL \url{https://aclanthology.org/2022.lrec-1.43}.

\bibitem[Lee et~al.(2019)Lee, Yoon, Kim, Kim, Kim, So, and Kang]{Lee_2019_biobert}
Jinhyuk Lee, Wonjin Yoon, Sungdong Kim, Donghyeon Kim, Sunkyu Kim, Chan~Ho So, and Jaewoo Kang.
\newblock Biobert: a pre-trained biomedical language representation model for biomedical text mining.
\newblock \emph{Bioinformatics}, 36\penalty0 (4):\penalty0 1234–1240, September 2019.
\newblock ISSN 1367-4811.
\newblock \doi{10.1093/bioinformatics/btz682}.
\newblock URL \url{http://dx.doi.org/10.1093/bioinformatics/btz682}.

\bibitem[Lehman et~al.(2023)Lehman, Hernandez, Mahajan, Wulff, Smith, Ziegler, Nadler, Szolovits, Johnson, and Alsentzer]{Lehman2023DoWS}
Eric~P. Lehman, Evan Hernandez, Diwakar Mahajan, Jonas Wulff, Micah~J. Smith, Zachary~M. Ziegler, Daniel Nadler, Peter Szolovits, Alistair E.~W. Johnson, and Emily Alsentzer.
\newblock Do we still need clinical language models?
\newblock \emph{ArXiv}, abs/2302.08091, 2023.

\bibitem[Lester et~al.(2021)Lester, Al-Rfou, and Constant]{lester-etal-2021-prompt-power}
Brian Lester, Rami Al-Rfou, and Noah Constant.
\newblock The power of scale for parameter-efficient prompt tuning.
\newblock In Marie-Francine Moens, Xuanjing Huang, Lucia Specia, and Scott Wen-tau Yih, editors, \emph{Proceedings of the 2021 Conference on Empirical Methods in Natural Language Processing}, pages 3045--3059, Online and Punta Cana, Dominican Republic, November 2021. Association for Computational Linguistics.
\newblock \doi{10.18653/v1/2021.emnlp-main.243}.
\newblock URL \url{https://aclanthology.org/2021.emnlp-main.243}.

\bibitem[Lewis et~al.(2020{\natexlab{a}})Lewis, Ott, Du, and Stoyanov]{lewis-etal-2020-bioclinroberta}
Patrick Lewis, Myle Ott, Jingfei Du, and Veselin Stoyanov.
\newblock Pretrained language models for biomedical and clinical tasks: Understanding and extending the state-of-the-art.
\newblock In Anna Rumshisky, Kirk Roberts, Steven Bethard, and Tristan Naumann, editors, \emph{Proceedings of the 3rd Clinical Natural Language Processing Workshop}, pages 146--157, Online, November 2020{\natexlab{a}}. Association for Computational Linguistics.
\newblock \doi{10.18653/v1/2020.clinicalnlp-1.17}.
\newblock URL \url{https://aclanthology.org/2020.clinicalnlp-1.17}.

\bibitem[Lewis et~al.(2020{\natexlab{b}})Lewis, Ott, Du, and Stoyanov]{lewis2020pretrained}
Patrick Lewis, Myle Ott, Jingfei Du, and Veselin Stoyanov.
\newblock Pretrained language models for biomedical and clinical tasks: understanding and extending the state-of-the-art.
\newblock In \emph{Proceedings of the 3rd Clinical Natural Language Processing Workshop}, pages 146--157, 2020{\natexlab{b}}.

\bibitem[Li et~al.(2016{\natexlab{a}})Li, Sun, Johnson, Sciaky, Wei, Leaman, Davis, Mattingly, Wiegers, and Lu]{Li2016-ud}
Jiao Li, Yueping Sun, Robin~J Johnson, Daniela Sciaky, Chih-Hsuan Wei, Robert Leaman, Allan~Peter Davis, Carolyn~J Mattingly, Thomas~C Wiegers, and Zhiyong Lu.
\newblock {BioCreative} {V} {CDR} task corpus: a resource for chemical disease relation extraction.
\newblock \emph{Database (Oxford)}, 2016:\penalty0 baw068, May 2016{\natexlab{a}}.

\bibitem[Li et~al.(2016{\natexlab{b}})Li, Sun, Johnson, Sciaky, Wei, Leaman, Davis, Mattingly, Wiegers, and Lu]{bc5cdr}
Jiao Li, Yueping Sun, Robin~J. Johnson, Daniela Sciaky, Chih-Hsuan Wei, Robert Leaman, Allan~Peter Davis, Carolyn~J. Mattingly, Thomas~C. Wiegers, and Zhiyong Lu.
\newblock Biocreative v cdr task corpus: A resource for chemical disease relation extraction.
\newblock \emph{Database}, 2016, 2016{\natexlab{b}}.
\newblock \doi{10.1093/database/baw068}.

\bibitem[Lin et~al.(2022)Lin, He, Ze, Wang, Hua, Dupuy, Gupta, Soltanolkotabi, Ren, and Avestimehr]{lin-etal-2022-fednlp}
Bill~Yuchen Lin, Chaoyang He, Zihang Ze, Hulin Wang, Yufen Hua, Christophe Dupuy, Rahul Gupta, Mahdi Soltanolkotabi, Xiang Ren, and Salman Avestimehr.
\newblock {F}ed{NLP}: Benchmarking federated learning methods for natural language processing tasks.
\newblock In Marine Carpuat, Marie-Catherine de~Marneffe, and Ivan~Vladimir Meza~Ruiz, editors, \emph{Findings of the Association for Computational Linguistics: NAACL 2022}, pages 157--175, Seattle, United States, July 2022. Association for Computational Linguistics.
\newblock \doi{10.18653/v1/2022.findings-naacl.13}.
\newblock URL \url{https://aclanthology.org/2022.findings-naacl.13}.

\bibitem[Longpre et~al.(2023)Longpre, Yauney, Reif, Lee, Roberts, Zoph, Zhou, Wei, Robinson, Mimno, and Ippolito]{longpre2023pretrainers_training}
Shayne Longpre, Gregory Yauney, Emily Reif, Katherine Lee, Adam Roberts, Barret Zoph, Denny Zhou, Jason Wei, Kevin Robinson, David Mimno, and Daphne Ippolito.
\newblock A pretrainer's guide to training data: Measuring the effects of data age, domain coverage, quality, \& toxicity, 2023.

\bibitem[Lu et~al.(2022)Lu, Dou, and Nguyen]{luClinicalT5}
Qiuhao Lu, Dejing Dou, and Thien Nguyen.
\newblock {C}linical{T}5: A generative language model for clinical text.
\newblock In \emph{Findings of the Association for Computational Linguistics: EMNLP 2022}, pages 5436--5443, Abu Dhabi, United Arab Emirates, December 2022. Association for Computational Linguistics.
\newblock URL \url{https://aclanthology.org/2022.findings-emnlp.398}.

\bibitem[Mehta et~al.(2023)Mehta, Patil, Chandar, and Strubell]{mehta2023empirical}
Sanket~Vaibhav Mehta, Darshan Patil, Sarath Chandar, and Emma Strubell.
\newblock An empirical investigation of the role of pre-training in lifelong learning, 2023.

\bibitem[Mueller et~al.(2022)Mueller, Andrews, and Dredze]{mueller2022text}
David Mueller, Nicholas Andrews, and Mark Dredze.
\newblock Do text-to-text multi-task learners suffer from task conflict?
\newblock \emph{arXiv preprint arXiv:2212.06645}, 2022.

\bibitem[Mullenbach et~al.(2021)Mullenbach, Pruksachatkun, Adler, Seale, Swartz, McKelvey, Dai, Yang, and Sontag]{CLIP}
James Mullenbach, Yada Pruksachatkun, Sean Adler, Jennifer Seale, Jordan Swartz, T.~Greg McKelvey, Hui Dai, Yi~Yang, and David~A. Sontag.
\newblock {CLIP:} {A} dataset for extracting action items for physicians from hospital discharge notes.
\newblock \emph{CoRR}, abs/2106.02524, 2021.
\newblock URL \url{https://arxiv.org/abs/2106.02524}.

\bibitem[Nakayama(2018)]{seqeval}
Hiroki Nakayama.
\newblock {seqeval}: A python framework for sequence labeling evaluation, 2018.
\newblock URL \url{https://github.com/chakki-works/seqeval}.
\newblock Software available from https://github.com/chakki-works/seqeval.

\bibitem[Ness et~al.(2024)Ness, Matton, Helm, Zhang, Bajwa, Priebe, and Horvitz]{ness2024medfuzzexploringrobustnesslarge}
Robert~Osazuwa Ness, Katie Matton, Hayden Helm, Sheng Zhang, Junaid Bajwa, Carey~E. Priebe, and Eric Horvitz.
\newblock Medfuzz: Exploring the robustness of large language models in medical question answering, 2024.
\newblock URL \url{https://arxiv.org/abs/2406.06573}.

\bibitem[{Phan} et~al.(2021){Phan}, {Anibal}, {Tran}, {Chanana}, {Bahadroglu}, {Peltekian}, and {Altan-Bonnet}]{phan2021SciFive}
Long~N. {Phan}, James~T. {Anibal}, Hieu {Tran}, Shaurya {Chanana}, Erol {Bahadroglu}, Alec {Peltekian}, and Gr{\'e}goire {Altan-Bonnet}.
\newblock {SciFive: a text-to-text transformer model for biomedical literature}.
\newblock \emph{arXiv e-prints}, art. arXiv:2106.03598, May 2021.
\newblock \doi{10.48550/arXiv.2106.03598}.

\bibitem[Radford et~al.(2019)Radford, Wu, Child, Luan, Amodei, and Sutskever]{Radford2019LanguageMA}
Alec Radford, Jeff Wu, Rewon Child, David Luan, Dario Amodei, and Ilya Sutskever.
\newblock Language models are unsupervised multitask learners.
\newblock 2019.
\newblock URL \url{https://api.semanticscholar.org/CorpusID:160025533}.

\bibitem[Raffel et~al.(2023)Raffel, Shazeer, Roberts, Lee, Narang, Matena, Zhou, Li, and Liu]{t5}
Colin Raffel, Noam Shazeer, Adam Roberts, Katherine Lee, Sharan Narang, Michael Matena, Yanqi Zhou, Wei Li, and Peter~J. Liu.
\newblock Exploring the limits of transfer learning with a unified text-to-text transformer, 2023.

\bibitem[Rajpurkar et~al.(2018)Rajpurkar, Zhang, and Liang]{Rajpurkar2018SQuAD2}
Pranav Rajpurkar, Jian Zhang, and Percy Liang.
\newblock Know what you don't know: Unanswerable questions for squad.
\newblock In \emph{ACL 2018}, 2018.

\bibitem[Ramshaw and Marcus(1995)]{seqeval-1995-text}
Lance Ramshaw and Mitch Marcus.
\newblock Text chunking using transformation-based learning.
\newblock In \emph{Third Workshop on Very Large Corpora}, 1995.
\newblock URL \url{https://www.aclweb.org/anthology/W95-0107}.

\bibitem[Romanov and Shivade(2018)]{romanov2018mednli}
Alexey Romanov and Chaitanya Shivade.
\newblock Lessons from natural language inference in the clinical domain, 2018.

\bibitem[Shazeer and Stern(2018)]{shazeer2018adafactor}
Noam Shazeer and Mitchell Stern.
\newblock Adafactor: Adaptive learning rates with sublinear memory cost, 2018.

\bibitem[Singhal et~al.(2022)Singhal, Azizi, Tu, Mahdavi, Wei, Chung, Scales, Tanwani, Cole-Lewis, Pfohl, Payne, Seneviratne, Gamble, Kelly, Scharli, Chowdhery, Mansfield, y~Arcas, Webster, Corrado, Matias, Chou, Gottweis, Tomasev, Liu, Rajkomar, Barral, Semturs, Karthikesalingam, and Natarajan]{singhal2022medpalm}
Karan Singhal, Shekoofeh Azizi, Tao Tu, S.~Sara Mahdavi, Jason Wei, Hyung~Won Chung, Nathan Scales, Ajay Tanwani, Heather Cole-Lewis, Stephen Pfohl, Perry Payne, Martin Seneviratne, Paul Gamble, Chris Kelly, Nathaneal Scharli, Aakanksha Chowdhery, Philip Mansfield, Blaise~Aguera y~Arcas, Dale Webster, Greg~S. Corrado, Yossi Matias, Katherine Chou, Juraj Gottweis, Nenad Tomasev, Yun Liu, Alvin Rajkomar, Joelle Barral, Christopher Semturs, Alan Karthikesalingam, and Vivek Natarajan.
\newblock Large language models encode clinical knowledge, 2022.

\bibitem[Singhal et~al.(2023)Singhal, Azizi, Tu, Mahdavi, Wei, Chung, Scales, Tanwani, Cole-Lewis, Pfohl, Payne, Seneviratne, Gamble, Kelly, Babiker, Sch{\"a}rli, Chowdhery, Mansfield, Demner-Fushman, Ag{\"u}era Y~Arcas, Webster, Corrado, Matias, Chou, Gottweis, Tomasev, Liu, Rajkomar, Barral, Semturs, Karthikesalingam, and Natarajan]{Singhal2023-zo}
Karan Singhal, Shekoofeh Azizi, Tao Tu, S~Sara Mahdavi, Jason Wei, Hyung~Won Chung, Nathan Scales, Ajay Tanwani, Heather Cole-Lewis, Stephen Pfohl, Perry Payne, Martin Seneviratne, Paul Gamble, Chris Kelly, Abubakr Babiker, Nathanael Sch{\"a}rli, Aakanksha Chowdhery, Philip Mansfield, Dina Demner-Fushman, Blaise Ag{\"u}era Y~Arcas, Dale Webster, Greg~S Corrado, Yossi Matias, Katherine Chou, Juraj Gottweis, Nenad Tomasev, Yun Liu, Alvin Rajkomar, Joelle Barral, Christopher Semturs, Alan Karthikesalingam, and Vivek Natarajan.
\newblock Large language models encode clinical knowledge.
\newblock \emph{Nature}, 620\penalty0 (7972):\penalty0 172--180, August 2023.

\bibitem[Soni et~al.(2022)Soni, Gudala, Pajouhi, and Roberts]{soni-etal-2022-radqa}
Sarvesh Soni, Meghana Gudala, Atieh Pajouhi, and Kirk Roberts.
\newblock {R}ad{QA}: A question answering dataset to improve comprehension of radiology reports.
\newblock In \emph{Proceedings of the Thirteenth Language Resources and Evaluation Conference}, pages 6250--6259, Marseille, France, June 2022. European Language Resources Association.
\newblock URL \url{https://aclanthology.org/2022.lrec-1.672}.

\bibitem[Spasic and Nenadic(2020)]{clinical_text_review}
Irena Spasic and Goran Nenadic.
\newblock Clinical text data in machine learning: Systematic review.
\newblock JMIR Medical Informatics, 03 2020.
\newblock \doi{10.2196/17984}.

\bibitem[Tang et~al.(2023)Tang, Sun, Idnay, Nestor, Soroush, Elias, Xu, Ding, Durrett, Rousseau, and et~al.]{Tang_Sun_et_al._2023_medical_evidence_summarization}
Liyan Tang, Zhaoyi Sun, Betina Idnay, Jordan~G. Nestor, Ali Soroush, Pierre~A. Elias, Ziyang Xu, Ying Ding, Greg Durrett, Justin~F. Rousseau, and et~al.
\newblock Evaluating large language models on medical evidence summarization.
\newblock \emph{npj Digital Medicine}, 6\penalty0 (1), Aug 2023.
\newblock \doi{10.1038/s41746-023-00896-7}.

\bibitem[Taylor et~al.(2022)Taylor, Kardas, Cucurull, Scialom, Hartshorn, Saravia, Poulton, Kerkez, and Stojnic]{taylor2022galactica}
Ross Taylor, Marcin Kardas, Guillem Cucurull, Thomas Scialom, Anthony Hartshorn, Elvis Saravia, Andrew Poulton, Viktor Kerkez, and Robert Stojnic.
\newblock Galactica: A large language model for science, 2022.

\bibitem[Toma et~al.(2023)Toma, Lawler, Ba, Krishnan, Rubin, and Wang]{toma2023clinicalcamelopenexpertlevel}
Augustin Toma, Patrick~R. Lawler, Jimmy Ba, Rahul~G. Krishnan, Barry~B. Rubin, and Bo~Wang.
\newblock Clinical camel: An open expert-level medical language model with dialogue-based knowledge encoding, 2023.
\newblock URL \url{https://arxiv.org/abs/2305.12031}.

\bibitem[Touvron et~al.(2023)Touvron, Lavril, Izacard, Martinet, Lachaux, Lacroix, Rozière, Goyal, Hambro, Azhar, Rodriguez, Joulin, Grave, and Lample]{touvron2023llama}
Hugo Touvron, Thibaut Lavril, Gautier Izacard, Xavier Martinet, Marie-Anne Lachaux, Timothée Lacroix, Baptiste Rozière, Naman Goyal, Eric Hambro, Faisal Azhar, Aurelien Rodriguez, Armand Joulin, Edouard Grave, and Guillaume Lample.
\newblock Llama: Open and efficient foundation language models, 2023.

\bibitem[Vaswani et~al.(2017)Vaswani, Shazeer, Parmar, Uszkoreit, Jones, Gomez, Kaiser, and Polosukhin]{vaswani2017attention}
Ashish Vaswani, Noam Shazeer, Niki Parmar, Jakob Uszkoreit, Llion Jones, Aidan~N. Gomez, Lukasz Kaiser, and Illia Polosukhin.
\newblock Attention is all you need, 2017.

\bibitem[Wu et~al.(2023{\natexlab{a}})Wu, Zhang, Zhang, Wang, and Xie]{wu2023pmcllama}
Chaoyi Wu, Xiaoman Zhang, Ya~Zhang, Yanfeng Wang, and Weidi Xie.
\newblock Pmc-llama: Further finetuning llama on medical papers, 2023{\natexlab{a}}.

\bibitem[Wu et~al.(2023{\natexlab{b}})Wu, Irsoy, Lu, Dabravolski, Dredze, Gehrmann, Kambadur, Rosenberg, and Mann]{wu2023bloomberggpt}
Shijie Wu, Ozan Irsoy, Steven Lu, Vadim Dabravolski, Mark Dredze, Sebastian Gehrmann, Prabhanjan Kambadur, David Rosenberg, and Gideon Mann.
\newblock Bloomberggpt: A large language model for finance, 2023{\natexlab{b}}.

\bibitem[Yang et~al.(2022)Yang, Chen, PourNejatian, Shin, Smith, Parisien, Compas, Martin, Flores, Zhang, Magoc, Harle, Lipori, Mitchell, Hogan, Shenkman, Bian, and Wu]{yang2022gatortron}
Xi~Yang, Aokun Chen, Nima PourNejatian, Hoo~Chang Shin, Kaleb~E Smith, Christopher Parisien, Colin Compas, Cheryl Martin, Mona~G Flores, Ying Zhang, Tanja Magoc, Christopher~A Harle, Gloria Lipori, Duane~A Mitchell, William~R Hogan, Elizabeth~A Shenkman, Jiang Bian, and Yonghui Wu.
\newblock Gatortron: A large clinical language model to unlock patient information from unstructured electronic health records, 2022.

\bibitem[Yasunaga et~al.(2022)Yasunaga, Leskovec, and Liang]{yasunaga2022linkbert}
Michihiro Yasunaga, Jure Leskovec, and Percy Liang.
\newblock Linkbert: Pretraining language models with document links, 2022.

\bibitem[Yue et~al.(2020)Yue, Jimenez~Gutierrez, and Sun]{yue-etal-2020-analysis-emrqa}
Xiang Yue, Bernal Jimenez~Gutierrez, and Huan Sun.
\newblock Clinical reading comprehension: A thorough analysis of the emr{QA} dataset.
\newblock In Dan Jurafsky, Joyce Chai, Natalie Schluter, and Joel Tetreault, editors, \emph{Proceedings of the 58th Annual Meeting of the Association for Computational Linguistics}, pages 4474--4486, Online, July 2020. Association for Computational Linguistics.
\newblock \doi{10.18653/v1/2020.acl-main.410}.
\newblock URL \url{https://aclanthology.org/2020.acl-main.410}.

\bibitem[Zhou et~al.(2023)Zhou, Yetisgen, Afshar, Gao, Savova, and Miller]{jamia_clinical_continued_pretraining}
Weipeng Zhou, Meliha Yetisgen, Majid Afshar, Yanjun Gao, Guergana Savova, and Timothy~A Miller.
\newblock {Improving model transferability for clinical note section classification models using continued pretraining}.
\newblock \emph{Journal of the American Medical Informatics Association}, 31\penalty0 (1):\penalty0 89--97, 09 2023.
\newblock ISSN 1527-974X.
\newblock \doi{10.1093/jamia/ocad190}.
\newblock URL \url{https://doi.org/10.1093/jamia/ocad190}.

\end{thebibliography}

\appendix

\section{Dataset Details}
\label{sec:appendix_dataset_details}

We adapted the data processing methods mostly from the original {MIMIC-T5}, {SciFive+MIMIC-T5} and {SciFive}'s paper and codebases.

\paragraph{MedNLI} MedNLI \citep{romanov2018mednli} is a natural language inference dataset derived from MIMIC-III \citep{mimic3}. Given two sentences - premise and hypothesis - it labels the relation between these two sentences as ``entailment", ``natural", and ``contradiction." 

\paragraph{RadQA} RadQA \citep{soni-etal-2022-radqa}
 is a question-answering dataset sourced from the radiology report in MIMIC-III \citep{mimic3}. It annotates the answer span to the questions; if no answers for the question, the answer text is empty. 

\paragraph{CLIP} CLIP \citep{CLIP} is a multi-label classification dataset of clinical action items, such as ``patient instructions" and ``appointment." Due to the long context of clinical records of these two tasks, we referred to the methodology utilized by {MIMIC-T5}, which involves segmenting long records and appropriately mapping labels to the segmented sequences. 

\paragraph{BC5CDR-disease} BC5CDR-disease \citep{bc5cdr} is a Disease Named Entity Recognition corpus, which annotates 1500 PubMed articles with disease entities labels. Suppose $(x, y)$ is the input-output pair: $y$ is the same as $x$ except that disease entities are enclosed with  ``disease*'' and ``*disease''.

\paragraph{NCBI-disease} NCBI-disease \citep{ncbi} is also a corpus for Disease Named Entity Recoginition from PubMed abtracts. Both BC5CDR-disease and NCBI-disease are preprocessed by {SciFive} and available in their repo.

\paragraph{HOC} HOC dataset \citep{Baker2015} is accessible on Huggingface. It is developed from PubMed publication abstracts. We use the sentence-level dataset, where each sentence is annotated for different hallmarks of cancer. 

\paragraph{Clinical Stigmatizing Language Datasets} Clinical Stigmatizing Language Datasets \citep{harrigian-etal-2023-characterization} characterizes stigmatizing languages in three classification tasks: 

1. Credibility \& Obstinacy (Disbelief, Difficult,
Exclude): expressions of doubt or resistance. 

2. Compliance (Negative, Neutral, Positive): adherence to medical advice.

3. Descriptors (Negative, Neutral, Positive, Exclude): characterization of patient behavior and demeanor.

\section{Prefix and Instructions}
\label{sec:appendix_prefix_instruction}

\paragraph{T5 Prefix}We used the same prefix as MIMIC-T5 on MedNLI, RadQA, and CLIP datasets. We referred to SCFIVE-T5's github repository for NCBI, BC5CDR, NCBI's prefix. On the stigmatizing language datasets, we referred to its keyword category as prefix before inputs: ``adamant" for the Credibility \& Obstinacy task; ``compliance'' for the Compliance task;       ``other" for the Descriptors task. 

\paragraph{FLAN-T5 Instructions} We didn't exhaustively experiment with different instructions for FLAN-T5. We referred to some instructions from MIMIC-T5 and most instructions are straight-forward.

\begin{itemize}

\item MedNLI
 \texttt{Answer entailment, contradiction or neutral. Premise: \{premise\} Hypothesis:\{hypothesis\}}

\item RadQA \texttt{Context: \{context\} Question: \{question\} If no answer is found in the context, do not reply; otherwise, give an answer from the context: }

\item CLIP \texttt{ Context: \{context\}. Label the above sentence as an empty string or as one or more of the following options, delimited by comma: 
Options: \{labels\}
}

\item HOC \texttt{
Sentence: \{input\} Assign the above sentence as zero or more of the following class labels: \{labels\}
}

\item BC5CR-disease \& NCBI-disease \texttt{
Sentence: \{input\} Identify and label disease terms in the sentence:
}

\item Stigmatizing Language Dataset 
\begin{itemize}
    \item Credibility \& Obstinacy
    
    \texttt{Classify this sentence as difficult, disbelief, or exclude, regarding the credibility and obstinacy of the patient: \{input\}}
    \item Compliance
    
    \texttt{Classify this sentence as negative, neutral, or positive, regarding the patient's compliance with medical advice: \{input\} }
    \item Descriptors
    
    \texttt{Classify this sentence as exclude, negative, neutral, or positive, regarding the patient's behavior and demeanor: \{input\} } 
\end{itemize}

\end{itemize}

\section{Model Details}
\label{sec:appendix_model_details}

The details of the pre-training corpus for each model are listed in \tableref{model-settings}.

\begin{table*}
        \centering
        \begin{tabular}{ccc}
        \toprule
             & \textbf{Weight Initialization} &  \textbf{pre-training Data}  \\
            \midrule
    \begin{tabular}{c}
         {T5-Sup} \\
         \citep{t5}
    \end{tabular} & Random & 
    \begin{tabular}{c}
        Colossal Clean
Crawled Corpus (C4)  \\
      +  Supervised Tasks \citep{t5}
    \end{tabular}  \\
\midrule

\begin{tabular}{c}
    {T5-Den} \\
     \citep{t5}
\end{tabular}  &  Random & 
\begin{tabular}{c}
    Colossal Clean
Crawled Corpus (C4) \\
\citep{t5} 
\end{tabular}\\

\midrule
            
            \begin{tabular}{c}
                {Clinical-T5} \\
                 \citep{Lehman2023DoWS} 
            \end{tabular} &  Random  &  
            \begin{tabular}{c}
          
            MIMIC-III \citep{mimic3} \\
            \&  MIMIC-IV \citep{mimic4}  \\
            \end{tabular}
            
            \\
            \midrule

            \begin{tabular}{c}
                 {Clinical-T5}  \\
                 \citep{luClinicalT5} 
            \end{tabular}
             &   
             \begin{tabular}{c}
                  SciFive-PubMed-PMC \\
                  \citep{phan2021SciFive}
             \end{tabular}
                  
      & \begin{tabular}{c}
        MIMIC-III \\
           \citep{mimic3}
      \end{tabular}  \\
           \bottomrule
        \end{tabular}
        \caption{\label{model-settings}
        pre-training corpus for the models we compared. All models compared have 770 million parameters. 
}
\end{table*}

\section{Training Details}
\label{sec:appendix_training_details}

\subsection{Training settings}

\paragraph{Training on Full Datasets}
We use code provided by prior work  \citep{phan2021SciFive, Lehman2023DoWS} to fine-tune the various T5 variants with pytorch==2.1.2. Throughout our experiments, we used the adafactor optimizer \citep{shazeer2018adafactor} with constant $lr=1e-4$, batch size of 64, and set max sequence lengths that depend on the dataset. For all fine-tuning runs, we train for a fixed amount of epochs (30) and pick the best checkpoint based on validation set performance. We chose accuracy for MedNLI, and F-1 for other datasets to pick the best checkpoint during the validation and then evaluated on the complete test split per model and dataset.

\paragraph{Training on Downsampled Datasets}
Due to variations in the downsampled training data, we conducted a learning rate search for all models in the range of $[1e-3, 5e-4, 1e-4, 1e-5]$ using a batch size of 16 with 5\% and 25\% downsampling training data and using a batch size of 4 with 1\% downsampling training data on stigmatizing language datasets and ran the learning rate search in the range of $[5e-3, 1e-3, 5e-4, 1e-4]$ with a batch size of 64 on MedNLI dataset. We trained each model for 30 epochs. We selected the best checkpoint for each model based on their average validation score across folds (i.e. macro-F1 for the stigmatizing task and accuracy for the MedNLI task) and subsequently evaluated on the test dataset. The best-found learning rates are reported in \tableref{tab: Best_lr_mimic_downsample}, \tableref{tab: Best_lr_internal_downsample}, and \tableref{tab: Best_lr_mednli}.

\begin{table}[]
    \centering
        \adjustbox{width=\linewidth}{
    \begin{tabular}{ccc}   
    \toprule
     1\% Training data \\ 
     \midrule 
      Model  & Learning Rate & Batch Size \\
      \hdashline
       T5-Sup  & 1e-4 & 4 \\
       FLAN-T5 & 1e-4 & 4\\
       MIMIC-T5 & 5e-4 & 4\\
       SCIFIVE+MIMIC-T5 & 1e-4 & 4 \\
    \bottomrule
    \end{tabular}}
    
    \medskip
    
    \adjustbox{width=\linewidth}{
    \begin{tabular}{ccc}   
    \toprule
     5\% Training data \\ 
     \midrule 
      Model  & Learning Rate & Batch Size \\
      \hdashline
       T5-Sup  & 1e-4  & 16  \\
       FLAN-T5 & 1e-4 & 16 \\
       MIMIC-T5 & 5e-4 & 16 \\
       SCIFIVE+MIMIC-T5 & 5e-4 & 16 \\
    \bottomrule
    \end{tabular}}
    
    \medskip
    
    \adjustbox{width=\linewidth}{
    \begin{tabular}{ccc}   
    \toprule
     25\% Training data \\ 
     \midrule 
      Model  & Learning Rate & Batch Size \\
      \hdashline
       T5-Sup  & 1e-4 & 16 \\
       FLAN-T5 & 1e-4 & 16 \\
       MIMIC-T5 & 1e-4 & 16 \\
       SCIFIVE+MIMIC-T5 & 1e-4 & 16 \\
    \bottomrule
    \end{tabular}}
    \caption{Best-found hyperparameter settings for down-sampling experiments on stigmatizing language dataset - MIMIC-IV}
    \label{tab: Best_lr_mimic_downsample}
\end{table}

\begin{table}[]
    \centering
    \adjustbox{width=\linewidth}{
    \begin{tabular}{ccc}   
    \toprule
     1\% Training data \\ 
     \midrule 
      Model  & Learning Rate & Batch Size \\
      \hdashline
       T5-Sup  & 1e-3 &  4 \\
       FLAN-T5 & 1e-4 & 4 \\
       MIMIC-T5 & 1e-3 & 4 \\
       SCIFIVE+MIMIC-T5 & 5e-4 & 4 \\
    \bottomrule
    \end{tabular}}

    \medskip
     \adjustbox{width=\linewidth}{
    \begin{tabular}{ccc}   
    \toprule
     5\% Training data \\ 
     \midrule 
      Model  & Learning Rate & Batch Size \\
      \hdashline
       T5-Sup  & 1e-4 & 16 \\
       FLAN-T5 & 1e-4 & 16 \\
       MIMIC-T5 & 1e-3 & 16 \\
       SCIFIVE+MIMIC-T5 & 5e-4 & 16 \\
    \bottomrule
    \end{tabular}}

    \medskip
        \adjustbox{width=\linewidth}{
    \begin{tabular}{ccc}   
    \toprule
     25 \% Training data \\ 
     \midrule 
      Model  & Learning Rate & Batch Size \\
      \hdashline
       T5-Sup  & 1e-4 & 16 \\
       FLAN-T5 & 1e-4 & 16 \\
       MIMIC-T5 & 5e-4 & 16 \\
       SCIFIVE+MIMIC-T5 & 1e-4 & 16 \\
    \bottomrule
    \end{tabular}}
    
    \caption{Best-found hyperparameter settings for down-sampling experiments on stigmatizing language dataset - Hospital System}
    \label{tab: Best_lr_internal_downsample}
\end{table}

\begin{table}[]
    \centering
    \adjustbox{width=\linewidth}{
    \begin{tabular}{ccc}
    \toprule
    1 \% Training Data \\
    \midrule
      Model  & Learning Rate & Batch Size \\
      \hdashline
       T5-Sup  & 1e-4  &  64 \\
       FLAN-T5 & 1e-4 & 64 \\
       MIMIC-T5 & 1e-3 & 64 \\
       SCIFIVE+MIMIC-T5 &  1e-4 & 64 \\
    \bottomrule
    \end{tabular}}
    \caption{Best-found hyperparameter settings for down-sampling experiments on MedNLI dataset with 1\% training data}
    \label{tab: Best_lr_mednli}
\end{table}

\subsection{Differences from Prior Work}

We summarize a list of changes we made that possibly contribute to the difference in results compared to prior {Clinical-T5}s works \citep{luClinicalT5, Lehman2023DoWS}:
\begin{itemize}
    \item We trained each model on each supervised dataset from scratch. 
    \item We implemented the 5-fold cross-validation experiment, focusing on variance in data. This is a different evaluation structure from prior works. 
    \item We added {T5-Den} and FLAN-T5 in addition to widely used {T5-Sup} into the comparison.
\end{itemize}

\subsection{Computing Resources}

We conducted our experiments on multiple 40G/80G A100 GPUs and Tesla M60 GPUs.  Each run of the fine-tuning takes a few hours, except for CLIP dataset which could take approximately 10 hours for 30 epochs with multiple GPUs. However, it should be noted that these models are approaching a size that is not necessarily easy to accommodate in clinical settings.

\section{Evaluation Details}

We use the Python 3.9 packages sklearn==1.1.2, and evaluate==0.3.0 for evaluations.

\paragraph{Classification} For the CLIP dataset, we followed the same evaluation strategy as \citealp{Lehman2023DoWS} by transforming predicted values into binary matrices, whose dimensions indicate the presence of class labels. And then used the package sklearn.metrics to calculate macro-F1 and micro-F1. On HOC dataset, we evaluate the outputs in the same manner as \citealp{phan2021SciFive} and \citealp{yasunaga2022linkbert}. For stigmatizing datasets, we also used sklearn.metrics for macro-F1 scores.

\paragraph{Natural Language Inference} For the MedNLI dataset, without additionally post-processing the outputs, we compute the F1 score and the accuracy directly between the labels and predicted values. 

\paragraph{Question Answering} We followed the same processing method as \citealp{Lehman2023DoWS} utilizing SQuAD 2.0 \citep{Rajpurkar2018SQuAD2}. 

\paragraph{Named Entity Recognition} As instructed by Clinical-T5 \citep{luClinicalT5}, we referred to the evaluation of SciFive  \citep{phan2021SciFive}. BC5CDR-disease and NCBI-disease datasets label the entity by inserting "disease*" and "*disease" around disease names in the input. We convert generated sentence outputs into sequences of "B-disease", "I-disease", "O", and padding and then evaluate using the metrics seqeval \citep{seqeval-1995-text, seqeval}.

\label{sec:appendix_evaluation}

\section{Reproducibility Experiments}
\label{sec:reproducibility_experiments}

We evaluated {T5}, {MIMIC-T5}, and {SciFive+MIMIC-T5} on the six datasets, following the original methodologies (i.e. a single train/dev/test split) proposed in each respective paper. Results of three clinical datasets are shown in \tableref{replicate_clinical_results} and results of three biomedical datasets are shown in \tableref{replicate_biomedical_results}. And these results are single-run.

\section{Cross-Validation Experiments}

We did 5-fold cross-validation in the following procedures: we merged the training and validation datasets and then shuffled them. Within each fold, 4 subsets are for training and one subset is for validation. For down-sampling experiments, we down-sampled the training data only while keeping the validation the same per fold. In the end, all the models were evaluated on the original test dataset. The detailed performances cross models are listed in \appendixref{sec:appendix_cross_validation_results} with paired T-test results.

Specifically, \ref{mednli_large_cv} shows 5-fold cross validation results on MedNLI Dataset. 
\tableref{tab:radqa_large_cv} shows 5-fold cross validation results on RadQA Dataset. 
\tableref{tab:clip_large_cv} shows 5-fold cross validation results on CLIP dataset.  \tableref{tab:hoc_large_Cv} shows 5-fold cross validation results on HOC dataset. 
\tableref{tab:bc5cdr_large_cv} shows 5-fold cross validation results on BC5CDR-disease dataset.
\tableref{tab:ncbi_large_cv} shows 5-fold cross validation results on NCBI-disease dataset. \tableref{tab:mimic_stigma_cv} and \tableref{tab:internal_stigma_cv} are results for the stigmatizing language datasets from MIMIC-IV and Hospital System separately. \tableref{tab:mimic_25_cv}, \tableref{tab:mimic_5_cv}
, and \tableref{tab:mimic_1_cv} illustrate the outcomes of down-sampling experiments on the MIMIC-IV dataset; \tableref{tab:internal_25_cv}, \tableref{tab:internal_5_cv}, and \tableref{tab:internal_1_cv} present the outcomes of down-sampling experiments on the Hospital System dataset.

\begin{table*}
    \centering
    \begin{tabular}{ccccc}
    \toprule
        Task & Dataset & Source  & Evaluated By & Metrics \\
        \midrule
        \textbf{NLI}  & MedNLI & MIMIC  & Both & Accuracy \\
        \midrule
        \textbf{Classif.} & CLIP & MIMIC & {MIMIC-T5} & F1(Macro, Micro) \\
        & HOC & Biomedical  & {SciFive+MIMIC-T5} & F1, P, R \\
        \midrule
        \textbf{QA}  & RadQA & MIMIC & {MIMIC-T5}& EM, F1 \\
        \midrule
        \textbf{NER}  & \begin{tabular}{c}
             BC5CDR  \\
             NCBI
        \end{tabular} & Biomedical & {SciFive+MIMIC-T5} & 
             F1, P, R \\
    \bottomrule
    \end{tabular}

    \caption{Statistics of datasets used in {MIMIC-T5} and {SciFive+MIMIC-T5}}
\label{tab:dataset-statistics}

\end{table*}

\begin{table*}
    \centering 
\begin{tabular}{cccccc}
\toprule
                       & MedNLI   & \multicolumn{2}{c}{RadQA} & \multicolumn{2}{c}{CLIP} \\
                       \cmidrule(lr){2-2}
                       \cmidrule(lr){3-4}
                       \cmidrule(lr){5-6}
                       & Accuracy & EM         & F1     & Macro-F1    & Micro-F1   \\
                       \midrule

{T5-Den}               & 85.16    & 52.44      & 71.42        & 64.68       & 79.11      \\

\hdashline

{T5-Sup}               & 84.95 & 54.07   & 71.14   & 64.20  & 78.93   \\

\midrule
{MIMIC-T5} & ~   & ~   & ~   & ~ & ~   \\
 \hfill --- Ours & 86.64    &  54.72     & 73.38        & 65.76       & 80.11      \\

\hfill --- Reported &  87.2     &  55.0     & 74.5  & 66.3      & 80.0        \\
\midrule 
{SciFive+MIMIC-T5}  & ~   & ~   & ~   & ~ & ~   \\
\hfill --- Ours&   85.79       &      52.28      &      70.74        & 63.79       & 78.24     \\
\hfill --- Reported & 85.86 & N/A & N/A & N/A & N/A \\

\bottomrule
\end{tabular} 
\caption{We reproduced the results of {MIMIC-T5} on three clinical datasets as reported and complemented evaluations of {SciFive+MIMIC-T5} on the clinical datasets that were not used previously. }
\label{replicate_clinical_results}
\end{table*}

\begin{table*}
\adjustbox{max width=\linewidth}{%
    \begin{tabular}{cccccccccc}
    \toprule
                       & \multicolumn{3}{c}{HOC}    & \multicolumn{3}{c}{BC5CDR} & \multicolumn{3}{c}{NCBI}   \\
                       \cmidrule(lr){2-4}
                       \cmidrule(lr){5-7}
                       \cmidrule(lr){8-10}
                       & F1 & P & R  & F1 & P & R & F1    & P & R \\
                       \midrule
{T5-Den}     & 85.57 & 85.81     & 85.34  & 80.43 & {81.74}     & 79.17  & 85.29 & 82.86     & 87.88  \\
\hdashline
{T5-Sup}        & {85.91} & {86.01} & 85.80  & 81.70 & 81.46 & 81.94 & 82.53 & 82.49   & 82.58 \\ \midrule

{MIMIC-T5} & ~ &~     & ~  & ~ & ~    & ~  & ~ &~    & ~  \\
\hfill --- Ours & 81.93 & 82.00     & 81.85  &   78.39    &   77.52        &    79.29    & 77.97 & 77.80     & 78.14  \\ 

\hfill --- Reported & N/A & N/A & N/A & N/A & N/A & N/A & N/A & N/A & N/A \\
\midrule

     {SciFive+MIMIC-T5} 
 & ~ &~     & ~  & ~ & ~    & ~  & ~ &~    & ~  \\
\hfill --- Ours & 85.42 & 85.52     & 85.32  & 
 82.44 & 81.22     & 83.70  & 85.67 & 83.18     & 88.31s  \\
 \hfill ---Reported &   84.78 & 85.37 & 84.79 &  80.35 &  79.24 & 81.49 & 86.73 & 86.37 & 87.09
\\
\bottomrule
\end{tabular}}
\caption{Evaluations on three biomedical datasets in {SciFive+MIMIC-T5} were also reproduced. {MIMIC-T5} and \textsc{T5}'s performances are also shown in the table. }
\label{replicate_biomedical_results}
\end{table*}

\label{sec:appendix_cross_validation_results}

\begin{table*}
\centering
\adjustbox{max width=\linewidth}{%
        \begin{tabular}{p{15mm}cccccc}
    MedNLI \\
    \toprule
    Metrics &
    ~ &
    \begin{tabular}{c}{T5-Den}  \\     \end{tabular} &
    \begin{tabular}{c}{T5-Sup}  \\    \end{tabular} &
    \begin{tabular}{c}{FLAN-T5}  \\    \end{tabular} &
    \begin{tabular}{c}{MIMIC-T5}  \\       \end{tabular} &
    \begin{tabular}{c}{SciFive+MIMIC-T5} \\        \end{tabular} \\ \midrule
    
        \multirow{3}{*}{Accuracy} &
        \multirow{2}{*}{Mean} &
        $85.89$ &
        $84.94$ &
        $86.03$ &
        $86.75$ &
        $85.64$ \\ 
        ~ &
        &
        $\spadesuit$ &
        $\clubsuit$ &
        $\blacklozenge  $ & 
        $\uparrow \spadesuit 0.86$, $\uparrow \clubsuit 1.81$, $\uparrow \blacklozenge  0.72$  &
        $\downarrow \spadesuit 0.25$, $\uparrow \clubsuit 0.70$, $\downarrow \blacklozenge  0.39$ \\
        \cmidrule{2-7}        
        ~ &
        95\% CI &
        [$85.34, 86.24$] &
        [$84.38, 85.49$] &
        [$85.72, 86.43$] &
        [$86.27, 87.29$] &
        [$85.13, 86.37$] \\
        
    \bottomrule 
    \end{tabular}} 
    
    \medskip

        \adjustbox{max width=\linewidth}{%
        \begin{tabular}{cccccc}
        MedNLI-Accuracy-T-test \\
        \toprule
        Models &
        \begin{tabular}{c} {T5-Den} \\  \end{tabular} &
        \begin{tabular}{c} {T5-Sup} \\  \end{tabular} &
        \begin{tabular}{c} {FLAN-T5} \\  \end{tabular} &
        \begin{tabular}{c}{MIMIC-T5}  \\  \end{tabular} &
        \begin{tabular}{c} {SciFive+MIMIC-T5}  \\  \end{tabular} \\ \midrule

        {T5-Den-Large} & -  & -    & -  & - & - \\ \midrule
        {T5-Sup-Large} & t(df=4)=-2.63, \textit{P}=.06
        & - & - & - & - \\ \midrule 
        {FLAN-T5-Large} & t(df=4)=0.31, \textit{P}=.77
        & t(df=4)=3.17, \textit{P}=.03 & - & - & - \\ \midrule 
        {MIMIC-T5-Large}  & t(df=4)=1.97, \textit{P}=.12
        & t(df=4)=4.14, \textit{P}=.01 
        & t(df=4)=1.75, \textit{P}=.15  & - & - \\ \midrule 
        {SciFive+MIMIC-T5-Large}  & t(df=4)=-0.45, \textit{P}=.68
        & t(df=4)=1.18, \textit{P}=.30 
        & t(df=4)=-1.45, \textit{P}=.22 
        & t(df=4)=-2.18, \textit{P}=.10  & - \\
        
            \bottomrule
            \end{tabular} }
        \caption{\label{mednli_large_cv}
        5-fold cross-validation results across models on MedNLI dataset and the t-test results among \textsc{T5}, {MIMIC-T5}, and {SciFive+MIMIC-T5}
}

\end{table*}

\begin{table*}
    \centering
 \adjustbox{max width=\linewidth}{%
        \begin{tabular}{p{15mm}cccccc}
    RadQA \\
    \toprule
    Metrics &
    ~ &
    \begin{tabular}{c}{T5-Den}  \\     \end{tabular} &
    \begin{tabular}{c}{T5-Sup}  \\    \end{tabular} &
    \begin{tabular}{c}{FLAN-T5}  \\    \end{tabular} &
    \begin{tabular}{c}{MIMIC-T5}  \\       \end{tabular} &
    \begin{tabular}{c}{SciFive+MIMIC-T5} \\        \end{tabular} \\ \midrule
    
        \multirow{3}{*}{EM} &
        \multirow{2}{*}{Mean} &
        $52.57$ &
        $51.95$ &
        $53.03$ &
        $54.50$ &
        $53.42$ \\ 
        ~ &
        &
        $\spadesuit$ &
        $\clubsuit$ &
        $\blacklozenge  $ & 
        $\uparrow \spadesuit 1.92$, $\uparrow \clubsuit 2.54$, $\uparrow \blacklozenge  1.47$  &
        $\uparrow \spadesuit 0.85$, $\uparrow \clubsuit 1.47$, $\uparrow \blacklozenge  0.39$ \\
        \cmidrule{2-7}        
        ~ &
        95\% CI &
        [$51.9, 53.27$] &
        [$51.04, 52.95$] &
        [$51.79, 54.01$] &
        [$54.04, 55.05$] &
        [$52.69, 54.22$] \\
        \midrule
        \multirow{3}{*}{F1} &
        \multirow{2}{*}{Mean} &
        $68.90$ &
        $68.68$ &
        $70.62$ &
        $73.40$ &
        $70.37$ \\ 
        ~ &
        &
        $\spadesuit$ &
        $\clubsuit$ &
        $\blacklozenge  $ & 
        $\uparrow \spadesuit 4.50$, $\uparrow \clubsuit 4.72$, $\uparrow \blacklozenge  2.78$  &
        $\uparrow \spadesuit 1.47$, $\uparrow \clubsuit 1.7$, $\downarrow \blacklozenge  0.25$ \\
        \cmidrule{2-7}        
        ~ &
        95\% CI &
        [$68.04, 69.79$] &
        [$67.82, 69.77$] &
        [$69.76, 71.34$] &
        [$72.75, 74.00$] &
        [$69.44, 71.49$] \\
        
    \bottomrule 
    \end{tabular}} 
    
    \medskip

        \adjustbox{max width=\linewidth}{%
        \begin{tabular}{cccccc}
        RadQA-EM-T-test \\
        \toprule
        Models &
        \begin{tabular}{c} {T5-Den} \\  \end{tabular} &
        \begin{tabular}{c} {T5-Sup} \\  \end{tabular} &
        \begin{tabular}{c} {FLAN-T5} \\  \end{tabular} &
        \begin{tabular}{c}{MIMIC-T5}  \\  \end{tabular} &
        \begin{tabular}{c} {SciFive+MIMIC-T5}  \\  \end{tabular} \\ \midrule

        {T5-Den-Large} & -  & -    & -  & - & - \\ \midrule
        {T5-Sup-Large} & t(df=4)=-0.91, \textit{P}=.41
        & - & - & - & - \\ \midrule 
        {FLAN-T5-Large} & t(df=4)=0.47, \textit{P}=.67
        & t(df=4)=1.4, \textit{P}=.23 & - & - & - \\ \midrule 
        {MIMIC-T5-Large}  & t(df=4)=4.3, \textit{P}=.01
        & t(df=4)=2.92, \textit{P}=.04 
        & t(df=4)=1.84, \textit{P}=.14  & - & - \\ \midrule 
        {SciFive+MIMIC-T5-Large}  & t(df=4)=1.19, \textit{P}=.30
        & t(df=4)=1.95, \textit{P}=.12 
        & t(df=4)=0.54, \textit{P}=.62 
        & t(df=4)=-1.97, \textit{P}=.12  & - \\
        
            \bottomrule
            \end{tabular} }
        
        \adjustbox{max width=\linewidth}{%
        \begin{tabular}{cccccc}
        RadQA-F1-T-test \\
        \toprule
        Models &
        \begin{tabular}{c} {T5-Den} \\  \end{tabular} &
        \begin{tabular}{c} {T5-Sup} \\  \end{tabular} &
        \begin{tabular}{c} {FLAN-T5} \\  \end{tabular} &
        \begin{tabular}{c}{MIMIC-T5}  \\  \end{tabular} &
        \begin{tabular}{c} {SciFive+MIMIC-T5}  \\  \end{tabular} \\ \midrule

        {T5-Den-Large} & -  & -    & -  & - & - \\ \midrule
        {T5-Sup-Large} & t(df=4)=-0.55, \textit{P}=.61
        & - & - & - & - \\ \midrule 
        {FLAN-T5-Large} & t(df=4)=2.53, \textit{P}=.06
        & t(df=4)=3.65, \textit{P}=.02 & - & - & - \\ \midrule 
        {MIMIC-T5-Large}  & t(df=4)=5.95, \textit{P}=.004
        & t(df=4)=5.2, \textit{P}=.007 
        & t(df=4)=3.5, \textit{P}=.02  & - & - \\ \midrule 
        {SciFive+MIMIC-T5-Large}  & t(df=4)=2.19, \textit{P}=.09
        & t(df=4)=2.15, \textit{P}=.10 
        & t(df=4)=-0.32, \textit{P}=.76 
        & t(df=4)=-5.47, \textit{P}=.005  & - \\
        
            \bottomrule
            \end{tabular} }
    \caption{5-fold cross-validation results across models on RadQA dataset and the t-test results among \textsc{T5}, {MIMIC-T5}, and {SciFive+MIMIC-T5}.}
    \label{tab:radqa_large_cv}

\end{table*}

\begin{table*}
\centering
\adjustbox{max width=\linewidth}{%
        \begin{tabular}{p{15mm}cccccc}
    CLIP \\
    \toprule
    Metrics &
    ~ &
    \begin{tabular}{c}{T5-Den}  \\     \end{tabular} &
    \begin{tabular}{c}{T5-Sup}  \\    \end{tabular} &
    \begin{tabular}{c}{FLAN-T5}  \\    \end{tabular} &
    \begin{tabular}{c}{MIMIC-T5}  \\       \end{tabular} &
    \begin{tabular}{c}{SciFive+MIMIC-T5} \\        \end{tabular} \\ \midrule
    
        \multirow{3}{*}{Macro-F1} &
        \multirow{2}{*}{Mean} &
        $63.86$ &
        $62.42$ &
        $64.33$ &
        $66.09$ &
        $63.46$ \\ 
        ~ &
        &
        $\spadesuit$ &
        $\clubsuit$ &
        $\blacklozenge  $ & 
        $\uparrow \spadesuit 2.23$, $\uparrow \clubsuit 3.67$, $\uparrow \blacklozenge  1.75$  &
        $\downarrow \spadesuit 0.4$, $\uparrow \clubsuit 1.04$, $\downarrow \blacklozenge  0.88$ \\
        \cmidrule{2-7}        
        ~ &
        95\% CI &
        [$62.18, 65.55$] &
        [$61.37, 63.4$] &
        [$62.94, 65.72$] &
        [$65.84, 66.26$] &
        [$62.2, 64.76$] \\
        \midrule
        \multirow{3}{*}{Micro-F1} &
        \multirow{2}{*}{Mean} &
        $78.62$ &
        $78.38$ &
        $79.40$ &
        $79.28$ &
        $78.23$ \\ 
        ~ &
        &
        $\spadesuit$ &
        $\clubsuit$ &
        $\blacklozenge  $ & 
        $\uparrow \spadesuit 0.67$, $\uparrow \clubsuit 0.9$, $\downarrow \blacklozenge  0.12$  &
        $\downarrow \spadesuit 0.38$, $\downarrow \clubsuit 0.15$, $\downarrow \blacklozenge  1.17$ \\
        \cmidrule{2-7}        
        ~ &
        95\% CI &
        [$77.48, 79.79$] &
        [$78.03, 78.77$] &
        [$78.85, 79.94$] &
        [$78.51, 80.01$] &
        [$77.93, 78.51$] \\
        
    \bottomrule 
    \end{tabular}} 
    
    \medskip

        \adjustbox{max width=\linewidth}{%
        \begin{tabular}{cccccc}
        CLIP-Macro-F1-T-test \\
        \toprule
        Models &
        \begin{tabular}{c} {T5-Den} \\  \end{tabular} &
        \begin{tabular}{c} {T5-Sup} \\  \end{tabular} &
        \begin{tabular}{c} {FLAN-T5} \\  \end{tabular} &
        \begin{tabular}{c}{MIMIC-T5}  \\  \end{tabular} &
        \begin{tabular}{c} {SciFive+MIMIC-T5}  \\  \end{tabular} \\ \midrule

        {T5-Den-Large} & -  & -    & -  & - & - \\ \midrule
        {T5-Sup-Large} & t(df=4)=-1.39, \textit{P}=.24
        & - & - & - & - \\ \midrule 
        {FLAN-T5-Large} & t(df=4)=0.38, \textit{P}=.72
        & t(df=4)=1.83, \textit{P}=.14 & - & - & - \\ \midrule 
        {MIMIC-T5-Large}  & t(df=4)=2.15, \textit{P}=.10
        & t(df=4)=5.86, \textit{P}=.004 
        & t(df=4)=2.02, \textit{P}=.11  & - & - \\ \midrule 
        {SciFive+MIMIC-T5-Large}  & t(df=4)=-0.3, \textit{P}=.78
        & t(df=4)=0.95, \textit{P}=.39 
        & t(df=4)=-0.73, \textit{P}=.51 
        & t(df=4)=-4.33, \textit{P}=.01  & - \\
        
            \bottomrule
            \end{tabular} }
        
        \adjustbox{max width=\linewidth}{%
        \begin{tabular}{cccccc}
        CLIP-Micro-F1-T-test \\
        \toprule
        Models &
        \begin{tabular}{c} {T5-Den} \\  \end{tabular} &
        \begin{tabular}{c} {T5-Sup} \\  \end{tabular} &
        \begin{tabular}{c} {FLAN-T5} \\  \end{tabular} &
        \begin{tabular}{c}{MIMIC-T5}  \\  \end{tabular} &
        \begin{tabular}{c} {SciFive+MIMIC-T5}  \\  \end{tabular} \\ \midrule

        {T5-Den-Large} & -  & -    & -  & - & - \\ \midrule
        {T5-Sup-Large} & t(df=4)=-0.34, \textit{P}=.75
        & - & - & - & - \\ \midrule 
        {FLAN-T5-Large} & t(df=4)=1.83, \textit{P}=.14
        & t(df=4)=2.55, \textit{P}=.06 & - & - & - \\ \midrule 
        {MIMIC-T5-Large}  & t(df=4)=0.64, \textit{P}=.56
        & t(df=4)=1.89, \textit{P}=.13 
        & t(df=4)=-0.18, \textit{P}=.87  & - & - \\ \midrule 
        {SciFive+MIMIC-T5-Large}  & t(df=4)=-0.67, \textit{P}=.54
        & t(df=4)=-0.49, \textit{P}=.65 
        & t(df=4)=-4.19, \textit{P}=.01 
        & t(df=4)=-2.19, \textit{P}=.09  & - \\
        
            \bottomrule
            \end{tabular} }
\caption{5-fold cross-validation results across models on CLIP dataset and the t-test results among \textsc{T5}, {MIMIC-T5}, and {SciFive+MIMIC-T5}}
\label{tab:clip_large_cv}
\end{table*}

\begin{table*}
\centering
\adjustbox{max width=\linewidth}{%
        \begin{tabular}{p{15mm}cccccc}
    HOC \\
    \toprule
    Metrics &
    ~ &
    \begin{tabular}{c}{T5-Den}  \\     \end{tabular} &
    \begin{tabular}{c}{T5-Sup}  \\    \end{tabular} &
    \begin{tabular}{c}{FLAN-T5}  \\    \end{tabular} &
    \begin{tabular}{c}{MIMIC-T5}  \\       \end{tabular} &
    \begin{tabular}{c}{SciFive+MIMIC-T5} \\        \end{tabular} \\ \midrule
    
        \multirow{3}{*}{F1} &
        \multirow{2}{*}{Mean} &
        $84.81$ &
        $84.76$ &
        $84.79$ &
        $82.75$ &
        $85.1$ \\ 
        ~ &
        &
        $\spadesuit$ &
        $\clubsuit$ &
        $\blacklozenge  $ & 
        $\downarrow \spadesuit 2.07$, $\downarrow \clubsuit 2.02$, $\downarrow \blacklozenge  2.05$  &
        $\uparrow \spadesuit 0.29$, $\uparrow \clubsuit 0.34$, $\uparrow \blacklozenge  0.31$ \\
        \cmidrule{2-7}        
        ~ &
        95\% CI &
        [$84.26, 85.49$] &
        [$84.38, 85.11$] &
        [$84.22, 85.26$] &
        [$82.53, 82.98$] &
        [$84.81, 85.4$] \\
        \midrule
        \multirow{3}{*}{P} &
        \multirow{2}{*}{Mean} &
        $84.72$ &
        $84.77$ &
        $84.60$ &
        $82.90$ &
        $85.04$ \\ 
        ~ &
        &
        $\spadesuit$ &
        $\clubsuit$ &
        $\blacklozenge  $ & 
        $\downarrow \spadesuit 1.82$, $\downarrow \clubsuit 1.87$, $\downarrow \blacklozenge  1.7$  &
        $\uparrow \spadesuit 0.32$, $\uparrow \clubsuit 0.27$, $\uparrow \blacklozenge  0.44$ \\
        \cmidrule{2-7}        
        ~ &
        95\% CI &
        [$84.04, 85.47$] &
        [$84.43, 85.11$] &
        [$83.94, 85.18$] &
        [$82.77, 83.04$] &
        [$84.67, 85.39$] \\
        \midrule
        \multirow{3}{*}{R} &
        \multirow{2}{*}{Mean} &
        $84.91$ &
        $84.76$ &
        $84.99$ &
        $82.60$ &
        $85.17$ \\ 
        ~ &
        &
        $\spadesuit$ &
        $\clubsuit$ &
        $\blacklozenge  $ & 
        $\downarrow \spadesuit 2.31$, $\downarrow \clubsuit 2.16$, $\downarrow \blacklozenge  2.4$  &
        $\uparrow \spadesuit 0.26$, $\uparrow \clubsuit 0.41$, $\uparrow \blacklozenge  0.18$ \\
        \cmidrule{2-7}        
        ~ &
        95\% CI &
        [$84.49, 85.52$] &
        [$84.28, 85.21$] &
        [$84.47, 85.4$] &
        [$82.26, 83.01$] &
        [$84.85, 85.52$] \\
        
    \bottomrule 
    \end{tabular}} 
    
    \medskip

        \adjustbox{max width=\linewidth}{%
        \begin{tabular}{cccccc}
        HOC-F1-T-test \\
        \toprule
        Models &
        \begin{tabular}{c} {T5-Den} \\  \end{tabular} &
        \begin{tabular}{c} {T5-Sup} \\  \end{tabular} &
        \begin{tabular}{c} {FLAN-T5} \\  \end{tabular} &
        \begin{tabular}{c}{MIMIC-T5}  \\  \end{tabular} &
        \begin{tabular}{c} {SciFive+MIMIC-T5}  \\  \end{tabular} \\ \midrule

        {T5-Den-Large} & -  & -    & -  & - & - \\ \midrule
        {T5-Sup-Large} & t(df=4)=-0.09, \textit{P}=.93
        & - & - & - & - \\ \midrule 
        {FLAN-T5-Large} & t(df=4)=-0.05, \textit{P}=.96
        & t(df=4)=0.07, \textit{P}=.95 & - & - & - \\ \midrule 
        {MIMIC-T5-Large}  & t(df=4)=-5.75, \textit{P}=.005
        & t(df=4)=-6.86, \textit{P}=.002 
        & t(df=4)=-7.02, \textit{P}=.002  & - & - \\ \midrule 
        {SciFive+MIMIC-T5-Large}  & t(df=4)=1.05, \textit{P}=.35
        & t(df=4)=1.31, \textit{P}=.26 
        & t(df=4)=1.08, \textit{P}=.34 
        & t(df=4)=10.06, \textit{P}=<.001  & - \\
        
            \bottomrule
            \end{tabular} }
        
        \adjustbox{max width=\linewidth}{%
        \begin{tabular}{cccccc}
        HOC-P-T-test \\
        \toprule
        Models &
        \begin{tabular}{c} {T5-Den} \\  \end{tabular} &
        \begin{tabular}{c} {T5-Sup} \\  \end{tabular} &
        \begin{tabular}{c} {FLAN-T5} \\  \end{tabular} &
        \begin{tabular}{c}{MIMIC-T5}  \\  \end{tabular} &
        \begin{tabular}{c} {SciFive+MIMIC-T5}  \\  \end{tabular} \\ \midrule

        {T5-Den-Large} & -  & -    & -  & - & - \\ \midrule
        {T5-Sup-Large} & t(df=4)=0.1, \textit{P}=.92
        & - & - & - & - \\ \midrule 
        {FLAN-T5-Large} & t(df=4)=-0.29, \textit{P}=.79
        & t(df=4)=-0.36, \textit{P}=.73 & - & - & - \\ \midrule 
        {MIMIC-T5-Large}  & t(df=4)=-4.3, \textit{P}=.01
        & t(df=4)=-6.62, \textit{P}=.003 
        & t(df=4)=-5.46, \textit{P}=.005  & - & - \\ \midrule 
        {SciFive+MIMIC-T5-Large}  & t(df=4)=0.86, \textit{P}=.44
        & t(df=4)=1.19, \textit{P}=.30 
        & t(df=4)=1.41, \textit{P}=.23 
        & t(df=4)=7.87, \textit{P}=.001  & - \\
        
            \bottomrule
            \end{tabular} }
        
        \adjustbox{max width=\linewidth}{%
        \begin{tabular}{cccccc}
        HOC-R-T-test \\
        \toprule
        Models &
        \begin{tabular}{c} {T5-Den} \\  \end{tabular} &
        \begin{tabular}{c} {T5-Sup} \\  \end{tabular} &
        \begin{tabular}{c} {FLAN-T5} \\  \end{tabular} &
        \begin{tabular}{c}{MIMIC-T5}  \\  \end{tabular} &
        \begin{tabular}{c} {SciFive+MIMIC-T5}  \\  \end{tabular} \\ \midrule

        {T5-Den-Large} & -  & -    & -  & - & - \\ \midrule
        {T5-Sup-Large} & t(df=4)=-0.28, \textit{P}=.79
        & - & - & - & - \\ \midrule 
        {FLAN-T5-Large} & t(df=4)=0.22, \textit{P}=.84
        & t(df=4)=0.65, \textit{P}=.55 & - & - & - \\ \midrule 
        {MIMIC-T5-Large}  & t(df=4)=-6.84, \textit{P}=.002
        & t(df=4)=-6.45, \textit{P}=.003 
        & t(df=4)=-8.09, \textit{P}=.001  & - & - \\ \midrule 
        {SciFive+MIMIC-T5-Large}  & t(df=4)=1.26, \textit{P}=.27
        & t(df=4)=1.11, \textit{P}=.33 
        & t(df=4)=0.57, \textit{P}=.60 
        & t(df=4)=10.45, \textit{P}=<.001  & - \\
        
            \bottomrule
            \end{tabular} }
\caption{5-fold cross-validation results across models on HOC dataset and the t-test results among \textsc{T5}, {MIMIC-T5}, and {SciFive+MIMIC-T5}}
\label{tab:hoc_large_Cv}
\end{table*}

\begin{table*}
\centering
\adjustbox{max width=\linewidth}{%
        \begin{tabular}{p{15mm}cccccc}
    BC5CDR \\
    \toprule
    Metrics &
    ~ &
    \begin{tabular}{c}{T5-Den}  \\     \end{tabular} &
    \begin{tabular}{c}{T5-Sup}  \\    \end{tabular} &
    \begin{tabular}{c}{FLAN-T5}  \\    \end{tabular} &
    \begin{tabular}{c}{MIMIC-T5}  \\       \end{tabular} &
    \begin{tabular}{c}{SciFive+MIMIC-T5} \\        \end{tabular} \\ \midrule
    
        \multirow{3}{*}{F1} &
        \multirow{2}{*}{Mean} &
        $82.90$ &
        $82.94$ &
        $83.30$ &
        $81.17$ &
        $83.61$ \\ 
        ~ &
        &
        $\spadesuit$ &
        $\clubsuit$ &
        $\blacklozenge  $ & 
        $\downarrow \spadesuit 1.73$, $\downarrow \clubsuit 1.77$, $\downarrow \blacklozenge  2.13$  &
        $\uparrow \spadesuit 0.71$, $\uparrow \clubsuit 0.68$, $\uparrow \blacklozenge  0.31$ \\
        \cmidrule{2-7}        
        ~ &
        95\% CI &
        [$82.33, 83.38$] &
        [$82.51, 83.45$] &
        [$82.89, 83.82$] &
        [$80.92, 81.53$] &
        [$83.38, 83.82$] \\
        \midrule
        \multirow{3}{*}{P} &
        \multirow{2}{*}{Mean} &
        $81.67$ &
        $81.62$ &
        $82.14$ &
        $80.62$ &
        $83.07$ \\ 
        ~ &
        &
        $\spadesuit$ &
        $\clubsuit$ &
        $\blacklozenge  $ & 
        $\downarrow \spadesuit 1.05$, $\downarrow \clubsuit 1.01$, $\downarrow \blacklozenge  1.52$  &
        $\uparrow \spadesuit 1.4$, $\uparrow \clubsuit 1.44$, $\uparrow \blacklozenge  0.93$ \\
        \cmidrule{2-7}        
        ~ &
        95\% CI &
        [$80.77, 82.51$] &
        [$81.30, 81.96$] &
        [$81.53, 82.72$] &
        [$80.20, 80.94$] &
        [$82.51, 83.6$] \\
        \midrule
        \multirow{3}{*}{R} &
        \multirow{2}{*}{Mean} &
        $84.19$ &
        $84.29$ &
        $84.50$ &
        $81.73$ &
        $84.17$ \\ 
        ~ &
        &
        $\spadesuit$ &
        $\clubsuit$ &
        $\blacklozenge  $ & 
        $\downarrow \spadesuit 2.46$, $\downarrow \clubsuit 2.56$, $\downarrow \blacklozenge  2.76$  &
        $\downarrow \spadesuit 0.02$, $\downarrow \clubsuit 0.12$, $\downarrow \blacklozenge  0.32$ \\
        \cmidrule{2-7}        
        ~ &
        95\% CI &
        [$83.64, 84.63$] &
        [$83.55, 85.05$] &
        [$84.23, 84.93$] &
        [$81.23, 82.31$] &
        [$83.60, 84.82$] \\
        
    \bottomrule 
    \end{tabular}} 
    
    \medskip

        \adjustbox{max width=\linewidth}{%
        \begin{tabular}{cccccc}
        BC5CDR-F1-T-test \\
        \toprule
        Models &
        \begin{tabular}{c} {T5-Den} \\  \end{tabular} &
        \begin{tabular}{c} {T5-Sup} \\  \end{tabular} &
        \begin{tabular}{c} {FLAN-T5} \\  \end{tabular} &
        \begin{tabular}{c}{MIMIC-T5}  \\  \end{tabular} &
        \begin{tabular}{c} {SciFive+MIMIC-T5}  \\  \end{tabular} \\ \midrule

        {T5-Den} & -  & -    & -  & - & - \\ \midrule
        {T5-Sup} & t(df=4)=0.06, \textit{P}=.96
        & - & - & - & - \\ \midrule 
        {FLAN-T5} & t(df=4)=0.82, \textit{P}=.46
        & t(df=4)=2.16, \textit{P}=.10 & - & - & - \\ \midrule 
        {MIMIC-T5}  & t(df=4)=-9.49, \textit{P}=<.001
        & t(df=4)=-4.01, \textit{P}=.02 
        & t(df=4)=-5.29, \textit{P}=.006  & - & - \\ \midrule 
        {SciFive+MIMIC-T5}  & t(df=4)=2.38, \textit{P}=.08
        & t(df=4)=1.7, \textit{P}=.16 
        & t(df=4)=0.76, \textit{P}=.49 
        & t(df=4)=14.45, \textit{P}=<.001  & - \\
        
            \bottomrule
            \end{tabular} }
        
        \adjustbox{max width=\linewidth}{%
        \begin{tabular}{cccccc}
        BC5CDR-P-T-test \\
        \toprule
        Models &
        \begin{tabular}{c} {T5-Den} \\  \end{tabular} &
        \begin{tabular}{c} {T5-Sup} \\  \end{tabular} &
        \begin{tabular}{c} {FLAN-T5} \\  \end{tabular} &
        \begin{tabular}{c}{MIMIC-T5}  \\  \end{tabular} &
        \begin{tabular}{c} {SciFive+MIMIC-T5}  \\  \end{tabular} \\ \midrule

        {T5-Den} & -  & -    & -  & - & - \\ \midrule
        {T5-Sup} & t(df=4)=-0.08, \textit{P}=.94
        & - & - & - & - \\ \midrule 
        {FLAN-T5} & t(df=4)=0.71, \textit{P}=.52
        & t(df=4)=2.37, \textit{P}=.08 & - & - & - \\ \midrule 
        {MIMIC-T5}  & t(df=4)=-3.07, \textit{P}=.04
        & t(df=4)=-3.23, \textit{P}=.03 
        & t(df=4)=-3.09, \textit{P}=.04  & - & - \\ \midrule 
        {SciFive+MIMIC-T5}  & t(df=4)=2.08, \textit{P}=.11
        & t(df=4)=3.41, \textit{P}=.03 
        & t(df=4)=1.73, \textit{P}=.16 
        & t(df=4)=6.04, \textit{P}=.004  & - \\
        
            \bottomrule
            \end{tabular} }
        
        \adjustbox{max width=\linewidth}{%
        \begin{tabular}{cccccc}
        BC5CDR-R-T-test \\
        \toprule
        Models &
        \begin{tabular}{c} {T5-Den} \\  \end{tabular} &
        \begin{tabular}{c} {T5-Sup} \\  \end{tabular} &
        \begin{tabular}{c} {FLAN-T5} \\  \end{tabular} &
        \begin{tabular}{c}{MIMIC-T5}  \\  \end{tabular} &
        \begin{tabular}{c} {SciFive+MIMIC-T5}  \\  \end{tabular} \\ \midrule

        {T5-Den} & -  & -    & -  & - & - \\ \midrule
        {T5-Sup} & t(df=4)=0.19, \textit{P}=.86
        & - & - & - & - \\ \midrule 
        {FLAN-T5} & t(df=4)=0.8, \textit{P}=.47
        & t(df=4)=0.78, \textit{P}=.48 & - & - & - \\ \midrule 
        {MIMIC-T5}  & t(df=4)=-6.28, \textit{P}=.003
        & t(df=4)=-3.88, \textit{P}=.02 
        & t(df=4)=-6.14, \textit{P}=.004  & - & - \\ \midrule 
        {SciFive+MIMIC-T5}  & t(df=4)=-0.07, \textit{P}=.95
        & t(df=4)=-0.17, \textit{P}=.88 
        & t(df=4)=-0.61, \textit{P}=.57 
        & t(df=4)=10.35, \textit{P}=<.001  & - \\
        
            \bottomrule
            \end{tabular} }
\caption{5-fold cross-validation results across models on BC5CDR-disease dataset and the t-test results among \textsc{T5}, {MIMIC-T5}, and {SciFive+MIMIC-T5} \label{tab:bc5cdr_large_cv}}
\end{table*}

\begin{table*}
\centering
\adjustbox{max width=\linewidth}{%
        \begin{tabular}{p{15mm}cccccc}
    NCBI \\
    \toprule
    Metrics &
    ~ &
    \begin{tabular}{c}{T5-Den}  \\     \end{tabular} &
    \begin{tabular}{c}{T5-Sup}  \\    \end{tabular} &
    \begin{tabular}{c}{FLAN-T5}  \\    \end{tabular} &
    \begin{tabular}{c}{MIMIC-T5}  \\       \end{tabular} &
    \begin{tabular}{c}{SciFive+MIMIC-T5} \\        \end{tabular} \\ \midrule
    
        \multirow{3}{*}{F1} &
        \multirow{2}{*}{Mean} &
        $85.08$ &
        $84.58$ &
        $84.44$ &
        $79.10$ &
        $85.43$ \\ 
        ~ &
        &
        $\spadesuit$ &
        $\clubsuit$ &
        $\blacklozenge  $ & 
        $\downarrow \spadesuit 5.98$, $\downarrow \clubsuit 5.49$, $\downarrow \blacklozenge  5.35$  &
        $\uparrow \spadesuit 0.36$, $\uparrow \clubsuit 0.85$, $\uparrow \blacklozenge  0.99$ \\
        \cmidrule{2-7}        
        ~ &
        95\% CI &
        [$84.61, 85.46$] &
        [$84.22, 85.02$] &
        [$83.94, 85.03$] &
        [$78.28, 79.94$] &
        [$84.89, 86.06$] \\
        \midrule
        \multirow{3}{*}{P} &
        \multirow{2}{*}{Mean} &
        $83.49$ &
        $83.58$ &
        $82.20$ &
        $78.35$ &
        $83.80$ \\ 
        ~ &
        &
        $\spadesuit$ &
        $\clubsuit$ &
        $\blacklozenge  $ & 
        $\downarrow \spadesuit 5.14$, $\downarrow \clubsuit 5.24$, $\downarrow \blacklozenge  3.85$  &
        $\uparrow \spadesuit 0.31$, $\uparrow \clubsuit 0.21$, $\uparrow \blacklozenge  1.60$ \\
        \cmidrule{2-7}        
        ~ &
        95\% CI &
        [$82.78, 84.16$] &
        [$83.13, 84.12$] &
        [$81.59, 83.04$] &
        [$77.57, 78.87$] &
        [$82.95, 84.68$] \\
        \midrule
        \multirow{3}{*}{R} &
        \multirow{2}{*}{Mean} &
        $86.73$ &
        $85.61$ &
        $86.82$ &
        $79.87$ &
        $87.14$ \\ 
        ~ &
        &
        $\spadesuit$ &
        $\clubsuit$ &
        $\blacklozenge  $ & 
        $\downarrow \spadesuit 6.86$, $\downarrow \clubsuit 5.74$, $\downarrow \blacklozenge  6.95$  &
        $\uparrow \spadesuit 0.41$, $\uparrow \clubsuit 1.54$, $\uparrow \blacklozenge  0.32$ \\
        \cmidrule{2-7}        
        ~ &
        95\% CI &
        [$86.31, 87.22$] &
        [$85.16, 86.15$] &
        [$86.31, 87.30$] &
        [$78.70, 81.30$] &
        [$86.86, 87.50$] \\
        
    \bottomrule 
    \end{tabular}} 
    
    \medskip

        \adjustbox{max width=\linewidth}{%
        \begin{tabular}{cccccc}
        NCBI-F1-T-test \\
        \toprule
        Models &
        \begin{tabular}{c} {T5-Den} \\  \end{tabular} &
        \begin{tabular}{c} {T5-Sup} \\  \end{tabular} &
        \begin{tabular}{c} {FLAN-T5} \\  \end{tabular} &
        \begin{tabular}{c}{MIMIC-T5}  \\  \end{tabular} &
        \begin{tabular}{c} {SciFive+MIMIC-T5}  \\  \end{tabular} \\ \midrule

        {T5-Den} & -  & -    & -  & - & - \\ \midrule
        {T5-Sup} & t(df=4)=-0.95, \textit{P}=.40
        & - & - & - & - \\ \midrule 
        {FLAN-T5} & t(df=4)=-1.73, \textit{P}=.16
        & t(df=4)=-0.33, \textit{P}=.76 & - & - & - \\ \midrule 
        {MIMIC-T5}  & t(df=4)=-12.52, \textit{P}=<.001
        & t(df=4)=-10.1, \textit{P}=<.001 
        & t(df=4)=-11.66, \textit{P}=<.001  & - & - \\ \midrule 
        {SciFive+MIMIC-T5}  & t(df=4)=0.94, \textit{P}=.40
        & t(df=4)=1.71, \textit{P}=.16 
        & t(df=4)=1.62, \textit{P}=.18 
        & t(df=4)=8.65, \textit{P}=<.001  & - \\
        
            \bottomrule
            \end{tabular} }
        
        \adjustbox{max width=\linewidth}{%
        \begin{tabular}{cccccc}
        NCBI-P-T-test \\
        \toprule
        Models &
        \begin{tabular}{c} {T5-Den} \\  \end{tabular} &
        \begin{tabular}{c} {T5-Sup} \\  \end{tabular} &
        \begin{tabular}{c} {FLAN-T5} \\  \end{tabular} &
        \begin{tabular}{c}{MIMIC-T5}  \\  \end{tabular} &
        \begin{tabular}{c} {SciFive+MIMIC-T5}  \\  \end{tabular} \\ \midrule

        {T5-Den} & -  & -    & -  & - & - \\ \midrule
        {T5-Sup} & t(df=4)=0.14, \textit{P}=.90
        & - & - & - & - \\ \midrule 
        {FLAN-T5} & t(df=4)=-1.88, \textit{P}=.13
        & t(df=4)=-3.52, \textit{P}=.02 & - & - & - \\ \midrule 
        {MIMIC-T5}  & t(df=4)=-7.08, \textit{P}=.002
        & t(df=4)=-14.76, \textit{P}=<.001 
        & t(df=4)=-8.28, \textit{P}=.001  & - & - \\ \midrule 
        {SciFive+MIMIC-T5}  & t(df=4)=0.74, \textit{P}=.50
        & t(df=4)=0.33, \textit{P}=.76 
        & t(df=4)=1.93, \textit{P}=.13 
        & t(df=4)=6.65, \textit{P}=.003  & - \\
        
            \bottomrule
            \end{tabular} }
        
        \adjustbox{max width=\linewidth}{%
        \begin{tabular}{cccccc}
        NCBI-R-T-test \\
        \toprule
        Models &
        \begin{tabular}{c} {T5-Den} \\  \end{tabular} &
        \begin{tabular}{c} {T5-Sup} \\  \end{tabular} &
        \begin{tabular}{c} {FLAN-T5} \\  \end{tabular} &
        \begin{tabular}{c}{MIMIC-T5}  \\  \end{tabular} &
        \begin{tabular}{c} {SciFive+MIMIC-T5}  \\  \end{tabular} \\ \midrule

        {T5-Den} & -  & -    & -  & - & - \\ \midrule
        {T5-Sup} & t(df=4)=-2.0, \textit{P}=.12
        & - & - & - & - \\ \midrule 
        {FLAN-T5} & t(df=4)=0.38, \textit{P}=.72
        & t(df=4)=2.71, \textit{P}=.05 & - & - & - \\ \midrule 
        {MIMIC-T5}  & t(df=4)=-10.43, \textit{P}=<.001
        & t(df=4)=-7.16, \textit{P}=.002 
        & t(df=4)=-12.53, \textit{P}=<.001  & - & - \\ \midrule 
        {SciFive+MIMIC-T5}  & t(df=4)=1.07, \textit{P}=.34
        & t(df=4)=3.25, \textit{P}=.03 
        & t(df=4)=0.72, \textit{P}=.51 
        & t(df=4)=8.63, \textit{P}=<.001  & - \\
        
            \bottomrule
            \end{tabular} }
\caption{5-fold cross-validation results across models on NCBI-disease dataset and the t-test results among \textsc{T5}, {MIMIC-T5}, and {SciFive+MIMIC-T5} \label{tab:ncbi_large_cv}}

\end{table*}

\begin{table*}
\centering
\adjustbox{max width=\linewidth}{%
            \begin{tabular}{p{15mm}ccccc}
            \multicolumn{5}{l}{MIMIC-IV-Credibility \& Obstinacy} \\
            \toprule
            Metrics &
            ~ &
            \begin{tabular}{c}{T5-Sup}  \\    \end{tabular} &
            \begin{tabular}{c}{FLAN-T5}  \\    \end{tabular} &
            \begin{tabular}{c}{MIMIC-T5}  \\       \end{tabular} &
            \begin{tabular}{c}{SciFive+MIMIC-T5} \\        \end{tabular} \\ \midrule      
        
    \multirow{3}{*}{F1} &
    \multirow{2}{*}{Mean} &
    $76.04$ &
    $77.44$ &
    $74.98$ &
    $74.65$ \\ 
    ~ &
    &
    $\clubsuit$ &
    $\blacklozenge  $ & 
    $\downarrow \clubsuit 1.06$, $\downarrow \blacklozenge  2.46$  &
    $\downarrow \clubsuit 1.39$, $\downarrow \blacklozenge  2.79$ \\
    \cmidrule{2-6}        
    ~ &
    95\% CI &
    [$74.14, 77.94$] &
    [$74.96, 79.34$] &
    [$72.11, 77.75$] &
    [$70.59, 78.82$] \\
    
        \bottomrule 
        \end{tabular}} 
        
        \adjustbox{max width=\linewidth}{%
            \begin{tabular}{p{15mm}ccccc}
            \multicolumn{5}{l}{MIMIC-IV-Compliance} \\
            \toprule
            Metrics &
            ~ &
            \begin{tabular}{c}{T5-Sup}  \\    \end{tabular} &
            \begin{tabular}{c}{FLAN-T5}  \\    \end{tabular} &
            \begin{tabular}{c}{MIMIC-T5}  \\       \end{tabular} &
            \begin{tabular}{c}{SciFive+MIMIC-T5} \\        \end{tabular} \\ \midrule      
        
    \multirow{3}{*}{F1} &
    \multirow{2}{*}{Mean} &
    $92.94$ &
    $92.93$ &
    $91.57$ &
    $90.98$ \\ 
    ~ &
    &
    $\clubsuit$ &
    $\blacklozenge  $ & 
    $\downarrow \clubsuit 1.37$, $\downarrow \blacklozenge  1.36$  &
    $\downarrow \clubsuit 1.96$, $\downarrow \blacklozenge  1.95$ \\
    \cmidrule{2-6}        
    ~ &
    95\% CI &
    [$91.74, 94.18$] &
    [$91.27, 94.48$] &
    [$90.38, 92.70$] &
    [$89.90, 92.12$] \\
    
        \bottomrule 
        \end{tabular}} 
        
        \adjustbox{max width=\linewidth}{%
            \begin{tabular}{p{15mm}ccccc}
            \multicolumn{5}{l}{MIMIC-IV-Descriptors} \\
            \toprule
            Metrics &
            ~ &
            \begin{tabular}{c}{T5-Sup}  \\    \end{tabular} &
            \begin{tabular}{c}{FLAN-T5}  \\    \end{tabular} &
            \begin{tabular}{c}{MIMIC-T5}  \\       \end{tabular} &
            \begin{tabular}{c}{SciFive+MIMIC-T5} \\        \end{tabular} \\ \midrule      
        
    \multirow{3}{*}{F1} &
    \multirow{2}{*}{Mean} &
    $85.73$ &
    $86.31$ &
    $84.76$ &
    $86.28$ \\ 
    ~ &
    &
    $\clubsuit$ &
    $\blacklozenge  $ & 
    $\downarrow \clubsuit 0.97$, $\downarrow \blacklozenge  1.55$  &
    $\uparrow \clubsuit 0.54$, $\downarrow \blacklozenge  0.03$ \\
    \cmidrule{2-6}        
    ~ &
    95\% CI &
    [$83.45, 87.99$] &
    [$85.32, 87.03$] &
    [$81.90, 88.15$] &
    [$85.00, 87.79$] \\
    
        \bottomrule 
        \end{tabular}} 
        
    \medskip
    
        \adjustbox{max width=\linewidth}{%
        \begin{tabular}{ccccc}
        MIMIC-IV-Credibility \& Obstinacy-F1-T-test \\
        \toprule
        Models &
        \begin{tabular}{c} {T5-Sup} \\  \end{tabular} &
        \begin{tabular}{c} {FLAN-T5} \\  \end{tabular} &
        \begin{tabular}{c}{MIMIC-T5}  \\  \end{tabular} &
        \begin{tabular}{c} {SciFive+MIMIC-T5}  \\  \end{tabular} \\ \midrule

    {T5-Sup}
    & - & - & - & - \\ \midrule 
    {FLAN-T5} 
    & t(df=4)=0.64, \textit{P}=.56 & - & - & - \\ \midrule 
    {MIMIC-T5} 
    & t(df=4)=-0.52, \textit{P}=.63 
    & t(df=4)=-1.05, \textit{P}=.35  & - & - \\ \midrule 
    {SciFive+MIMIC-T5} 
    & t(df=4)=-0.81, \textit{P}=.46 
    & t(df=4)=-1.08, \textit{P}=.34 
    & t(df=4)=-0.12, \textit{P}=.91  & - \\
    
            \bottomrule
            \end{tabular} }
        
        \adjustbox{max width=\linewidth}{%
        \begin{tabular}{ccccc}
        MIMIC-IV-Compliance-F1-T-test \\
        \toprule
        Models &
        \begin{tabular}{c} {T5-Sup} \\  \end{tabular} &
        \begin{tabular}{c} {FLAN-T5} \\  \end{tabular} &
        \begin{tabular}{c}{MIMIC-T5}  \\  \end{tabular} &
        \begin{tabular}{c} {SciFive+MIMIC-T5}  \\  \end{tabular} \\ \midrule

    {T5-Sup}
    & - & - & - & - \\ \midrule 
    {FLAN-T5} 
    & t(df=4)=-0.0, \textit{P}=1.00 & - & - & - \\ \midrule 
    {MIMIC-T5} 
    & t(df=4)=-1.2, \textit{P}=.30 
    & t(df=4)=-1.2, \textit{P}=.30  & - & - \\ \midrule 
    {SciFive+MIMIC-T5} 
    & t(df=4)=-1.5, \textit{P}=.21 
    & t(df=4)=-1.93, \textit{P}=.13 
    & t(df=4)=-0.84, \textit{P}=.45  & - \\
    
            \bottomrule
            \end{tabular} }
        
        \adjustbox{max width=\linewidth}{%
        \begin{tabular}{ccccc}
        MIMIC-IV-Descriptors-F1-T-test \\
        \toprule
        Models &
        \begin{tabular}{c} {T5-Sup} \\  \end{tabular} &
        \begin{tabular}{c} {FLAN-T5} \\  \end{tabular} &
        \begin{tabular}{c}{MIMIC-T5}  \\  \end{tabular} &
        \begin{tabular}{c} {SciFive+MIMIC-T5}  \\  \end{tabular} \\ \midrule

    {T5-Sup}
    & - & - & - & - \\ \midrule 
    {FLAN-T5} 
    & t(df=4)=0.37, \textit{P}=.73 & - & - & - \\ \midrule 
    {MIMIC-T5} 
    & t(df=4)=-0.74, \textit{P}=.50 
    & t(df=4)=-0.9, \textit{P}=.42  & - & - \\ \midrule 
    {SciFive+MIMIC-T5} 
    & t(df=4)=0.33, \textit{P}=.76 
    & t(df=4)=-0.04, \textit{P}=.97 
    & t(df=4)=0.75, \textit{P}=.50  & - \\
    
            \bottomrule
            \end{tabular} }
\caption{5-fold cross-validation results across models on stigmatizing dataset from MIMIC-IV and the t-test results among T5-Sup, FLAN-T5, {MIMIC-T5}, and {SciFive+MIMIC-T5} \label{tab:mimic_stigma_cv}}
\end{table*}

\begin{table*}
\centering
\adjustbox{max width=\linewidth}{%
            \begin{tabular}{p{15mm}ccccc}
            \multicolumn{5}{l}{Hospital System-Credibility \& Obstinacy} \\
            \toprule
            Metrics &
            ~ &
            \begin{tabular}{c}{T5-Sup}  \\    \end{tabular} &
            \begin{tabular}{c}{FLAN-T5}  \\    \end{tabular} &
            \begin{tabular}{c}{MIMIC-T5}  \\       \end{tabular} &
            \begin{tabular}{c}{SciFive+MIMIC-T5} \\        \end{tabular} \\ \midrule      
        
    \multirow{3}{*}{F1} &
    \multirow{2}{*}{Mean} &
    $88.21$ &
    $90.38$ &
    $88.07$ &
    $87.31$ \\ 
    ~ &
    &
    $\clubsuit$ &
    $\blacklozenge  $ & 
    $\downarrow \clubsuit 0.15$, $\downarrow \blacklozenge  2.31$  &
    $\downarrow \clubsuit 0.9$, $\downarrow \blacklozenge  3.07$ \\
    \cmidrule{2-6}        
    ~ &
    95\% CI &
    [$86.12, 90.59$] &
    [$88.26, 92.33$] &
    [$83.11, 94.14$] &
    [$82.11, 91.76$] \\
    
        \bottomrule 
        \end{tabular}} 
        
        \adjustbox{max width=\linewidth}{%
            \begin{tabular}{p{15mm}ccccc}
            \multicolumn{5}{l}{Hospital System-Compliance} \\
            \toprule
            Metrics &
            ~ &
            \begin{tabular}{c}{T5-Sup}  \\    \end{tabular} &
            \begin{tabular}{c}{FLAN-T5}  \\    \end{tabular} &
            \begin{tabular}{c}{MIMIC-T5}  \\       \end{tabular} &
            \begin{tabular}{c}{SciFive+MIMIC-T5} \\        \end{tabular} \\ \midrule      
        
    \multirow{3}{*}{F1} &
    \multirow{2}{*}{Mean} &
    $86.71$ &
    $88.03$ &
    $84.72$ &
    $87.21$ \\ 
    ~ &
    &
    $\clubsuit$ &
    $\blacklozenge  $ & 
    $\downarrow \clubsuit 2.0$, $\downarrow \blacklozenge  3.31$  &
    $\uparrow \clubsuit 0.49$, $\downarrow \blacklozenge  0.82$ \\
    \cmidrule{2-6}        
    ~ &
    95\% CI &
    [$85.52, 88.19$] &
    [$86.84, 88.92$] &
    [$80.99, 88.67$] &
    [$86.82, 87.63$] \\
    
        \bottomrule 
        \end{tabular}} 
        
        \adjustbox{max width=\linewidth}{%
            \begin{tabular}{p{15mm}ccccc}
            \multicolumn{5}{l}{Hospital System-Descriptors} \\
            \toprule
            Metrics &
            ~ &
            \begin{tabular}{c}{T5-Sup}  \\    \end{tabular} &
            \begin{tabular}{c}{FLAN-T5}  \\    \end{tabular} &
            \begin{tabular}{c}{MIMIC-T5}  \\       \end{tabular} &
            \begin{tabular}{c}{SciFive+MIMIC-T5} \\        \end{tabular} \\ \midrule      
        
    \multirow{3}{*}{F1} &
    \multirow{2}{*}{Mean} &
    $90.58$ &
    $86.82$ &
    $88.85$ &
    $89.98$ \\ 
    ~ &
    &
    $\clubsuit$ &
    $\blacklozenge  $ & 
    $\downarrow \clubsuit 1.73$, $\uparrow \blacklozenge  2.03$  &
    $\downarrow \clubsuit 0.6$, $\uparrow \blacklozenge  3.17$ \\
    \cmidrule{2-6}        
    ~ &
    95\% CI &
    [$89.25, 91.37$] &
    [$79.94, 90.69$] &
    [$88.24, 89.46$] &
    [$89.21, 90.67$] \\
    
        \bottomrule 
        \end{tabular}} 
        
    \medskip
    
        \adjustbox{max width=\linewidth}{%
        \begin{tabular}{ccccc}
        Hospital System-Credibility \& Obstinacy-F1-T-test \\
        \toprule
        Models &
        \begin{tabular}{c} {T5-Sup} \\  \end{tabular} &
        \begin{tabular}{c} {FLAN-T5} \\  \end{tabular} &
        \begin{tabular}{c}{MIMIC-T5}  \\  \end{tabular} &
        \begin{tabular}{c} {SciFive+MIMIC-T5}  \\  \end{tabular} \\ \midrule

    {T5-Sup}
    & - & - & - & - \\ \midrule 
    {FLAN-T5} 
    & t(df=4)=1.02, \textit{P}=.37 & - & - & - \\ \midrule 
    {MIMIC-T5} 
    & t(df=4)=-0.07, \textit{P}=.95 
    & t(df=4)=-0.63, \textit{P}=.56  & - & - \\ \midrule 
    {SciFive+MIMIC-T5} 
    & t(df=4)=-0.34, \textit{P}=.75 
    & t(df=4)=-1.48, \textit{P}=.21 
    & t(df=4)=-0.23, \textit{P}=.83  & - \\
    
            \bottomrule
            \end{tabular} }
        
        \adjustbox{max width=\linewidth}{%
        \begin{tabular}{ccccc}
        Hospital System-Compliance-F1-T-test \\
        \toprule
        Models &
        \begin{tabular}{c} {T5-Sup} \\  \end{tabular} &
        \begin{tabular}{c} {FLAN-T5} \\  \end{tabular} &
        \begin{tabular}{c}{MIMIC-T5}  \\  \end{tabular} &
        \begin{tabular}{c} {SciFive+MIMIC-T5}  \\  \end{tabular} \\ \midrule

    {T5-Sup}
    & - & - & - & - \\ \midrule 
    {FLAN-T5} 
    & t(df=4)=1.75, \textit{P}=.16 & - & - & - \\ \midrule 
    {MIMIC-T5} 
    & t(df=4)=-1.04, \textit{P}=.36 
    & t(df=4)=-1.52, \textit{P}=.20  & - & - \\ \midrule 
    {SciFive+MIMIC-T5} 
    & t(df=4)=0.52, \textit{P}=.63 
    & t(df=4)=-0.91, \textit{P}=.42 
    & t(df=4)=1.06, \textit{P}=.35  & - \\
    
            \bottomrule
            \end{tabular} }
        
        \adjustbox{max width=\linewidth}{%
        \begin{tabular}{ccccc}
        Hospital System-Descriptors-F1-T-test \\
        \toprule
        Models &
        \begin{tabular}{c} {T5-Sup} \\  \end{tabular} &
        \begin{tabular}{c} {FLAN-T5} \\  \end{tabular} &
        \begin{tabular}{c}{MIMIC-T5}  \\  \end{tabular} &
        \begin{tabular}{c} {SciFive+MIMIC-T5}  \\  \end{tabular} \\ \midrule

    {T5-Sup}
    & - & - & - & - \\ \midrule 
    {FLAN-T5} 
    & t(df=4)=-0.99, \textit{P}=.38 & - & - & - \\ \midrule 
    {MIMIC-T5} 
    & t(df=4)=-2.5, \textit{P}=.07 
    & t(df=4)=0.6, \textit{P}=.58  & - & - \\ \midrule 
    {SciFive+MIMIC-T5} 
    & t(df=4)=-0.63, \textit{P}=.57 
    & t(df=4)=0.9, \textit{P}=.42 
    & t(df=4)=1.71, \textit{P}=.16  & - \\
    
            \bottomrule
            \end{tabular} }
\caption{5-fold cross-validation results across models on stigmatizing dataset from Hospital System and the t-test results among T5-Sup, FLAN-T5, {MIMIC-T5}, and {SciFive+MIMIC-T5} \label{tab:internal_stigma_cv}}

\end{table*}

\begin{table*}
\centering
\adjustbox{max width=\linewidth}{%
            \begin{tabular}{p{15mm}ccccc}
            \multicolumn{5}{l}{MIMIC-IV-Credibility \& Obstinacy - 25\% downsampling data} \\
            \toprule
            Metrics &
            ~ &
            \begin{tabular}{c}{T5-Sup}  \\    \end{tabular} &
            \begin{tabular}{c}{FLAN-T5}  \\    \end{tabular} &
            \begin{tabular}{c}{MIMIC-T5}  \\       \end{tabular} &
            \begin{tabular}{c}{SciFive+MIMIC-T5} \\        \end{tabular} \\ \midrule      
        
    \multirow{3}{*}{F1} &
    \multirow{2}{*}{Mean} &
    $63.81$ &
    $71.92$ &
    $64.44$ &
    $57.67$ \\ 
    ~ &
    &
    $\clubsuit$ &
    $\blacklozenge  $ & 
    $\uparrow \clubsuit 0.63$, $\downarrow \blacklozenge  7.48$  &
    $\downarrow \clubsuit 6.15$, $\downarrow \blacklozenge  14.25$ \\
    \cmidrule{2-6}        
    ~ &
    95\% CI &
    [$62.15, 65.77$] &
    [$69.6, 74.16$] &
    [$61.69, 66.63$] &
    [$54.39, 61.06$] \\
    
        \bottomrule 
        \end{tabular}} 
        
        \adjustbox{max width=\linewidth}{%
            \begin{tabular}{p{15mm}ccccc}
            \multicolumn{5}{l}{MIMIC-IV-Compliance- 25\% downsampling data} \\
            \toprule
            Metrics &
            ~ &
            \begin{tabular}{c}{T5-Sup}  \\    \end{tabular} &
            \begin{tabular}{c}{FLAN-T5}  \\    \end{tabular} &
            \begin{tabular}{c}{MIMIC-T5}  \\       \end{tabular} &
            \begin{tabular}{c}{SciFive+MIMIC-T5} \\        \end{tabular} \\ \midrule      
        
    \multirow{3}{*}{F1} &
    \multirow{2}{*}{Mean} &
    $89.6$ &
    $91.44$ &
    $86.02$ &
    $89.58$ \\ 
    ~ &
    &
    $\clubsuit$ &
    $\blacklozenge  $ & 
    $\downarrow \clubsuit 3.58$, $\downarrow \blacklozenge  5.42$  &
    $\downarrow \clubsuit 0.02$, $\downarrow \blacklozenge  1.86$ \\
    \cmidrule{2-6}        
    ~ &
    95\% CI &
    [$86.8, 92.19$] &
    [$89.15, 93.09$] &
    [$84.63, 87.5$] &
    [$87.19, 92.12$] \\
    
        \bottomrule 
        \end{tabular}} 
        
        \adjustbox{max width=\linewidth}{%
            \begin{tabular}{p{15mm}ccccc}
            \multicolumn{5}{l}{MIMIC-IV-Descriptors- 25\% downsampling data} \\
            \toprule
            Metrics &
            ~ &
            \begin{tabular}{c}{T5-Sup}  \\    \end{tabular} &
            \begin{tabular}{c}{FLAN-T5}  \\    \end{tabular} &
            \begin{tabular}{c}{MIMIC-T5}  \\       \end{tabular} &
            \begin{tabular}{c}{SciFive+MIMIC-T5} \\        \end{tabular} \\ \midrule      
        
    \multirow{3}{*}{F1} &
    \multirow{2}{*}{Mean} &
    $76.83$ &
    $78.21$ &
    $75.31$ &
    $82.46$ \\ 
    ~ &
    &
    $\clubsuit$ &
    $\blacklozenge  $ & 
    $\downarrow \clubsuit 1.52$, $\downarrow \blacklozenge  2.9$  &
    $\uparrow \clubsuit 5.63$, $\uparrow \blacklozenge  4.25$ \\
    \cmidrule{2-6}        
    ~ &
    95\% CI &
    [$72.49, 79.4$] &
    [$77.29, 79.19$] &
    [$72.32, 77.52$] &
    [$82.02, 82.85$] \\
    
        \bottomrule 
        \end{tabular}} 
        
    \medskip
    
        \adjustbox{max width=\linewidth}{%
        \begin{tabular}{ccccc}
        MIMIC-IV-Credibility \& Obstinacy-F1-T-test \\
        \toprule
        Models &
        \begin{tabular}{c} {T5-Sup} \\  \end{tabular} &
        \begin{tabular}{c} {FLAN-T5} \\  \end{tabular} &
        \begin{tabular}{c}{MIMIC-T5}  \\  \end{tabular} &
        \begin{tabular}{c} {SciFive+MIMIC-T5}  \\  \end{tabular} \\ \midrule

    {T5-Sup}
    & - & - & - & - \\ \midrule 
    {FLAN-T5} 
    & t(df=4)=3.37, \textit{P}=.03 & - & - & - \\ \midrule 
    {MIMIC-T5} 
    & t(df=4)=0.42, \textit{P}=.69 
    & t(df=4)=-2.66, \textit{P}=.06  & - & - \\ \midrule 
    {SciFive+MIMIC-T5} 
    & t(df=4)=-3.08, \textit{P}=.04 
    & t(df=4)=-4.9, \textit{P}=.008 
    & t(df=4)=-2.42, \textit{P}=.07  & - \\
    
            \bottomrule
            \end{tabular} }
        
        \adjustbox{max width=\linewidth}{%
        \begin{tabular}{ccccc}
        MIMIC-IV-Compliance-F1-T-test \\
        \toprule
        Models &
        \begin{tabular}{c} {T5-Sup} \\  \end{tabular} &
        \begin{tabular}{c} {FLAN-T5} \\  \end{tabular} &
        \begin{tabular}{c}{MIMIC-T5}  \\  \end{tabular} &
        \begin{tabular}{c} {SciFive+MIMIC-T5}  \\  \end{tabular} \\ \midrule

    {T5-Sup}
    & - & - & - & - \\ \midrule 
    {FLAN-T5} 
    & t(df=4)=1.2, \textit{P}=.30 & - & - & - \\ \midrule 
    {MIMIC-T5} 
    & t(df=4)=-4.16, \textit{P}=.01 
    & t(df=4)=-4.54, \textit{P}=.01  & - & - \\ \midrule 
    {SciFive+MIMIC-T5} 
    & t(df=4)=-0.01, \textit{P}=.99 
    & t(df=4)=-1.47, \textit{P}=.21 
    & t(df=4)=1.9, \textit{P}=.13  & - \\
    
            \bottomrule
            \end{tabular} }
        
        \adjustbox{max width=\linewidth}{%
        \begin{tabular}{ccccc}
        MIMIC-IV-Descriptors-F1-T-test \\
        \toprule
        Models &
        \begin{tabular}{c} {T5-Sup} \\  \end{tabular} &
        \begin{tabular}{c} {FLAN-T5} \\  \end{tabular} &
        \begin{tabular}{c}{MIMIC-T5}  \\  \end{tabular} &
        \begin{tabular}{c} {SciFive+MIMIC-T5}  \\  \end{tabular} \\ \midrule

    {T5-Sup}
    & - & - & - & - \\ \midrule 
    {FLAN-T5} 
    & t(df=4)=0.72, \textit{P}=.51 & - & - & - \\ \midrule 
    {MIMIC-T5} 
    & t(df=4)=-0.82, \textit{P}=.46 
    & t(df=4)=-1.67, \textit{P}=.17  & - & - \\ \midrule 
    {SciFive+MIMIC-T5} 
    & t(df=4)=2.66, \textit{P}=.06 
    & t(df=4)=5.61, \textit{P}=.005 
    & t(df=4)=4.64, \textit{P}=.010  & - \\
    
            \bottomrule
            \end{tabular} }
\caption{5-fold cross-validation results across models on stigmatizing dataset from MIMIC-IV with 25\% downsampled training data and the t-test results among T5-Sup, FLAN-T5, {MIMIC-T5}, and {SciFive+MIMIC-T5} \label{tab:mimic_25_cv}}
\end{table*}

\begin{table*}
\centering
\adjustbox{max width=\linewidth}{%
            \begin{tabular}{p{15mm}ccccc}
            \multicolumn{5}{l}{MIMIC-IV-Credibility \& Obstinacy - 5\% downsampling data} \\
            \toprule
            Metrics &
            ~ &
            \begin{tabular}{c}{T5-Sup}  \\    \end{tabular} &
            \begin{tabular}{c}{FLAN-T5}  \\    \end{tabular} &
            \begin{tabular}{c}{MIMIC-T5}  \\       \end{tabular} &
            \begin{tabular}{c}{SciFive+MIMIC-T5} \\        \end{tabular} \\ \midrule      
        
    \multirow{3}{*}{F1} &
    \multirow{2}{*}{Mean} &
    $51.62$ &
    $66.33$ &
    $58.58$ &
    $53.14$ \\ 
    ~ &
    &
    $\clubsuit$ &
    $\blacklozenge  $ & 
    $\uparrow \clubsuit 6.96$, $\downarrow \blacklozenge  7.76$  &
    $\uparrow \clubsuit 1.52$, $\downarrow \blacklozenge  13.2$ \\
    \cmidrule{2-6}        
    ~ &
    95\% CI &
    [$42.57, 58.72$] &
    [$64.37, 67.61$] &
    [$53.25, 64.66$] &
    [$49.77, 56.59$] \\
    
        \bottomrule 
        \end{tabular}} 
        
        \adjustbox{max width=\linewidth}{%
            \begin{tabular}{p{15mm}ccccc}
            \multicolumn{5}{l}{MIMIC-IV-Compliance- 5\% downsampling data} \\
            \toprule
            Metrics &
            ~ &
            \begin{tabular}{c}{T5-Sup}  \\    \end{tabular} &
            \begin{tabular}{c}{FLAN-T5}  \\    \end{tabular} &
            \begin{tabular}{c}{MIMIC-T5}  \\       \end{tabular} &
            \begin{tabular}{c}{SciFive+MIMIC-T5} \\        \end{tabular} \\ \midrule      
        
    \multirow{3}{*}{F1} &
    \multirow{2}{*}{Mean} &
    $70.2$ &
    $82.33$ &
    $72.98$ &
    $69.36$ \\ 
    ~ &
    &
    $\clubsuit$ &
    $\blacklozenge  $ & 
    $\uparrow \clubsuit 2.78$, $\downarrow \blacklozenge  9.35$  &
    $\downarrow \clubsuit 0.84$, $\downarrow \blacklozenge  12.97$ \\
    \cmidrule{2-6}        
    ~ &
    95\% CI &
    [$67.48, 72.75$] &
    [$74.22, 87.16$] &
    [$70.25, 76.61$] &
    [$55.76, 79.52$] \\
    
        \bottomrule 
        \end{tabular}} 
        
        \adjustbox{max width=\linewidth}{%
            \begin{tabular}{p{15mm}ccccc}
            \multicolumn{5}{l}{MIMIC-IV-Descriptors- 5\% downsampling data} \\
            \toprule
            Metrics &
            ~ &
            \begin{tabular}{c}{T5-Sup}  \\    \end{tabular} &
            \begin{tabular}{c}{FLAN-T5}  \\    \end{tabular} &
            \begin{tabular}{c}{MIMIC-T5}  \\       \end{tabular} &
            \begin{tabular}{c}{SciFive+MIMIC-T5} \\        \end{tabular} \\ \midrule      
        
    \multirow{3}{*}{F1} &
    \multirow{2}{*}{Mean} &
    $61.5$ &
    $63.16$ &
    $62.5$ &
    $67.3$ \\ 
    ~ &
    &
    $\clubsuit$ &
    $\blacklozenge  $ & 
    $\uparrow \clubsuit 0.99$, $\downarrow \blacklozenge  0.66$  &
    $\uparrow \clubsuit 5.8$, $\uparrow \blacklozenge  4.14$ \\
    \cmidrule{2-6}        
    ~ &
    95\% CI &
    [$59.2, 63.54$] &
    [$49.6, 71.45$] &
    [$60.26, 64.6$] &
    [$64.92, 69.49$] \\
    
        \bottomrule 
        \end{tabular}} 
        
    \medskip
    
        \adjustbox{max width=\linewidth}{%
        \begin{tabular}{ccccc}
        MIMIC-IV-Credibility \& Obstinacy-F1-T-test \\
        \toprule
        Models &
        \begin{tabular}{c} {T5-Sup} \\  \end{tabular} &
        \begin{tabular}{c} {FLAN-T5} \\  \end{tabular} &
        \begin{tabular}{c}{MIMIC-T5}  \\  \end{tabular} &
        \begin{tabular}{c} {SciFive+MIMIC-T5}  \\  \end{tabular} \\ \midrule

    {T5-Sup}
    & - & - & - & - \\ \midrule 
    {FLAN-T5} 
    & t(df=4)=2.78, \textit{P}=.05 & - & - & - \\ \midrule 
    {MIMIC-T5} 
    & t(df=4)=1.35, \textit{P}=.25 
    & t(df=4)=-2.55, \textit{P}=.06  & - & - \\ \midrule 
    {SciFive+MIMIC-T5} 
    & t(df=4)=0.27, \textit{P}=.80 
    & t(df=4)=-5.36, \textit{P}=.006 
    & t(df=4)=-1.25, \textit{P}=.28  & - \\
    
            \bottomrule
            \end{tabular} }
        
        \adjustbox{max width=\linewidth}{%
        \begin{tabular}{ccccc}
        MIMIC-IV-Compliance-F1-T-test \\
        \toprule
        Models &
        \begin{tabular}{c} {T5-Sup} \\  \end{tabular} &
        \begin{tabular}{c} {FLAN-T5} \\  \end{tabular} &
        \begin{tabular}{c}{MIMIC-T5}  \\  \end{tabular} &
        \begin{tabular}{c} {SciFive+MIMIC-T5}  \\  \end{tabular} \\ \midrule

    {T5-Sup}
    & - & - & - & - \\ \midrule 
    {FLAN-T5} 
    & t(df=4)=2.4, \textit{P}=.07 & - & - & - \\ \midrule 
    {MIMIC-T5} 
    & t(df=4)=0.93, \textit{P}=.40 
    & t(df=4)=-1.83, \textit{P}=.14  & - & - \\ \midrule 
    {SciFive+MIMIC-T5} 
    & t(df=4)=-0.11, \textit{P}=.92 
    & t(df=4)=-1.38, \textit{P}=.24 
    & t(df=4)=-0.36, \textit{P}=.73  & - \\
    
            \bottomrule
            \end{tabular} }
        
        \adjustbox{max width=\linewidth}{%
        \begin{tabular}{ccccc}
        MIMIC-IV-Descriptors-F1-T-test \\
        \toprule
        Models &
        \begin{tabular}{c} {T5-Sup} \\  \end{tabular} &
        \begin{tabular}{c} {FLAN-T5} \\  \end{tabular} &
        \begin{tabular}{c}{MIMIC-T5}  \\  \end{tabular} &
        \begin{tabular}{c} {SciFive+MIMIC-T5}  \\  \end{tabular} \\ \midrule

    {T5-Sup}
    & - & - & - & - \\ \midrule 
    {FLAN-T5} 
    & t(df=4)=0.25, \textit{P}=.81 & - & - & - \\ \midrule 
    {MIMIC-T5} 
    & t(df=4)=1.7, \textit{P}=.16 
    & t(df=4)=-0.11, \textit{P}=.92  & - & - \\ \midrule 
    {SciFive+MIMIC-T5} 
    & t(df=4)=9.13, \textit{P}=<.001 
    & t(df=4)=0.58, \textit{P}=.59 
    & t(df=4)=5.05, \textit{P}=.007  & - \\
    
            \bottomrule
            \end{tabular} }
\caption{5-fold cross-validation results across models on stigmatizing dataset from MIMIC-IV with 5\% downsampled training data and the t-test results among T5-Sup, FLAN-T5, {MIMIC-T5}, and {SciFive+MIMIC-T5} \label{tab:mimic_5_cv}}

\end{table*}

\begin{table*}
\centering

\adjustbox{max width=\linewidth}{%
            \begin{tabular}{p{15mm}ccccc}
            \multicolumn{5}{l}{MIMIC-IV-Credibility \& Obstinacy - 1\% downsampling data} \\
            \toprule
            Metrics &
            ~ &
            \begin{tabular}{c}{T5-Sup}  \\    \end{tabular} &
            \begin{tabular}{c}{FLAN-T5}  \\    \end{tabular} &
            \begin{tabular}{c}{MIMIC-T5}  \\       \end{tabular} &
            \begin{tabular}{c}{SciFive+MIMIC-T5} \\        \end{tabular} \\ \midrule      
        
    \multirow{3}{*}{F1} &
    \multirow{2}{*}{Mean} &
    $27.7$ &
    $51.02$ &
    $37.08$ &
    $33.49$ \\ 
    ~ &
    &
    $\clubsuit$ &
    $\blacklozenge  $ & 
    $\uparrow \clubsuit 9.38$, $\downarrow \blacklozenge  13.94$  &
    $\uparrow \clubsuit 5.78$, $\downarrow \blacklozenge  17.54$ \\
    \cmidrule{2-6}        
    ~ &
    95\% CI &
    [$22.94, 32.17$] &
    [$45.89, 55.52$] &
    [$29.74, 42.96$] &
    [$27.39, 40.18$] \\
    
        \bottomrule 
        \end{tabular}} 
        
        \adjustbox{max width=\linewidth}{%
            \begin{tabular}{p{15mm}ccccc}
            \multicolumn{5}{l}{MIMIC-IV-Compliance- 1\% downsampling data} \\
            \toprule
            Metrics &
            ~ &
            \begin{tabular}{c}{T5-Sup}  \\    \end{tabular} &
            \begin{tabular}{c}{FLAN-T5}  \\    \end{tabular} &
            \begin{tabular}{c}{MIMIC-T5}  \\       \end{tabular} &
            \begin{tabular}{c}{SciFive+MIMIC-T5} \\        \end{tabular} \\ \midrule      
        
    \multirow{3}{*}{F1} &
    \multirow{2}{*}{Mean} &
    $54.32$ &
    $66.18$ &
    $42.1$ &
    $50.44$ \\ 
    ~ &
    &
    $\clubsuit$ &
    $\blacklozenge  $ & 
    $\downarrow \clubsuit 12.22$, $\downarrow \blacklozenge  24.09$  &
    $\downarrow \clubsuit 3.88$, $\downarrow \blacklozenge  15.75$ \\
    \cmidrule{2-6}        
    ~ &
    95\% CI &
    [$49.27, 59.91$] &
    [$65.28, 66.94$] &
    [$38.57, 46.91$] &
    [$43.57, 56.06$] \\
    
        \bottomrule 
        \end{tabular}} 
        
        \adjustbox{max width=\linewidth}{%
            \begin{tabular}{p{15mm}ccccc}
            \multicolumn{5}{l}{MIMIC-IV-Descriptors- 1\% downsampling data} \\
            \toprule
            Metrics &
            ~ &
            \begin{tabular}{c}{T5-Sup}  \\    \end{tabular} &
            \begin{tabular}{c}{FLAN-T5}  \\    \end{tabular} &
            \begin{tabular}{c}{MIMIC-T5}  \\       \end{tabular} &
            \begin{tabular}{c}{SciFive+MIMIC-T5} \\        \end{tabular} \\ \midrule      
        
    \multirow{3}{*}{F1} &
    \multirow{2}{*}{Mean} &
    $43.41$ &
    $47.85$ &
    $29.09$ &
    $35.14$ \\ 
    ~ &
    &
    $\clubsuit$ &
    $\blacklozenge  $ & 
    $\downarrow \clubsuit 14.32$, $\downarrow \blacklozenge  18.75$  &
    $\downarrow \clubsuit 8.27$, $\downarrow \blacklozenge  12.7$ \\
    \cmidrule{2-6}        
    ~ &
    95\% CI &
    [$40.3, 46.32$] &
    [$42.16, 55.4$] &
    [$25.76, 33.03$] &
    [$29.11, 41.98$] \\
    
        \bottomrule 
        \end{tabular}} 
        
    \medskip
    
        \adjustbox{max width=\linewidth}{%
        \begin{tabular}{ccccc}
        MIMIC-IV-Credibility \& Obstinacy-F1-T-test \\
        \toprule
        Models &
        \begin{tabular}{c} {T5-Sup} \\  \end{tabular} &
        \begin{tabular}{c} {FLAN-T5} \\  \end{tabular} &
        \begin{tabular}{c}{MIMIC-T5}  \\  \end{tabular} &
        \begin{tabular}{c} {SciFive+MIMIC-T5}  \\  \end{tabular} \\ \midrule

    {T5-Sup}
    & - & - & - & - \\ \midrule 
    {FLAN-T5} 
    & t(df=4)=5.92, \textit{P}=.004 & - & - & - \\ \midrule 
    {MIMIC-T5} 
    & t(df=4)=3.43, \textit{P}=.03 
    & t(df=4)=-2.43, \textit{P}=.07  & - & - \\ \midrule 
    {SciFive+MIMIC-T5} 
    & t(df=4)=1.77, \textit{P}=.15 
    & t(df=4)=-3.58, \textit{P}=.02 
    & t(df=4)=-1.76, \textit{P}=.15  & - \\
    
            \bottomrule
            \end{tabular} }
        
        \adjustbox{max width=\linewidth}{%
        \begin{tabular}{ccccc}
        MIMIC-IV-Compliance-F1-T-test \\
        \toprule
        Models &
        \begin{tabular}{c} {T5-Sup} \\  \end{tabular} &
        \begin{tabular}{c} {FLAN-T5} \\  \end{tabular} &
        \begin{tabular}{c}{MIMIC-T5}  \\  \end{tabular} &
        \begin{tabular}{c} {SciFive+MIMIC-T5}  \\  \end{tabular} \\ \midrule

    {T5-Sup}
    & - & - & - & - \\ \midrule 
    {FLAN-T5} 
    & t(df=4)=3.46, \textit{P}=.03 & - & - & - \\ \midrule 
    {MIMIC-T5} 
    & t(df=4)=-8.5, \textit{P}=.001 
    & t(df=4)=-9.03, \textit{P}=<.001  & - & - \\ \midrule 
    {SciFive+MIMIC-T5} 
    & t(df=4)=-1.14, \textit{P}=.32 
    & t(df=4)=-4.2, \textit{P}=.01 
    & t(df=4)=2.98, \textit{P}=.04  & - \\
    
            \bottomrule
            \end{tabular} }
        
        \adjustbox{max width=\linewidth}{%
        \begin{tabular}{ccccc}
        MIMIC-IV-Descriptors-F1-T-test \\
        \toprule
        Models &
        \begin{tabular}{c} {T5-Sup} \\  \end{tabular} &
        \begin{tabular}{c} {FLAN-T5} \\  \end{tabular} &
        \begin{tabular}{c}{MIMIC-T5}  \\  \end{tabular} &
        \begin{tabular}{c} {SciFive+MIMIC-T5}  \\  \end{tabular} \\ \midrule

    {T5-Sup}
    & - & - & - & - \\ \midrule 
    {FLAN-T5} 
    & t(df=4)=1.23, \textit{P}=.29 & - & - & - \\ \midrule 
    {MIMIC-T5} 
    & t(df=4)=-3.94, \textit{P}=.02 
    & t(df=4)=-4.89, \textit{P}=.008  & - & - \\ \midrule 
    {SciFive+MIMIC-T5} 
    & t(df=4)=-1.63, \textit{P}=.18 
    & t(df=4)=-2.06, \textit{P}=.11 
    & t(df=4)=2.25, \textit{P}=.09  & - \\
    
            \bottomrule
            \end{tabular} }
 \caption{5-fold cross-validation results across models on stigmatizing dataset from MIMIC-IV with 1\% downsampled training data and the t-test results among T5-Sup, FLAN-T5, {MIMIC-T5}, and {SciFive+MIMIC-T5} \label{tab:mimic_1_cv}}
\end{table*}

\begin{table*}
\centering
\adjustbox{max width=\linewidth}{%
            \begin{tabular}{p{15mm}ccccc}
            \multicolumn{5}{l}{Hospital System-Credibility \& Obstinacy - 25\% downsampling data} \\
            \toprule
            Metrics &
            ~ &
            \begin{tabular}{c}{T5-Sup}  \\    \end{tabular} &
            \begin{tabular}{c}{FLAN-T5}  \\    \end{tabular} &
            \begin{tabular}{c}{MIMIC-T5}  \\       \end{tabular} &
            \begin{tabular}{c}{SciFive+MIMIC-T5} \\        \end{tabular} \\ \midrule      
        
    \multirow{3}{*}{F1} &
    \multirow{2}{*}{Mean} &
    $85.1$ &
    $86.59$ &
    $75.9$ &
    $84.7$ \\ 
    ~ &
    &
    $\clubsuit$ &
    $\blacklozenge  $ & 
    $\downarrow \clubsuit 9.19$, $\downarrow \blacklozenge  10.69$  &
    $\downarrow \clubsuit 0.4$, $\downarrow \blacklozenge  1.89$ \\
    \cmidrule{2-6}        
    ~ &
    95\% CI &
    [$81.26, 87.88$] &
    [$83.57, 88.98$] &
    [$71.71, 80.43$] &
    [$80.58, 88.32$] \\
    
        \bottomrule 
        \end{tabular}} 
        
        \adjustbox{max width=\linewidth}{%
            \begin{tabular}{p{15mm}ccccc}
            \multicolumn{5}{l}{Hospital System-Compliance- 25\% downsampling data} \\
            \toprule
            Metrics &
            ~ &
            \begin{tabular}{c}{T5-Sup}  \\    \end{tabular} &
            \begin{tabular}{c}{FLAN-T5}  \\    \end{tabular} &
            \begin{tabular}{c}{MIMIC-T5}  \\       \end{tabular} &
            \begin{tabular}{c}{SciFive+MIMIC-T5} \\        \end{tabular} \\ \midrule      
        
    \multirow{3}{*}{F1} &
    \multirow{2}{*}{Mean} &
    $82.52$ &
    $85.3$ &
    $79.32$ &
    $81.53$ \\ 
    ~ &
    &
    $\clubsuit$ &
    $\blacklozenge  $ & 
    $\downarrow \clubsuit 3.2$, $\downarrow \blacklozenge  5.98$  &
    $\downarrow \clubsuit 0.99$, $\downarrow \blacklozenge  3.77$ \\
    \cmidrule{2-6}        
    ~ &
    95\% CI &
    [$78.41, 84.92$] &
    [$82.98, 87.65$] &
    [$76.29, 83.15$] &
    [$75.26, 86.29$] \\
    
        \bottomrule 
        \end{tabular}} 
        
        \adjustbox{max width=\linewidth}{%
            \begin{tabular}{p{15mm}ccccc}
            \multicolumn{5}{l}{Hospital System-Descriptors- 25\% downsampling data} \\
            \toprule
            Metrics &
            ~ &
            \begin{tabular}{c}{T5-Sup}  \\    \end{tabular} &
            \begin{tabular}{c}{FLAN-T5}  \\    \end{tabular} &
            \begin{tabular}{c}{MIMIC-T5}  \\       \end{tabular} &
            \begin{tabular}{c}{SciFive+MIMIC-T5} \\        \end{tabular} \\ \midrule      
        
    \multirow{3}{*}{F1} &
    \multirow{2}{*}{Mean} &
    $86.45$ &
    $85.98$ &
    $87.19$ &
    $86.51$ \\ 
    ~ &
    &
    $\clubsuit$ &
    $\blacklozenge  $ & 
    $\uparrow \clubsuit 0.75$, $\uparrow \blacklozenge  1.22$  &
    $\uparrow \clubsuit 0.06$, $\uparrow \blacklozenge  0.53$ \\
    \cmidrule{2-6}        
    ~ &
    95\% CI &
    [$85.24, 88.16$] &
    [$85.14, 86.69$] &
    [$85.32, 89.41$] &
    [$85.09, 88.14$] \\
    
        \bottomrule 
        \end{tabular}} 
        
    \medskip
    
        \adjustbox{max width=\linewidth}{%
        \begin{tabular}{ccccc}
        Hospital System-Credibility \& Obstinacy-F1-T-test \\
        \toprule
        Models &
        \begin{tabular}{c} {T5-Sup} \\  \end{tabular} &
        \begin{tabular}{c} {FLAN-T5} \\  \end{tabular} &
        \begin{tabular}{c}{MIMIC-T5}  \\  \end{tabular} &
        \begin{tabular}{c} {SciFive+MIMIC-T5}  \\  \end{tabular} \\ \midrule

    {T5-Sup}
    & - & - & - & - \\ \midrule 
    {FLAN-T5} 
    & t(df=4)=0.54, \textit{P}=.62 & - & - & - \\ \midrule 
    {MIMIC-T5} 
    & t(df=4)=-2.36, \textit{P}=.08 
    & t(df=4)=-2.94, \textit{P}=.04  & - & - \\ \midrule 
    {SciFive+MIMIC-T5} 
    & t(df=4)=-0.15, \textit{P}=.89 
    & t(df=4)=-0.54, \textit{P}=.62 
    & t(df=4)=4.8, \textit{P}=.009  & - \\
    
            \bottomrule
            \end{tabular} }
        
        \adjustbox{max width=\linewidth}{%
        \begin{tabular}{ccccc}
        Hospital System-Compliance-F1-T-test \\
        \toprule
        Models &
        \begin{tabular}{c} {T5-Sup} \\  \end{tabular} &
        \begin{tabular}{c} {FLAN-T5} \\  \end{tabular} &
        \begin{tabular}{c}{MIMIC-T5}  \\  \end{tabular} &
        \begin{tabular}{c} {SciFive+MIMIC-T5}  \\  \end{tabular} \\ \midrule

    {T5-Sup}
    & - & - & - & - \\ \midrule 
    {FLAN-T5} 
    & t(df=4)=1.79, \textit{P}=.15 & - & - & - \\ \midrule 
    {MIMIC-T5} 
    & t(df=4)=-1.57, \textit{P}=.19 
    & t(df=4)=-2.98, \textit{P}=.04  & - & - \\ \midrule 
    {SciFive+MIMIC-T5} 
    & t(df=4)=-0.78, \textit{P}=.48 
    & t(df=4)=-1.37, \textit{P}=.24 
    & t(df=4)=0.88, \textit{P}=.43  & - \\
    
            \bottomrule
            \end{tabular} }
        
        \adjustbox{max width=\linewidth}{%
        \begin{tabular}{ccccc}
        Hospital System-Descriptors-F1-T-test \\
        \toprule
        Models &
        \begin{tabular}{c} {T5-Sup} \\  \end{tabular} &
        \begin{tabular}{c} {FLAN-T5} \\  \end{tabular} &
        \begin{tabular}{c}{MIMIC-T5}  \\  \end{tabular} &
        \begin{tabular}{c} {SciFive+MIMIC-T5}  \\  \end{tabular} \\ \midrule

    {T5-Sup}
    & - & - & - & - \\ \midrule 
    {FLAN-T5} 
    & t(df=4)=-0.43, \textit{P}=.69 & - & - & - \\ \midrule 
    {MIMIC-T5} 
    & t(df=4)=0.53, \textit{P}=.62 
    & t(df=4)=1.05, \textit{P}=.35  & - & - \\ \midrule 
    {SciFive+MIMIC-T5} 
    & t(df=4)=0.08, \textit{P}=.94 
    & t(df=4)=0.49, \textit{P}=.65 
    & t(df=4)=-0.5, \textit{P}=.65  & - \\
    
            \bottomrule
            \end{tabular} }
\caption{5-fold cross-validation results across models on stigmatizing dataset from Hospital System with 25\% downsampled training data and the t-test results among T5-Sup, FLAN-T5, {MIMIC-T5}, and {SciFive+MIMIC-T5} \label{tab:internal_25_cv}}
        
\end{table*}

\begin{table*}
\centering

\adjustbox{max width=\linewidth}{%
            \begin{tabular}{p{15mm}ccccc}
            \multicolumn{5}{l}{Hospital System-Credibility \& Obstinacy - 5\% downsampling data} \\
            \toprule
            Metrics &
            ~ &
            \begin{tabular}{c}{T5-Sup}  \\    \end{tabular} &
            \begin{tabular}{c}{FLAN-T5}  \\    \end{tabular} &
            \begin{tabular}{c}{MIMIC-T5}  \\       \end{tabular} &
            \begin{tabular}{c}{SciFive+MIMIC-T5} \\        \end{tabular} \\ \midrule      
        
    \multirow{3}{*}{F1} &
    \multirow{2}{*}{Mean} &
    $48.94$ &
    $75.66$ &
    $48.88$ &
    $50.2$ \\ 
    ~ &
    &
    $\clubsuit$ &
    $\blacklozenge  $ & 
    $\downarrow \clubsuit 0.06$, $\downarrow \blacklozenge  26.78$  &
    $\uparrow \clubsuit 1.26$, $\downarrow \blacklozenge  25.46$ \\
    \cmidrule{2-6}        
    ~ &
    95\% CI &
    [$43.03, 55.34$] &
    [$67.77, 82.11$] &
    [$46.72, 50.4$] &
    [$48.32, 52.7$] \\
    
        \bottomrule 
        \end{tabular}} 
        
        \adjustbox{max width=\linewidth}{%
            \begin{tabular}{p{15mm}ccccc}
            \multicolumn{5}{l}{Hospital System - Compliance - 5\% downsampling data} \\
            \toprule
            Metrics &
            ~ &
            \begin{tabular}{c}{T5-Sup}  \\    \end{tabular} &
            \begin{tabular}{c}{FLAN-T5}  \\    \end{tabular} &
            \begin{tabular}{c}{MIMIC-T5}  \\       \end{tabular} &
            \begin{tabular}{c}{SciFive+MIMIC-T5} \\        \end{tabular} \\ \midrule      
        
    \multirow{3}{*}{F1} &
    \multirow{2}{*}{Mean} &
    $58.62$ &
    $71.09$ &
    $51.82$ &
    $56.84$ \\ 
    ~ &
    &
    $\clubsuit$ &
    $\blacklozenge  $ & 
    $\downarrow \clubsuit 6.8$, $\downarrow \blacklozenge  19.27$  &
    $\downarrow \clubsuit 1.78$, $\downarrow \blacklozenge  14.25$ \\
    \cmidrule{2-6}        
    ~ &
    95\% CI &
    [$55.48, 60.66$] &
    [$63.96, 76.14$] &
    [$49.19, 54.12$] &
    [$52.65, 60.79$] \\
    
        \bottomrule 
        \end{tabular}} 
        
        \adjustbox{max width=\linewidth}{%
            \begin{tabular}{p{15mm}ccccc}
            \multicolumn{5}{l}{Hospital System - Descriptors - 5\% downsampling data} \\
            \toprule
            Metrics &
            ~ &
            \begin{tabular}{c}{T5-Sup}  \\    \end{tabular} &
            \begin{tabular}{c}{FLAN-T5}  \\    \end{tabular} &
            \begin{tabular}{c}{MIMIC-T5}  \\       \end{tabular} &
            \begin{tabular}{c}{SciFive+MIMIC-T5} \\        \end{tabular} \\ \midrule      
        
    \multirow{3}{*}{F1} &
    \multirow{2}{*}{Mean} &
    $69.95$ &
    $79.82$ &
    $69.71$ &
    $76.41$ \\ 
    ~ &
    &
    $\clubsuit$ &
    $\blacklozenge  $ & 
    $\downarrow \clubsuit 0.24$, $\downarrow \blacklozenge  10.11$  &
    $\uparrow \clubsuit 6.46$, $\downarrow \blacklozenge  3.41$ \\
    \cmidrule{2-6}        
    ~ &
    95\% CI &
    [$68.13, 71.33$] &
    [$77.17, 82.3$] &
    [$63.77, 73.81$] &
    [$75.0, 77.68$] \\
    
        \bottomrule 
        \end{tabular}} 
        
    \medskip
    
        \adjustbox{max width=\linewidth}{%
        \begin{tabular}{ccccc}
        Hospital System-Credibility \& Obstinacy-F1-T-test \\
        \toprule
        Models &
        \begin{tabular}{c} {T5-Sup} \\  \end{tabular} &
        \begin{tabular}{c} {FLAN-T5} \\  \end{tabular} &
        \begin{tabular}{c}{MIMIC-T5}  \\  \end{tabular} &
        \begin{tabular}{c} {SciFive+MIMIC-T5}  \\  \end{tabular} \\ \midrule

    {T5-Sup}
    & - & - & - & - \\ \midrule 
    {FLAN-T5} 
    & t(df=4)=3.89, \textit{P}=.02 & - & - & - \\ \midrule 
    {MIMIC-T5} 
    & t(df=4)=-0.02, \textit{P}=.99 
    & t(df=4)=-5.36, \textit{P}=.006  & - & - \\ \midrule 
    {SciFive+MIMIC-T5} 
    & t(df=4)=0.42, \textit{P}=.70 
    & t(df=4)=-4.42, \textit{P}=.01 
    & t(df=4)=0.93, \textit{P}=.40  & - \\
    
            \bottomrule
            \end{tabular} }
        
        \adjustbox{max width=\linewidth}{%
        \begin{tabular}{ccccc}
        Hospital System-Compliance-F1-T-test \\
        \toprule
        Models &
        \begin{tabular}{c} {T5-Sup} \\  \end{tabular} &
        \begin{tabular}{c} {FLAN-T5} \\  \end{tabular} &
        \begin{tabular}{c}{MIMIC-T5}  \\  \end{tabular} &
        \begin{tabular}{c} {SciFive+MIMIC-T5}  \\  \end{tabular} \\ \midrule

    {T5-Sup}
    & - & - & - & - \\ \midrule 
    {FLAN-T5} 
    & t(df=4)=2.52, \textit{P}=.07 & - & - & - \\ \midrule 
    {MIMIC-T5} 
    & t(df=4)=-4.8, \textit{P}=.009 
    & t(df=4)=-5.18, \textit{P}=.007  & - & - \\ \midrule 
    {SciFive+MIMIC-T5} 
    & t(df=4)=-1.17, \textit{P}=.31 
    & t(df=4)=-3.12, \textit{P}=.04 
    & t(df=4)=2.5, \textit{P}=.07  & - \\
    
            \bottomrule
            \end{tabular} }
        
        \adjustbox{max width=\linewidth}{%
        \begin{tabular}{ccccc}
        Hospital System-Descriptors-F1-T-test \\
        \toprule
        Models &
        \begin{tabular}{c} {T5-Sup} \\  \end{tabular} &
        \begin{tabular}{c} {FLAN-T5} \\  \end{tabular} &
        \begin{tabular}{c}{MIMIC-T5}  \\  \end{tabular} &
        \begin{tabular}{c} {SciFive+MIMIC-T5}  \\  \end{tabular} \\ \midrule

    {T5-Sup}
    & - & - & - & - \\ \midrule 
    {FLAN-T5} 
    & t(df=4)=6.01, \textit{P}=.004 & - & - & - \\ \midrule 
    {MIMIC-T5} 
    & t(df=4)=-0.07, \textit{P}=.95 
    & t(df=4)=-2.23, \textit{P}=.09  & - & - \\ \midrule 
    {SciFive+MIMIC-T5} 
    & t(df=4)=4.25, \textit{P}=.01 
    & t(df=4)=-3.41, \textit{P}=.03 
    & t(df=4)=1.57, \textit{P}=.19  & - \\
    
            \bottomrule
            \end{tabular} }
\caption{5-fold cross-validation results across models on stigmatizing dataset from Hospital System with 5\% downsampled training data and the t-test results among T5-Sup, FLAN-T5, {MIMIC-T5}, and {SciFive+MIMIC-T5} \label{tab:internal_5_cv}}
\end{table*}

\begin{table*}
\centering
\adjustbox{max width=\linewidth}{%
            \begin{tabular}{p{15mm}ccccc}
            \multicolumn{5}{l}{Hospital System-Credibility \& Obstinacy - 1\% downsampling data} \\
            \toprule
            Metrics &
            ~ &
            \begin{tabular}{c}{T5-Sup}  \\    \end{tabular} &
            \begin{tabular}{c}{FLAN-T5}  \\    \end{tabular} &
            \begin{tabular}{c}{MIMIC-T5}  \\       \end{tabular} &
            \begin{tabular}{c}{SciFive+MIMIC-T5} \\        \end{tabular} \\ \midrule      
        
    \multirow{3}{*}{F1} &
    \multirow{2}{*}{Mean} &
    $23.29$ &
    $54.45$ &
    $34.95$ &
    $31.79$ \\ 
    ~ &
    &
    $\clubsuit$ &
    $\blacklozenge  $ & 
    $\uparrow \clubsuit 11.66$, $\downarrow \blacklozenge  19.5$  &
    $\uparrow \clubsuit 8.5$, $\downarrow \blacklozenge  22.66$ \\
    \cmidrule{2-6}        
    ~ &
    95\% CI &
    [$17.55, 27.46$] &
    [$45.76, 61.91$] &
    [$26.12, 46.24$] &
    [$25.45, 38.43$] \\
    
        \bottomrule 
        \end{tabular}} 
        
        \adjustbox{max width=\linewidth}{%
            \begin{tabular}{p{15mm}ccccc}
            \multicolumn{5}{l}{Hospital System-Compliance- 1\% downsampling data} \\
            \toprule
            Metrics &
            ~ &
            \begin{tabular}{c}{T5-Sup}  \\    \end{tabular} &
            \begin{tabular}{c}{FLAN-T5}  \\    \end{tabular} &
            \begin{tabular}{c}{MIMIC-T5}  \\       \end{tabular} &
            \begin{tabular}{c}{SciFive+MIMIC-T5} \\        \end{tabular} \\ \midrule      
        
    \multirow{3}{*}{F1} &
    \multirow{2}{*}{Mean} &
    $45.44$ &
    $70.6$ &
    $41.92$ &
    $38.52$ \\ 
    ~ &
    &
    $\clubsuit$ &
    $\blacklozenge  $ & 
    $\downarrow \clubsuit 3.52$, $\downarrow \blacklozenge  28.68$  &
    $\downarrow \clubsuit 6.92$, $\downarrow \blacklozenge  32.08$ \\
    \cmidrule{2-6}        
    ~ &
    95\% CI &
    [$40.34, 52.88$] &
    [$68.81, 72.05$] &
    [$36.33, 46.04$] &
    [$33.42, 42.52$] \\
    
        \bottomrule 
        \end{tabular}} 
        
        \adjustbox{max width=\linewidth}{%
            \begin{tabular}{p{15mm}ccccc}
            \multicolumn{5}{l}{Hospital System-Descriptors- 1\% downsampling data} \\
            \toprule
            Metrics &
            ~ &
            \begin{tabular}{c}{T5-Sup}  \\    \end{tabular} &
            \begin{tabular}{c}{FLAN-T5}  \\    \end{tabular} &
            \begin{tabular}{c}{MIMIC-T5}  \\       \end{tabular} &
            \begin{tabular}{c}{SciFive+MIMIC-T5} \\        \end{tabular} \\ \midrule      
        
    \multirow{3}{*}{F1} &
    \multirow{2}{*}{Mean} &
    $53.78$ &
    $58.36$ &
    $44.52$ &
    $55.52$ \\ 
    ~ &
    &
    $\clubsuit$ &
    $\blacklozenge  $ & 
    $\downarrow \clubsuit 9.26$, $\downarrow \blacklozenge  13.84$  &
    $\uparrow \clubsuit 1.74$, $\downarrow \blacklozenge  2.83$ \\
    \cmidrule{2-6}        
    ~ &
    95\% CI &
    [$45.07, 60.86$] &
    [$52.25, 64.75$] &
    [$32.8, 54.06$] &
    [$53.29, 58.91$] \\
    
        \bottomrule 
        \end{tabular}} 
        
    \medskip
    
        \adjustbox{max width=\linewidth}{%
        \begin{tabular}{ccccc}
        Hospital System-Credibility \& Obstinacy-F1-T-test \\
        \toprule
        Models &
        \begin{tabular}{c} {T5-Sup} \\  \end{tabular} &
        \begin{tabular}{c} {FLAN-T5} \\  \end{tabular} &
        \begin{tabular}{c}{MIMIC-T5}  \\  \end{tabular} &
        \begin{tabular}{c} {SciFive+MIMIC-T5}  \\  \end{tabular} \\ \midrule

    {T5-Sup}
    & - & - & - & - \\ \midrule 
    {FLAN-T5} 
    & t(df=4)=4.77, \textit{P}=.009 & - & - & - \\ \midrule 
    {MIMIC-T5} 
    & t(df=4)=1.39, \textit{P}=.24 
    & t(df=4)=-5.22, \textit{P}=.006  & - & - \\ \midrule 
    {SciFive+MIMIC-T5} 
    & t(df=4)=1.49, \textit{P}=.21 
    & t(df=4)=-6.22, \textit{P}=.003 
    & t(df=4)=-0.56, \textit{P}=.61  & - \\
    
            \bottomrule
            \end{tabular} }
        
        \adjustbox{max width=\linewidth}{%
        \begin{tabular}{ccccc}
        Hospital System-Compliance-F1-T-test \\
        \toprule
        Models &
        \begin{tabular}{c} {T5-Sup} \\  \end{tabular} &
        \begin{tabular}{c} {FLAN-T5} \\  \end{tabular} &
        \begin{tabular}{c}{MIMIC-T5}  \\  \end{tabular} &
        \begin{tabular}{c} {SciFive+MIMIC-T5}  \\  \end{tabular} \\ \midrule

    {T5-Sup}
    & - & - & - & - \\ \midrule 
    {FLAN-T5} 
    & t(df=4)=6.2, \textit{P}=.003 & - & - & - \\ \midrule 
    {MIMIC-T5} 
    & t(df=4)=-0.85, \textit{P}=.44 
    & t(df=4)=-9.06, \textit{P}=<.001  & - & - \\ \midrule 
    {SciFive+MIMIC-T5} 
    & t(df=4)=-1.63, \textit{P}=.18 
    & t(df=4)=-14.93, \textit{P}=<.001 
    & t(df=4)=-1.02, \textit{P}=.36  & - \\
    
            \bottomrule
            \end{tabular} }
        
        \adjustbox{max width=\linewidth}{%
        \begin{tabular}{ccccc}
        Hospital System-Descriptors-F1-T-test \\
        \toprule
        Models &
        \begin{tabular}{c} {T5-Sup} \\  \end{tabular} &
        \begin{tabular}{c} {FLAN-T5} \\  \end{tabular} &
        \begin{tabular}{c}{MIMIC-T5}  \\  \end{tabular} &
        \begin{tabular}{c} {SciFive+MIMIC-T5}  \\  \end{tabular} \\ \midrule

    {T5-Sup}
    & - & - & - & - \\ \midrule 
    {FLAN-T5} 
    & t(df=4)=0.63, \textit{P}=.56 & - & - & - \\ \midrule 
    {MIMIC-T5} 
    & t(df=4)=-2.76, \textit{P}=.05 
    & t(df=4)=-1.67, \textit{P}=.17  & - & - \\ \midrule 
    {SciFive+MIMIC-T5} 
    & t(df=4)=0.44, \textit{P}=.68 
    & t(df=4)=-0.75, \textit{P}=.50 
    & t(df=4)=2.24, \textit{P}=.09  & - \\
    
            \bottomrule
            \end{tabular} }
\caption{5-fold cross-validation results across models on stigmatizing dataset from Hospital System with 1\% downsampled training data and the t-test results among T5-Sup, FLAN-T5, {MIMIC-T5}, and {SciFive+MIMIC-T5} \label{tab:internal_1_cv}}
\end{table*}

\end{document}